\documentclass{article}
\usepackage{kr}

\usepackage{times}
\usepackage{soul}
\usepackage{url}
\usepackage[hidelinks]{hyperref}
\usepackage[utf8]{inputenc}
\usepackage[small]{caption}
\usepackage{graphicx}
\usepackage{amsmath}
\usepackage{amsthm}
\usepackage{booktabs}
\usepackage{algorithm}
\usepackage{algorithmic}
\usepackage{multirow}

\title{Using Language Models For Knowledge Acquisition in Natural Language Reasoning Problems}

\author{%
Fangzhen Lin$^1$\and
Ziyi Shou$^1$\and
Chengcai Chen$^2$\\
\affiliations
$^1$Department of Computer Science and Engineering, Hong Kong University of Science and Technology\\
$^2$Xiaoi Robot Inc., Shanghai, China\\
\emails
\{flin,zshou\}@cse.ust.hk
}

\begin{document}

\maketitle

\begin{abstract}
  For a natural language problem that requires some non-trivial reasoning to solve, there are at least two
  ways to do it using a large language model (LLM).
  One is to ask it to solve it directly. The other
  is to use it to extract the facts from the problem text and then use a theorem prover to solve it.
  In this note, we compare the two methods using ChatGPT and GPT4 on a series of logic word puzzles, and conclude that the latter is the right approach.
\end{abstract}

\section{Introduction}

ChatGPT, released in November 2022 by OpenAI\footnote{\url{https://chat.openai.com/chat}}, took the world by storm.
There has been many anecdotes of how good it is
in writing essays, composing poems, answering questions and so on. There
was even a report that it could pass a University of Pennsylvania Wharton MBA exam~\footnote{\url{https://mackinstitute.wharton.upenn.edu/wp-content/uploads/2023/01/Christian-Terwiesch-Chat-GTP-1.24.pdf}}. Its successor, GPT4, released in March 2023, has been claimed to be
even better for solving more complex problems.
But there were also anecdotes of the mistakes that they made in solving some simple
mathematics word problems.

In this note, through a series of logic word puzzles, mainly those adapted from
Smullyan's ``Ladies or Tigers?'' puzzles \cite{Smullyan},
  we conclude that ChatGPT is not capable of
  non-trivial reasoning. GPT4, while a little better than ChatGPT, is also nowhere near
  competent in logical reasoning.
  To us, that large language models (LLMs) are not capable of logical reasoning
  is to be expected, and does not need a paper to document it.
  What we want to show is that this is not the right way of using a 
  pure data driven machine learning language model like ChatGPT, however large it is.
  It is not the
  role of such a language model to prove theorems, write correct computer programs, or re-invent
  mathematics and logic. What it could do is as a useful tool to extract knowledge from natural
  language texts, i.e. a knowledge acquisition tool. Indeed, our experiments showed that ChatGPT and GPT4
  did much better when asked to translate the texts in the puzzles to logic sentences. It is
  not perfect, but the potential is clearly there. It can certainly be improved by some further
  training or fine-tuning. This compares with training a language model directly for reasoning, for which
  there has been a large body of work,
  including recent work on chain-of-thought prompting \cite{CoT2022}, 
  a step-by-step explanation of
  how a conclusion is derived. While useful in many Q\&A tasks, it is not enough for tasks that
  require substantial reasoning to solve.
  There has also been
  work on training language models
to perform systematic reasoning such as following some pre-defined rules
(e.g. \cite{talmor2020leap,Hahn20,Creswell22a,Creswell22b,paradox-learning-reasoning22}).
We found the study reported in \cite{paradox-learning-reasoning22} particularly interesting.
It shows that for a simple rule-based logic studied in the paper, BERT can ``attains near-perfect accuracy on in-distribution test examples while failing to generalize to other data distributions over the exact same problem space.''
Perhaps the title of
  a recent survey paper by Helwe {\it et al.} \shortcite{Helwe2021ReasoningWT}, ``Reasoning with Transformer-based Models: Deep Learning, but Shallow Reasoning'', says it all. 

  The rest of the paper describes our experiments and discusses their results.
  Before we do that, two general remarks about our experiments are in order.
One is that given ChatGPT and GPT4 can remember and make use of the earlier interactions in the
same session, we always use a fresh new session whenever necessary.
The other is that 
they do not always give the
same answer to the same question when asked repeatedly.
This always happened with our logic word puzzles.
We conjecture that this is in fact an indication of its lack of
confidence in any of its answers, as in the end, it all comes down to
probabilities. So after trying it out with a couple of ``Ladies or Tigers?'' puzzles initially,
we stopped repeating our queries, and took the first responses that we got from them. One can consider
our experiments as a sample of how these two systems would respond.

\section{Who's in the car?}\label{sec:who-is-in-the-car}

\begin{figure}[!ht]
  \includegraphics[width=\linewidth]{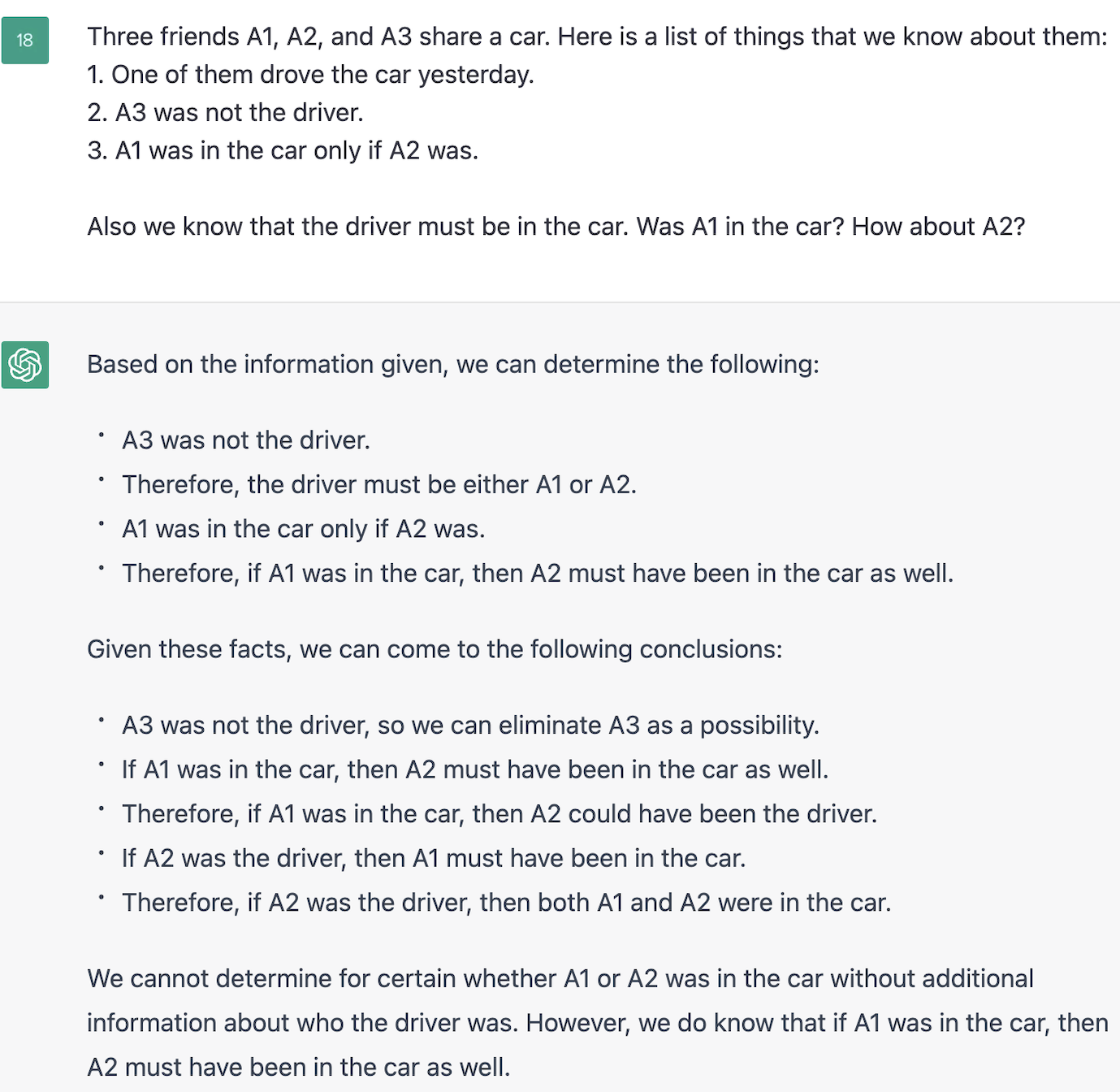}
  \caption{Who's in the car?}
  \label{car1}
\end{figure}

Our first puzzle is about who was in the car, an exam question from an AI course taught by one of the authors:
\begin{quote}{\it
  Three friends A1, A2, and A3 share a car. Here is a list of things that we know about them:
  \begin{enumerate}
\item One of them drove the car yesterday.
\item A3 was not the driver.
\item A1 was in the car only if A2 was.
  \end{enumerate}
Also we know that the driver must be in the car. Was A1 in the car? How about A2?
}
\end{quote}
Figure~\ref{car1} shows the puzzle and a response
by ChatGPT (queried on March 1, 2023).
It was
asked once and never repeated.
The correct answer is that A2 was in the car, but the others may or may not be.
ChatGPT's answer came in three parts. First, for each statement, it determined its fact and any implications from it.
We see that it did this part correctly, determined that the driver must be either
A1 or A2. The second part outlines its reasoning in its attempt to answer the questions. We see it
did not do this part well, drawing problematic conclusions such as ``if A2 was the driver, then A1 must
have been in the car.''  The last part gave its answer which was that not enough information to
determine for certain whether A1 or A2 was in the car. The correct reasoning is of course by case
analysis: if A1 was in the car, then so was A2; if A1 was not in the car, then A2 was
the driver, thus in the car.

\begin{figure}[!ht]
  \includegraphics[width=\linewidth]{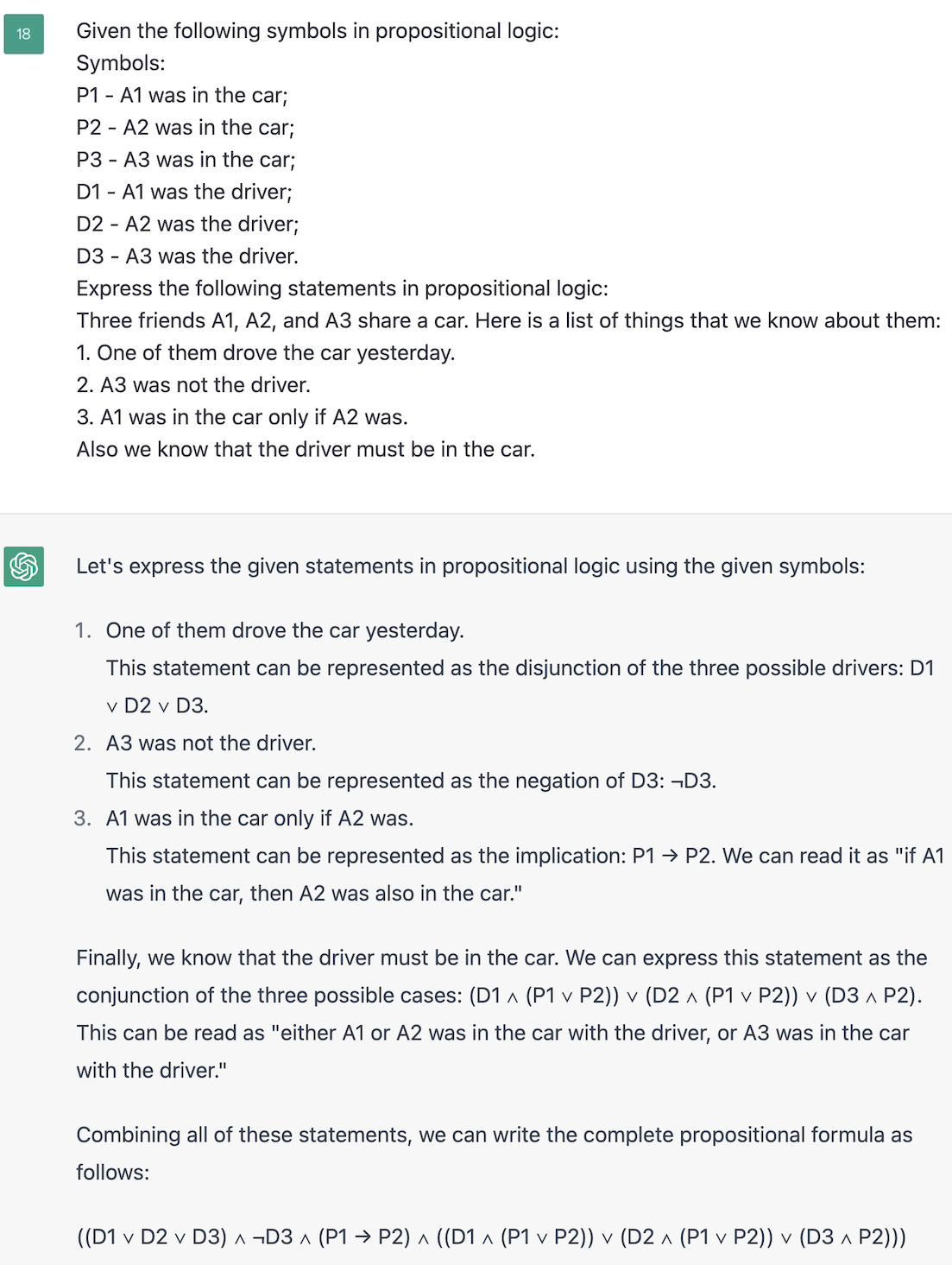}
  \caption{Who's in the car - fact encoding}
  \label{car2}
  \end{figure}

If ChatGPT could understand each statement correctly, why not ask it to just encode it in a logical
language and use a theorem prover to draw conclusions. In other words, use it as an encoder,
a translator, or a knowledge acquisition tool. Figure~\ref{car2} shows how we used it for this
purpose.
To make the task more specific, we provide logical symbols by which the statements are to be encoded.
In the beginning of our experiments with the ``Ladies or Tigers?'' puzzles, we also asked ChatGPT to translate the
statements to propositional logic without providing the symbols. Its responses varied too much.
So we decided to be more specific by providing symbols to test its ability in using pre-defined symbols.
The final generated formula is usable and will indeed give correct answers using any of the
current SAT solvers: the formula entails P2 but not P1. However, there is a glitch:
its translation of the fact that the driver must be in the car is strange, and wrong strictly
speaking.
The straightforward encoding is of course the conjunction of $D_i\rightarrow P_i$, for each $i$.
Somehow it decided to enumerate the cases and concluded that P2 goes with D3:
$(D_1\land (P_1\lor P_2))\lor ( D_2\land (P_1\lor P_2))\lor (D_3\lor P_2)$.

This shows the importance of checking the generated sentences 
before submitting the final KB to a theorem prover. Checking the correctness of a translated sentence is of course easier than checking the correctness of an answer to a complex question, and we
consider this a key advantage of using LLMs such as ChatGPT as a knowledge acquisition/encoding tool rather than a
problem solver.

Having done our experiment on ChatGPT, we repeated the two queries, one to solve the puzzle and the other to translate the text to logic, using GPT4 when it was released on
March 14, 2023. It answered them both correctly - see Appendix \ref{sec:gpt4-car} for the screenshots of our queries and GPT4's answers.
So GPT4 is more capable. However, it is not by much. It is easy to come up with similar
puzzles to show its limitation in reasoning.

\section{The Alpine Club}

Our next puzzle to test GPT4 is one about members of the Alpine Club adapted from
Exercise 16.4 in Nilsson's AI textbook \cite{Nilsson98}:
\begin{quote}{\it
Tony, Mike, and John belong to the Alpine Club. Every member of the Alpine Club is
either a skier or a mountain climber or both. No mountain climber likes rain, and all
skiers like snow. Mike dislikes whatever Tony likes and likes whatever Tony dislikes.
Tony likes rain and snow.

Who is a member of the Alpine Club who is a mountain climber but not a skier?
  }
\end{quote}
The answer is ``Mike.'' Figure~\ref{club1} below shows the response from GPT4 (queried on March 29, 2023).
\begin{figure}[!ht]
  \includegraphics[width=\linewidth]{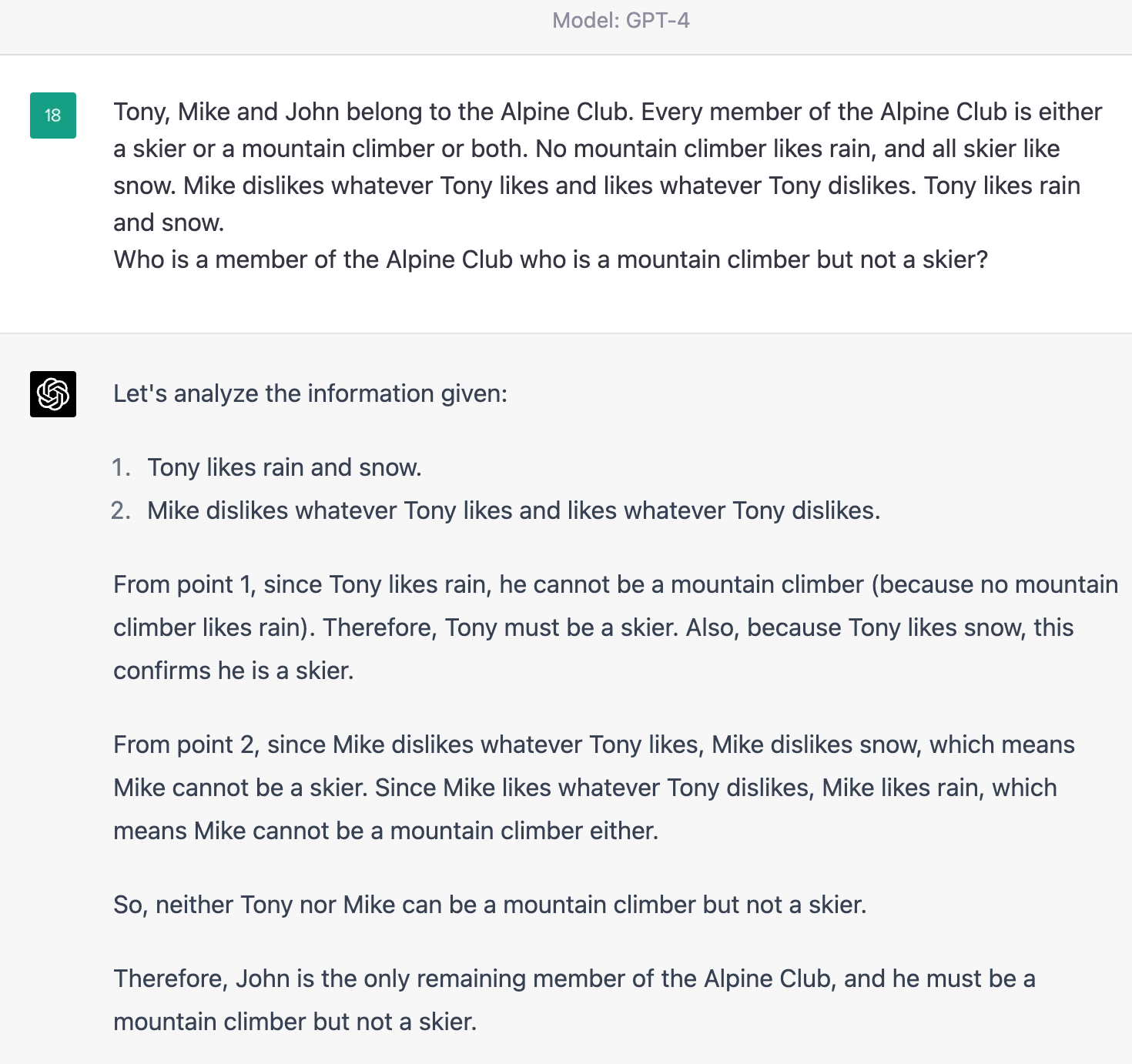}
  \caption{Who's a mountain climber - direct query}
  \label{club1}
  \end{figure}

\begin{figure}[!ht]
  \includegraphics[width=\linewidth]{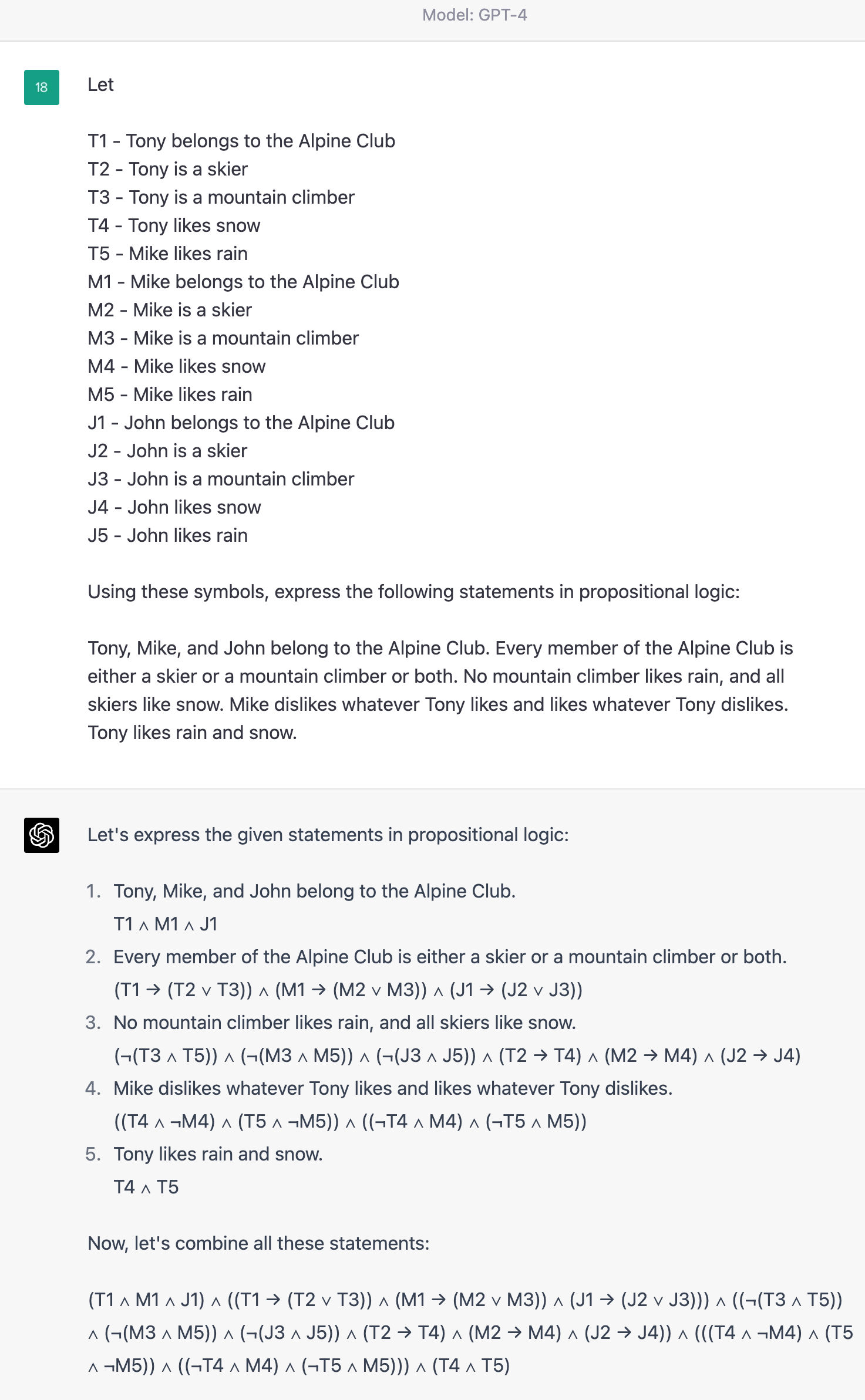}
  \caption{Who's a mountain climber - fact encoding}
  \label{club2}
  \end{figure}

One can see that both its answer and reasoning were wrong. It's clear that GPT4 did not
understand the statement ``Mike dislikes whatever Tony likes and likes whatever Tony dislikes.'' Indeed, when we asked it to translate the text to logic, it got it right except for
that statement, which was somehow translated  into a contradictory sentence.
Figure~\ref{club2} contains the screenshot of GPT4's response (queried on March 29, 2023).

\section{Ladies or Tigers?}

Our next experiment uses puzzles adapted from Smullyan's ``Ladies or Tigers?''
puzzles \cite{Smullyan}, which were based on the 1882 short story ``The Lady, or the Tiger?''
by Stockton \shortcite{Stockton}.
  The Wikipedia page for it says ```The Lady, or the Tiger?'
has entered the English language as an allegorical expression, a shorthand indication or signifier, for a problem that is unsolvable.''\footnote{\url{https://en.wikipedia.org/wiki/The_Lady,_or_the_Tiger}, accessed on March 5, 2023.}

\begin{figure}[!ht]
  \includegraphics[width=\linewidth]{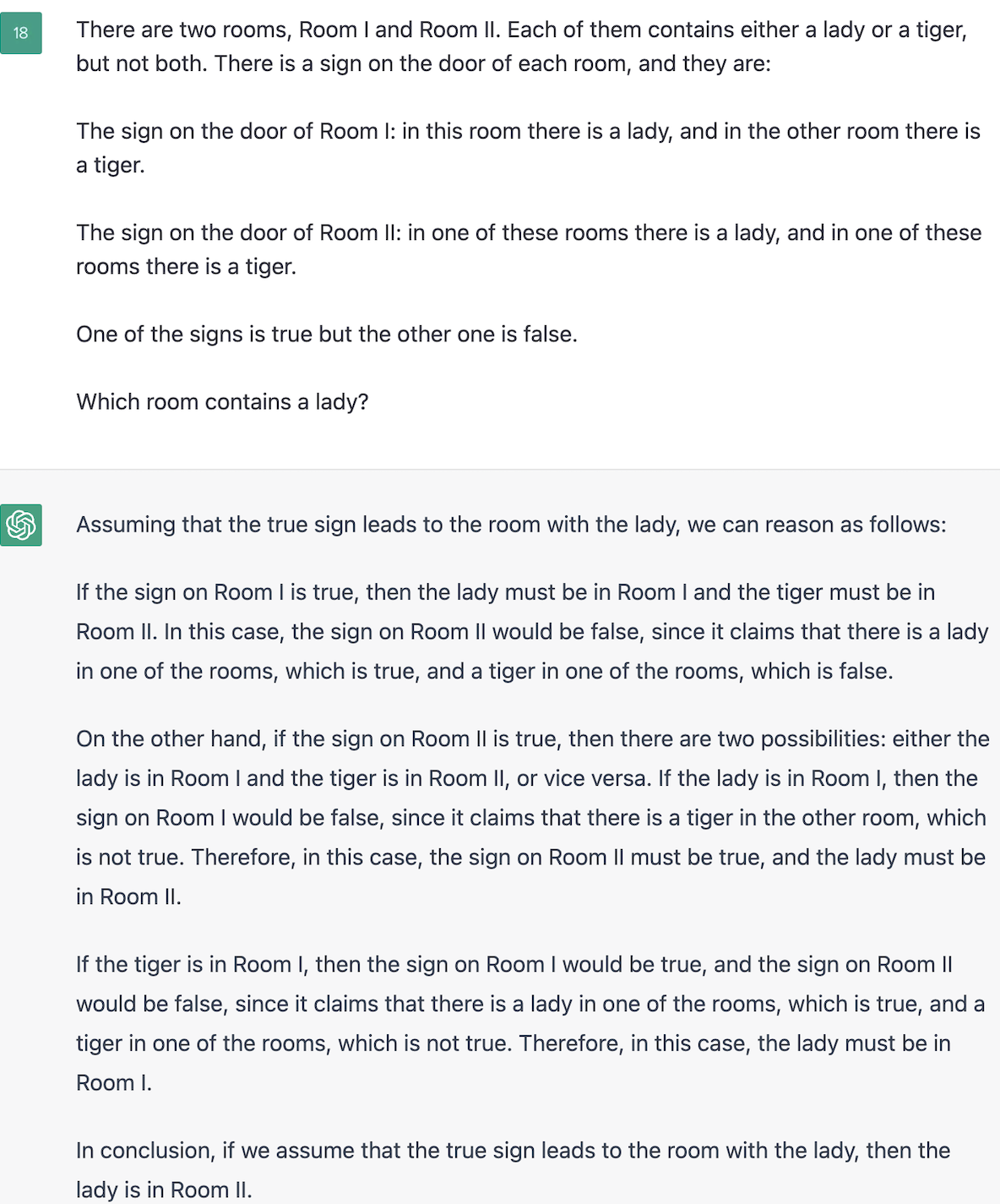}
  \caption{The first trial}
  \label{lot1}
  \end{figure}

Figure~\ref{lot1} shows the puzzle that corresponds to the First Trial in Smullyan's book, and a response from ChatGPT.
The correct answer is that Room II contains a lady. ChatGPT concluded that ``if we assume that the true
sign leads to the room with the lady, then the lady is in Room II,'' a useless conditional answer.
For one, the condition is actually false. For another, such a conditional answer is like kicking the can back to the user.

The problem is not just with this wrong type of answer. One can see that ChatGPT made many mistakes in
its reasoning. For example, it somehow concluded that
if the sign on Room I is true, then the sign on Room II would be false, when the opposite is true:
if the sign on Room I is true, then the sign on Room II would be true as well.

\begin{figure}[!ht]
  \includegraphics[width=\linewidth]{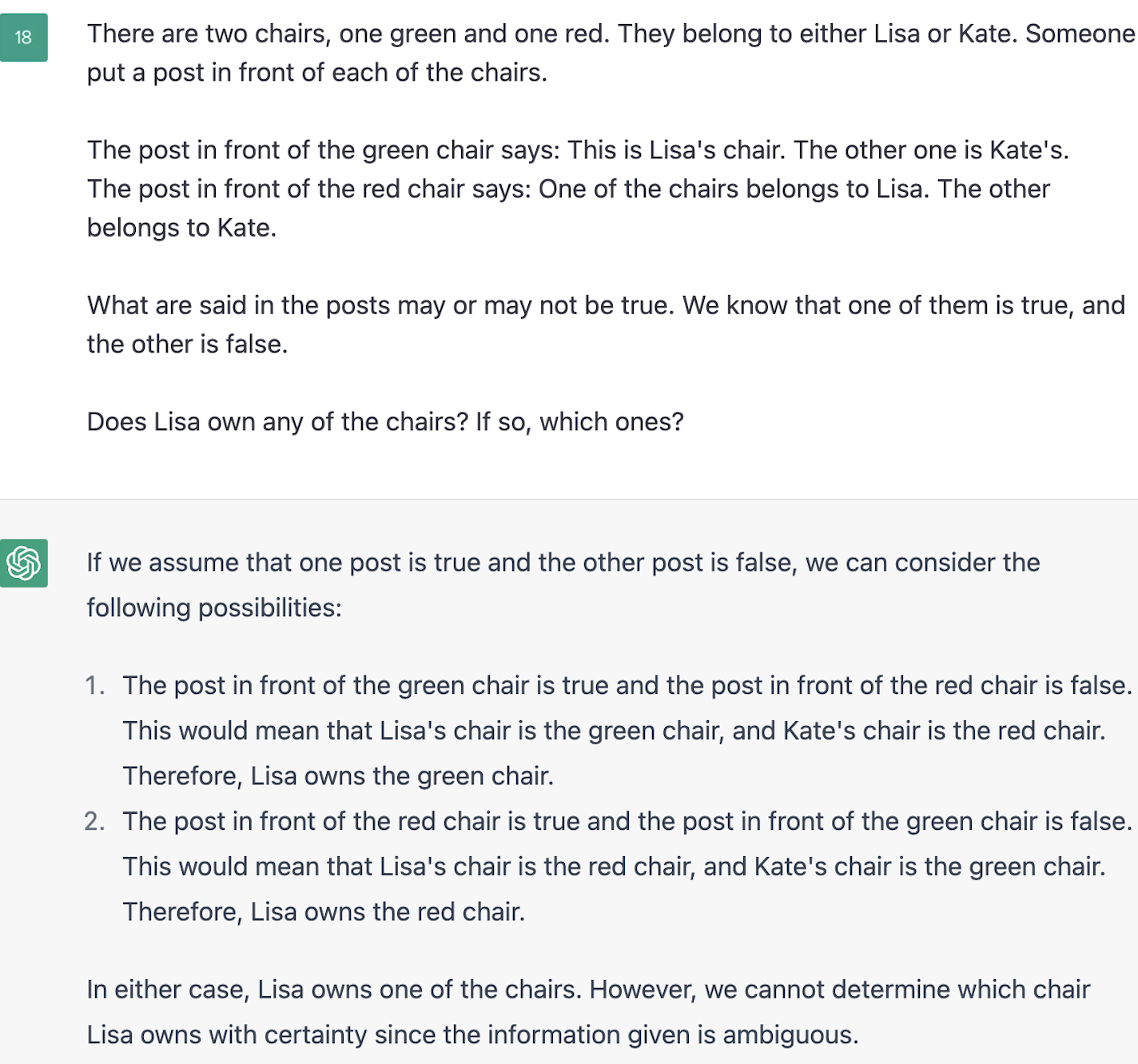}
  \caption{Whose chair}
  \label{chair1}
  \end{figure}

ChatGPT's conclusion and the reasoning for this puzzle is puzzling (no puns intended). One possibility is
data contamination. 
Given that the puzzles and their answers have
been widely available on
internet, they could have been in ChatGPT's training set already. 
It is a common practice in work on few-shot learning with LLMs to do
a test set contamination analysis  (e.g. \cite{Brownetal2020}).
Since we have no access to ChatGPT's training data, there is no way we can do this.
But it dose not really matter here as ChatGPT performed badly on the 12 puzzles that correspond to
the 12 trials given in Smullyan's book.
Still, we decided to use a completely different story line to create a new puzzle with the same logical structure. 
Figure~\ref{chair1} shows our session with ChatGPT on the new puzzle.\\[1ex]
As one can see, ChatGPT's reasoning is sound but it just could not come to the conclusion that the sign
in front of the green chair is false, thus Lisa owns the red chair. This suggests that the
``Ladies or Tigers?'' one may indeed have been contaminated - it wanted to conclude a lady is in Room II but could not.

\begin{figure}[!ht]
  \includegraphics[width=\linewidth]{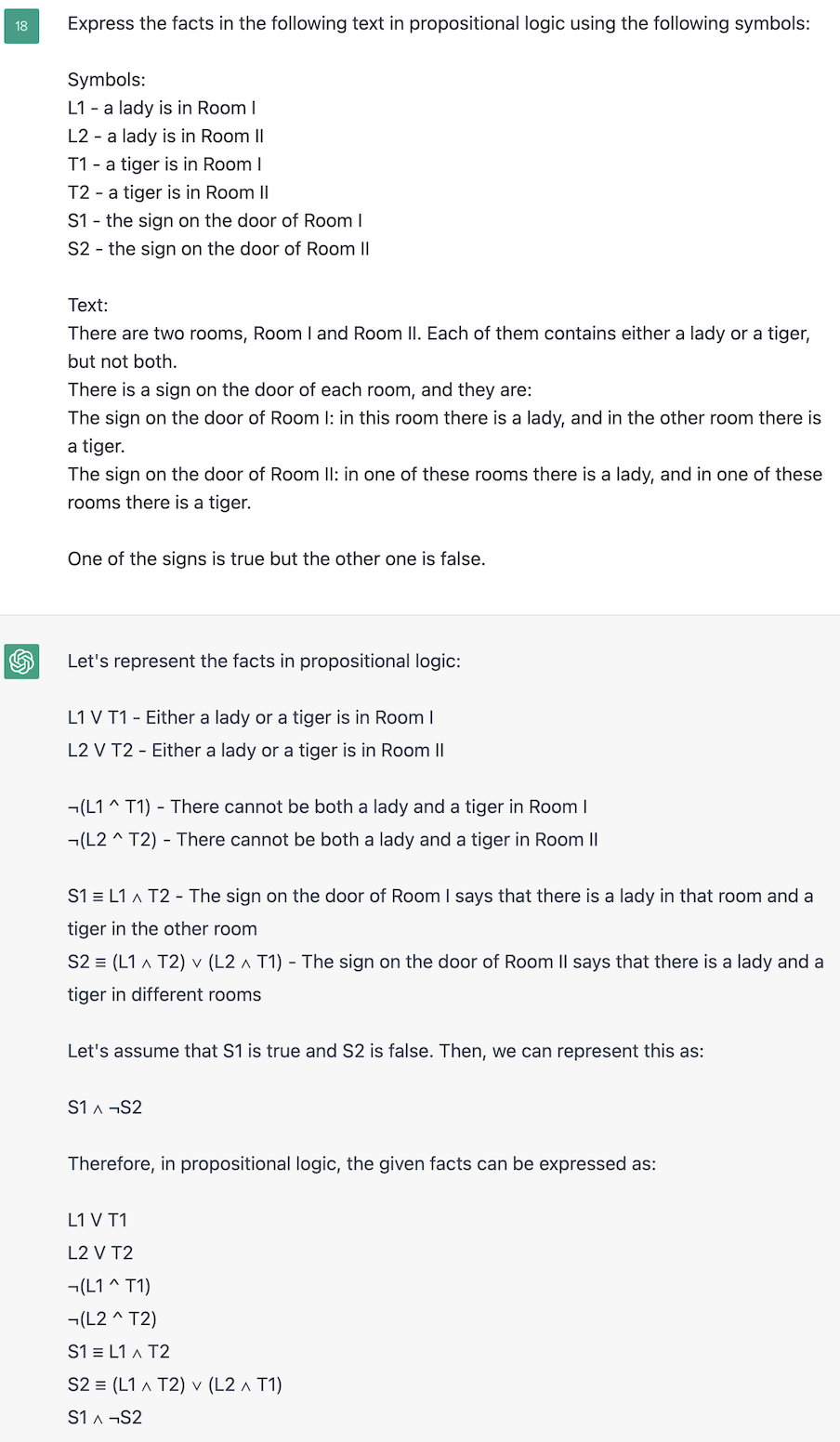}
  \caption{The first trial - facts in one go}
  \label{lot1-fact}
  \end{figure}

Again we can see that ChatGPT could not perform the required reasoning.
We now consider using it
for knowledge acquisition.
Figure~\ref{lot1-fact}
shows how we asked ChatGPT to translate the text of the
puzzle  to logic
using the provided symbols.
As one can see, the result is correct except that it refused to encode the statement ``one of the
signs is true but the other is false.'' Instead, it took the liberty to assume 
that the sign on
room I is true and the one on room II is false, a strange behavior that is consistent with the
observations that others have made about reasoning in language models (e.g. \cite{saparov2023language}).

This motivates us to ask ChatGPT to encode the statements one at a time.
For this example, the response that we got from ChatGPT was correct (screenshot given in Appendix as supplementary materials).

For GPT4, it gave a correct final answer (a lady in Room II) to the first trial of the ``Ladies or Tigers?'' puzzle, but an incorrect reasoning: it considered only two out of the four possible scenarios of what the two rooms can contain. It gave a correct translation from the puzzle text
to logic. The screenshots are given in Appendix.

Table~\ref{lot-summary}
summaries our experiments with ChatGPT and GPT4 using the 12 puzzles that correspond
to the 12 trials in Chapter 2 ``Ladies or Tigers?'' in \cite{Smullyan}. The screenshots of all the
interactions with them are included in Appendix. 
 For GPT4, queries for Trial I were done on March 15, 2023, and for the rest on March 30, 2023. A notable difference between GPT4's responses on these
 two dates is that 
 it encoded the statement ``There are two rooms, Room I and Room II. Each of them contains either a lady or a tiger, but not both'' for Trial I correctly, but not for
Trials II - VIII and similar statements in other trials. Its encoding of this statement for
Trial II is
\[
(L1\land\neg T1)\land (L2\land \neg T2)\land (T1\land\neg L1)\land (T2\land \neg L2),
\]
which is a contradictory sentence as it put the four cases together as a conjunction
instead of a disjunction.

\begin{table*}[t]
\begin{center}
  \begin{tabular}{| r | c|c | c|c | c|c | c|c |} 
    \hline
    \multirow{2}{*}{Trials} & \multicolumn{2}{c|}{QA} & \multicolumn{2}{c|}{QR} & \multicolumn{2}{c|}{T2L} & \multicolumn{2}{c|}{S2L} \\ \cline{2-9} 
    & ChatGPT & GPT4 & ChatGPT & GPT4 & ChatGPT & GPT4 & ChatGPT & GPT4 \\ \hline\hline

 1 &no & yes &no & no &6/7 & 5/5 &4/4 & 4/4 \\ \hline
 2 &no& no &no & no &3/5 & 4/5 &3/4& 2/4 \\ \hline
 3 & incomplete & incomplete & no & no &6/6& 4/4 &4/4& 3/4 \\ \hline
 4&yes & yes &no & yes &4/5 & 5/5 &5/5& 4/5 \\ \hline
 5&yes& yes &no & no &6/6 & 5/5 &3/5& 4/5 \\ \hline
 6& no & yes &no & yes &2/6 & 5/5 &3/5& 4/5 \\ \hline
 7&no & yes &no & yes &4/6& 4/5 &3/5& 4/5 \\ \hline
 8&no & yes &no & no &3/6& 5/5 &3/4& 2/5 \\ \hline
 9&no & yes &no & no &6/6& 6/6 &3/3& 5/6 \\ \hline
 10&no & no &no & no &7/11 & 3/5 &4/5& 4/6 \\ \hline
 11&no & no &no & no &3/9 & 3/5 &6/7& 6/7 \\ \hline
 12&no & no &no& no &6/12 & 21/21 &12/13& 13/13 \\ \hline
  \end{tabular}
\end{center}
Legends: QA - whether ChatGPT's final answer to the puzzle is correct; QR - whether the reasoning is correct; T2L - 
the ratio of correct sentences when the puzzle is given as a text for it to translate; S2L - the
ratio of correct sentences when the text is broken into statements and asked one at a time.
  \caption{The Lady or the Tiger puzzles}\label{lot-summary}

\end{table*}

\section{Translating statements to logic}\label{sec:s2l}

To further evaluate the ability of ChatGPT and GPT4 in encoding natural language statements to propositional logic,
we created 15 example sentences about three balls and their colors. They were given as a single query by
asking them to encode each of them in propositional logic using the provided symbols. The
results given below were the first responses from ChatGPT and GPT4, respectively, and we did not repeat the query. Again the 
screenshots are included in Appendix 
\\[1ex]
Task: Given the following propositional symbols:
\begin{center}
  \begin{tabular}{ll}
X1 -  A is a red ball \ \ & X2 - A is a green ball\\
X3 - A is a white ball & Y1 - B is a red ball\\
Y2 - B is a green ball & Y3 - B is a white ball\\
Z1 - C is a red ball & Z2 - C is a green ball\\
Z3 - C is a white ball &
  \end{tabular}
\end{center}
express each of the following statements in propositional logic:
\begin{enumerate}
\item There are three balls: A, B, and C. Exactly one of them is red.\\
  ChatGPT (wrong): (X1 XOR X2) XOR X3 AND NOT (X1 AND X2)\\
  GPT4 (wrong): (X1$\land$ $\neg$Y1$\land$ $\neg$Z1)$\land$ ($\neg$X1$\land$ Y1$\land$ $\neg$Z1)$\land$ ($\neg$X1$\land$ $\neg$Y1$\land$ Z1)
\item There are three balls: A, B, and C. At least one of them is red.\\
  ChatGPT (wrong): X1 OR X2 OR X3\\
  GPT4 (correct):  X1$\lor$ Y1$\lor$ Z1
\item There are two balls: A and C. One of them is red, the other is white.\\
  ChatGPT and GPT4 (correct): (X1 AND Z3) OR (X3 AND Z1)
\item There are two balls: A and C. A is green. The other one is white.\\
  ChatGPT and GPT4 (correct): X2 AND Z3
\item If A is a red ball, so is B and C.\\
  ChatGPT and GPT4 (correct):  X1 $\rightarrow$ Y1 AND Z1
\item If A is a red ball and B is a green ball, then C is a white ball.\\
  ChatGPT and GPT4 (correct): X1 AND Y2 $\rightarrow$ Z3
\item If A is not a green ball, then it is a red ball.\\
  ChatGPT and GPT4 (correct): NOT X2 $\rightarrow$ X1
\item A is a red ball if and only if B is a green ball.\\
  ChatGPT and GPT4 (correct): X1 $\leftrightarrow$ Y2
\item There are three balls: A, B, and C. It cannot be the case that all these three balls are red.\\
  ChatGPT and GPT4 (correct): NOT (X1 AND Y1 AND Z1)
\item The ball A is both red and green.\\
  ChatGPT and GPT4 (correct): X1 AND X2
\item The ball B is neither red nor green.\\
  ChatGPT (correct): NOT (Y1 OR Y2)\\
  GPT4 (correct) $\neg$Y1 $\land$ $\neg$Y2
\item Among the possible colors red, green and white, the ball C can have at most two colors.\\
  ChatGPT (wrong): (Z1 AND Z2) OR (Z1 AND Z3) OR (Z2 AND Z3)\\
GPT4 (wrong):  ($\neg$Z1$\lor$ $\neg$Z2) $\land$ ($\neg$Z1$\lor$ $\neg$Z3) $\land$ ($\neg$Z2$\lor$ $\neg$Z3)
\item There are three balls, A, B, and C. Any of them can be white but none of them is green.\\
  ChatGPT and GPT4 (wrong?): (X3 OR Y3 OR Z3) AND NOT (X2 OR Y2 OR Z2)
\item The balls A and B are of different colors.\\
  ChatGPT (wrong): (X1 AND Y2) OR (X2 AND Y1)\\
GPT4 (wrong?):  (X1 $\land$ $\neg$Y1)$\lor$ (X2 $\land$ $\neg$Y2)$\lor$ (X3 $\land$ $\neg$Y3)$\lor$ ($\neg$X1 $\land$ Y1)$\lor$ ($\neg$X2 $\land$ Y2)$\lor$ ($\neg$X3 $\land$ Y3)
\item If A is a red ball, then B and C are balls of the same color.\\
  ChatGPT (wrong):  X1 $\rightarrow$ Y2 AND Z2\\
GPT4 (correct):  X1 $\rightarrow$ ((Y1 $\land$ Z1)$\lor$ (Y2 $\land$ Z2)$\lor$ (Y3 $\land$ Z3))
\end{enumerate}
One can see that ChatGPT did poorly on statements with the terms ``exactly one'',
``at most two'', and ``at least one''. It did well on the others. Its answers to 
statements (13) - (15) are interesting. We put a question mark ``wrong?'' on (13): does ``any of them
can be white'' implies ``one of them is white?'' We think not, but ChatGPT apparently assumed that.
ChatGPT's answers to (14) and (15) are interesting because they revealed that it took the liberty to
assume that there are only two colors (for 14) or the same color means color ``green'' (for 15),
the same type of presumption that we saw earlier with the ``Ladies or Tigers?'' puzzle.
In comparison, GPT4 did much better. It got (2) wrong for the same reason as it did for some
of the statements in ``Ladies or Tigers?'' - putting cases together using conjunctions
instead of disjunctions.
It still could not handle ``at most two'' in (12), and
still assume that a ball must have one of the given color in (14).

\section{Concluding remarks}

Again the main point of this paper is that one should not use a LLM like ChatGPT for
tasks that require non-trivial reasoning. Instead, one should consider using it as a tool
for knowledge acquisition from natural language texts. Thus
the research direction that we are
advocating is two fold - to use ChatGPT and possibly other LLMs as a front end for complex
reasoning tasks and to train better LLMs for this purpose.

What we are advocating here is in line with neuro-symbolic AI, reliable and explainable AI, and in general
the need to have high-level cognitive reasoning in any general AI systems. Our specific call for 
training better language models for capturing facts in a text is related to work on pronoun resolution, relation and entity extraction, abstract meaning
representation, and others. The connections are
clear but it is beyond the scope of
this short note to discuss them in detail here.

\section*{Acknowledgement}

Our thanks to Yan Yang and Zhijin Wang for motivating discussions related to the topics of
this work.

\bibliographystyle{kr}
\bibliography{kr23}
    
\clearpage
\appendix

The appendix contains the screenshots of our sessions with ChatGPT and GPT4. It is divided into four sections: the GPT4 sessions for ``Who's in the car'' mentioned in Section~\ref{sec:who-is-in-the-car}, the ChatGPT and GPT4 sessions for translating statements to propositional logic mentioned in Section~\ref{sec:s2l}, the ChatGPT sessions for all ``Ladies or Tigers?'' puzzles, and the GPT4 sessions for all ``Ladies or Tigers?'' puzzles.

\section{Who’s in the car}\label{sec:gpt4-car}
This section contains the screenshots of our sessions with GPT4 (March 15, 2023) on the 'Who's in the car'' problem in Section \ref{sec:who-is-in-the-car}.\\
Asking it to solve the puzzle directly:\\
\includegraphics[width=\columnwidth]{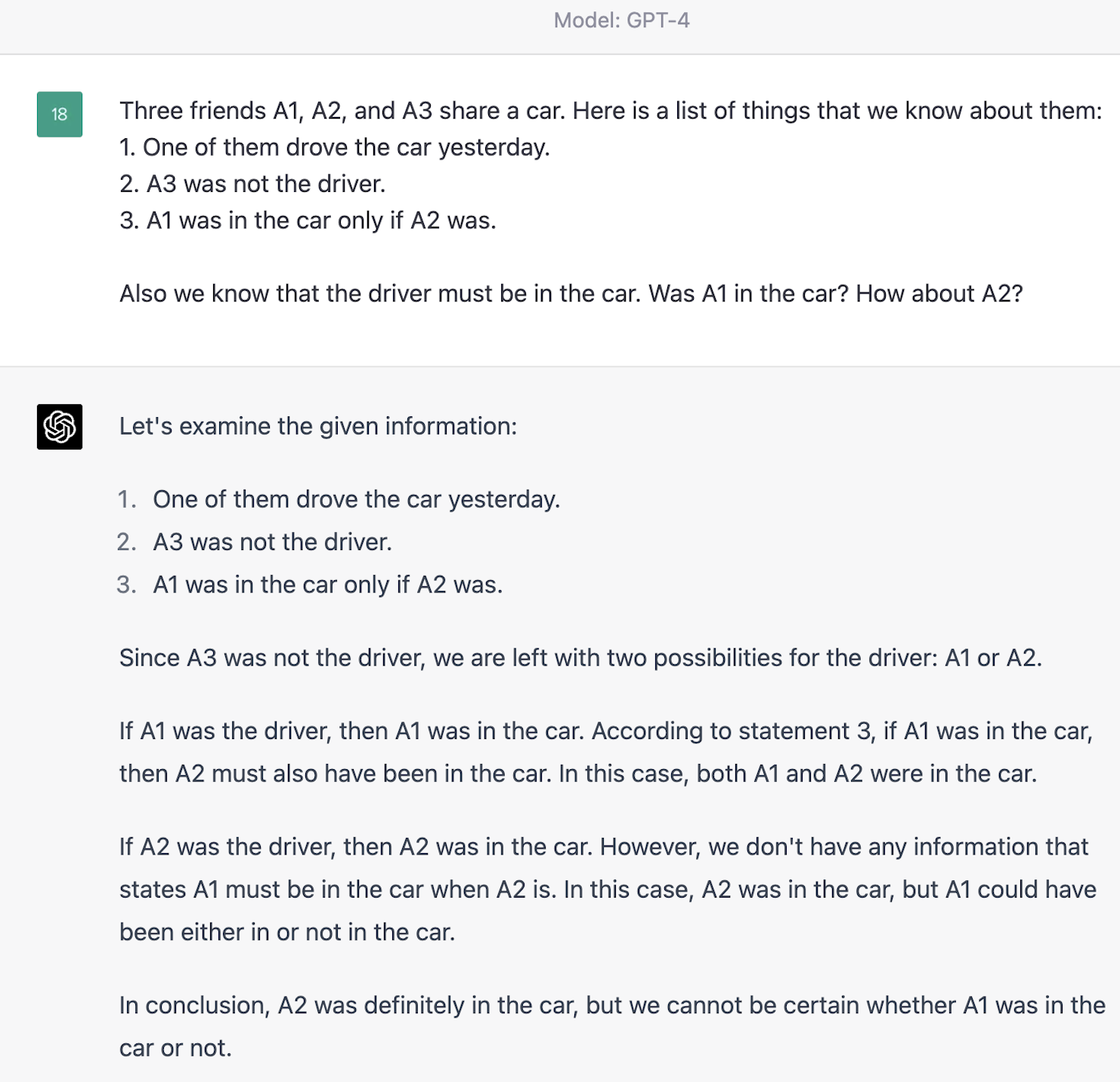}
\newpage
\noindent
Asking it to translate the text to logic:\\
\includegraphics[width=\columnwidth]{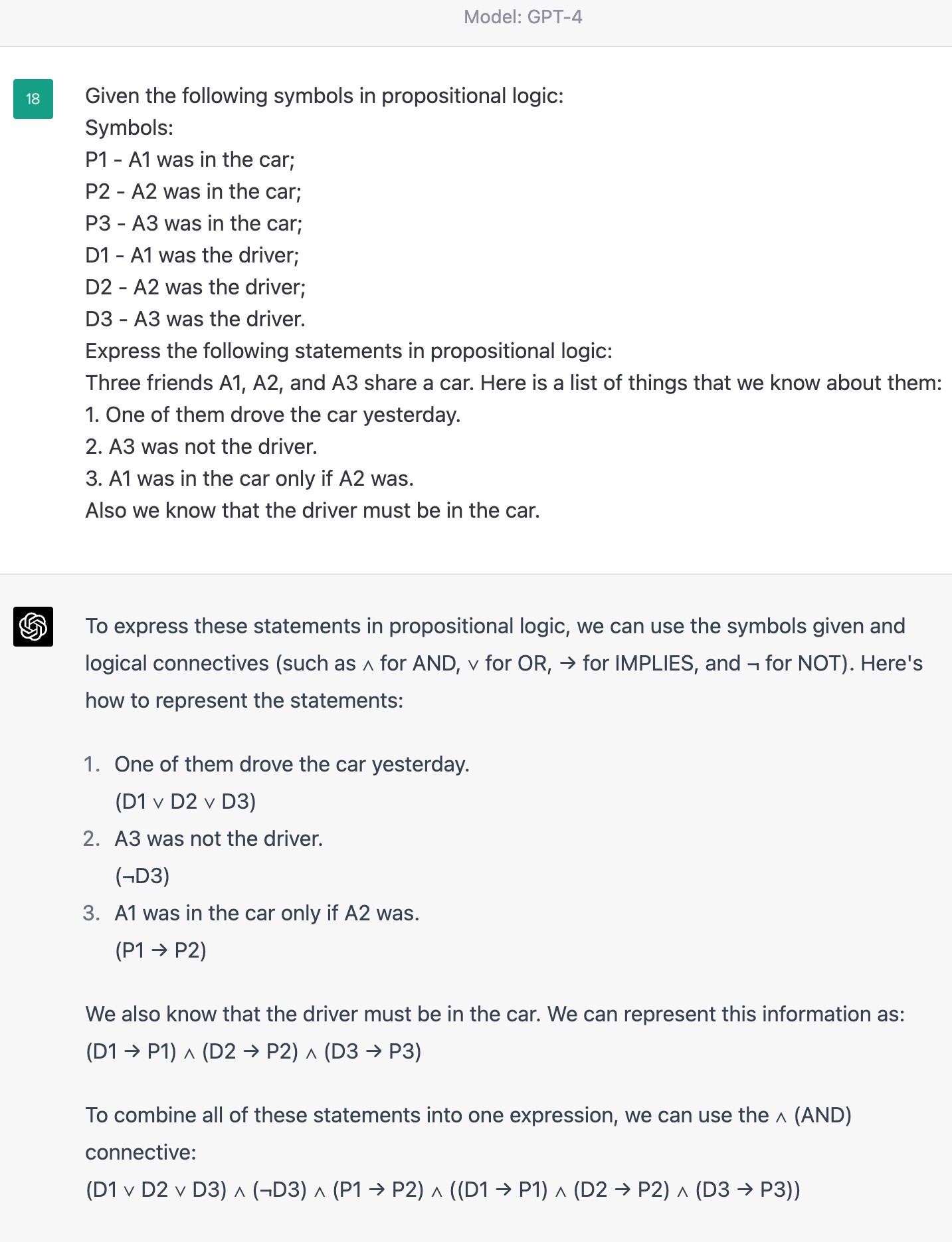}

\newpage
\section{Translating statements to logic}
The following is the screenshot of our session with ChatGPT (March 7, 2023) asking it to translate the 15 statements in Section 4 to propositional logic:\\[1ex]
\includegraphics[width=\columnwidth]{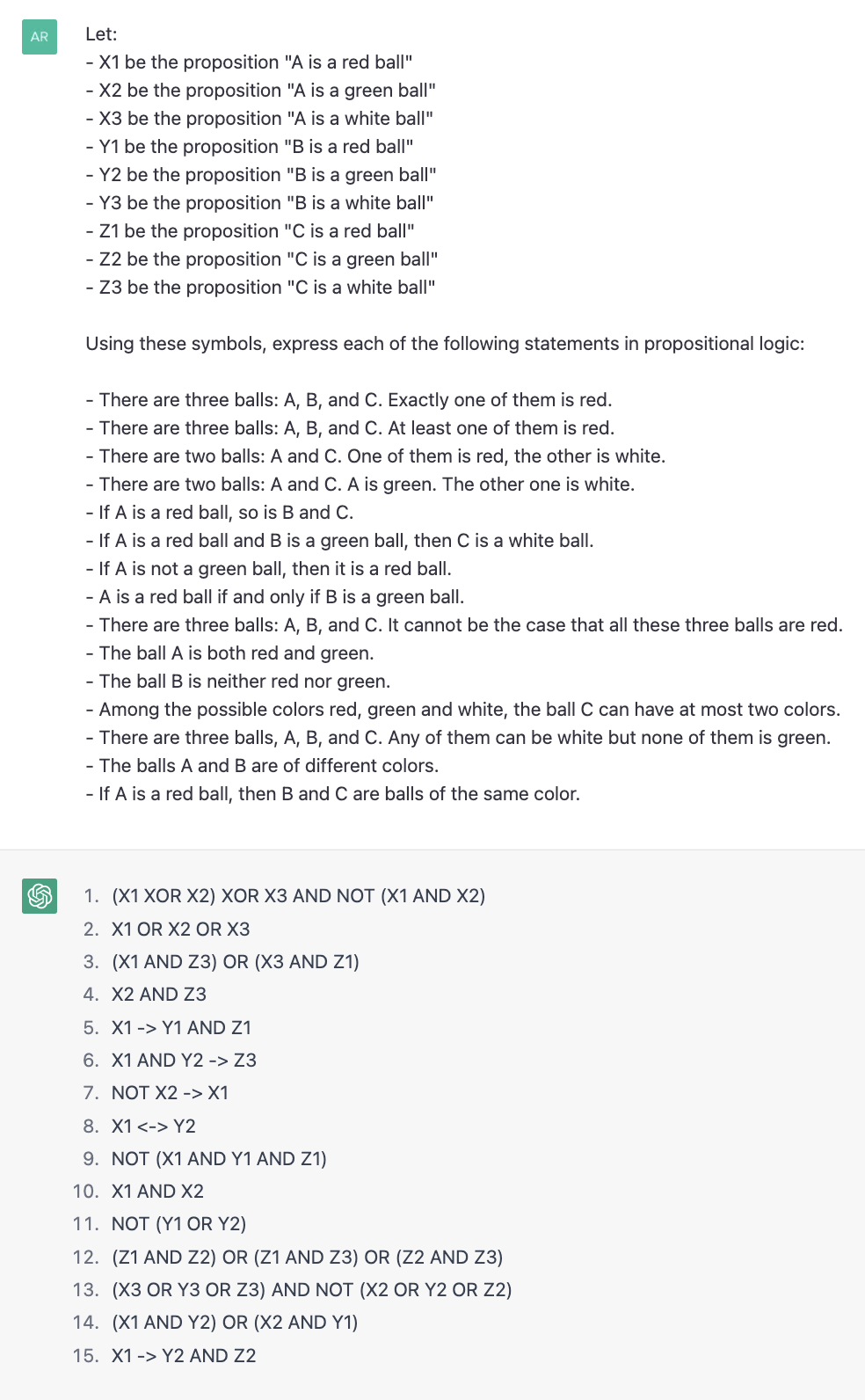}
\newpage\noindent
The following is the screenshot of our session with GPT4 (March 15, 2023):\\[1ex]
\includegraphics[width=\columnwidth]{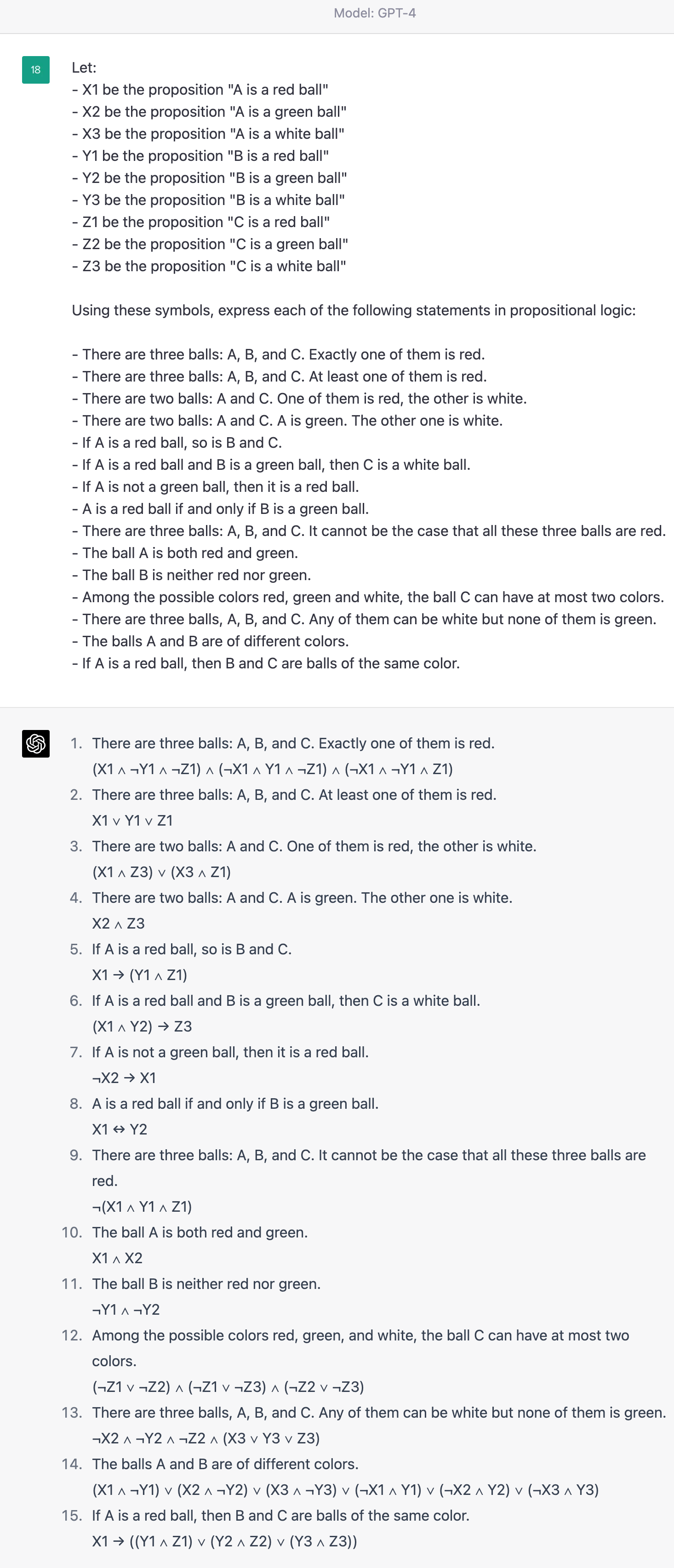}

\newpage

\section{Ladies or Tigers? - ChatGPT}\label{sec:appendix-lot}
This section contains the screenshots of our sessions with ChatGPT (February 28, 2023)  on the 12 ``Ladies or Tigers?'' puzzles.

\subsection*{Trial I}
Direct query to ChatGPT:\\
\includegraphics[width=\columnwidth]{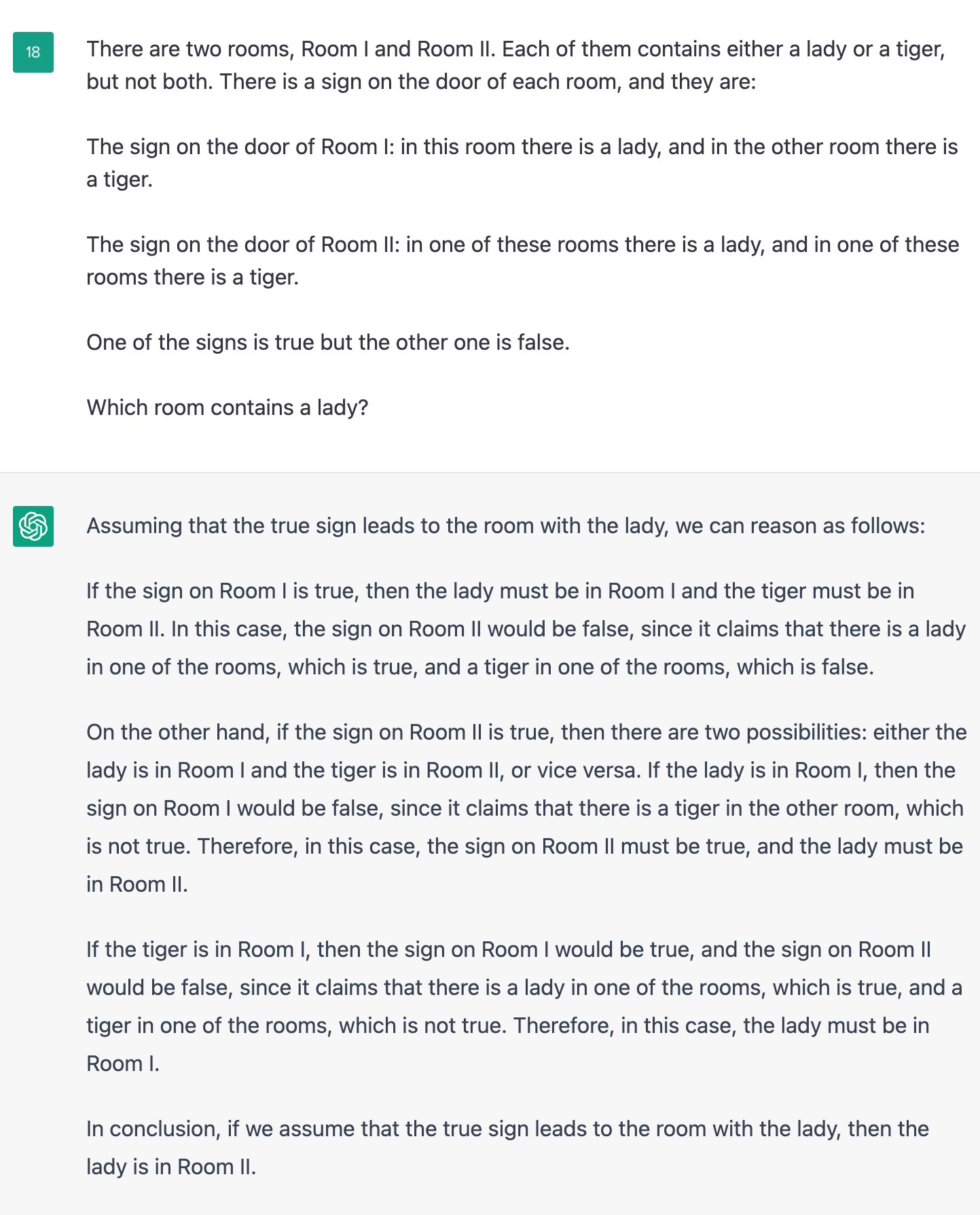}

\newpage
\noindent 
Asking it to translate the text to logic:\\
\includegraphics[width=\columnwidth]{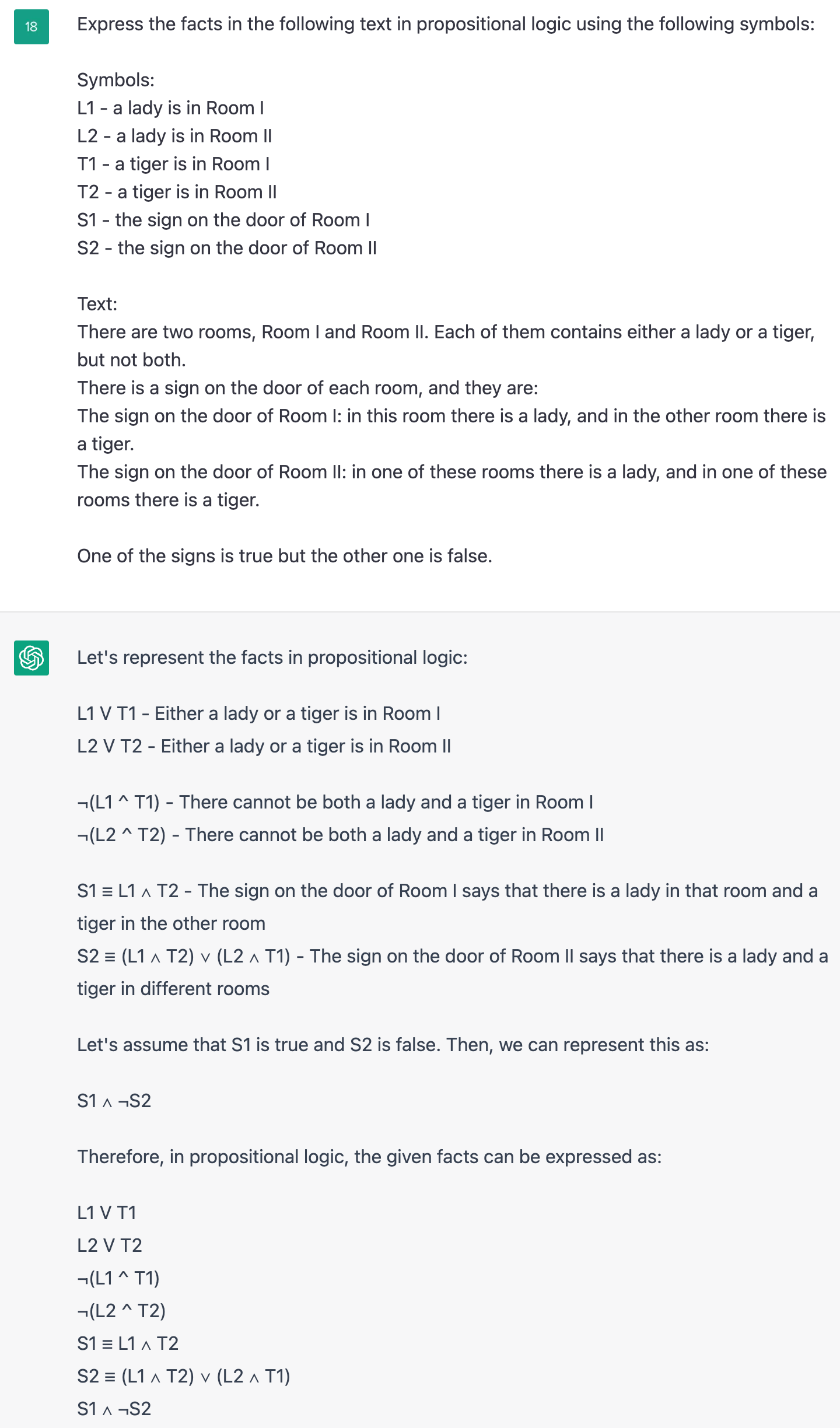}

\newpage
\noindent 
Asking it to translate the statements in the text one at a time:\\
\includegraphics[width=.86\columnwidth]{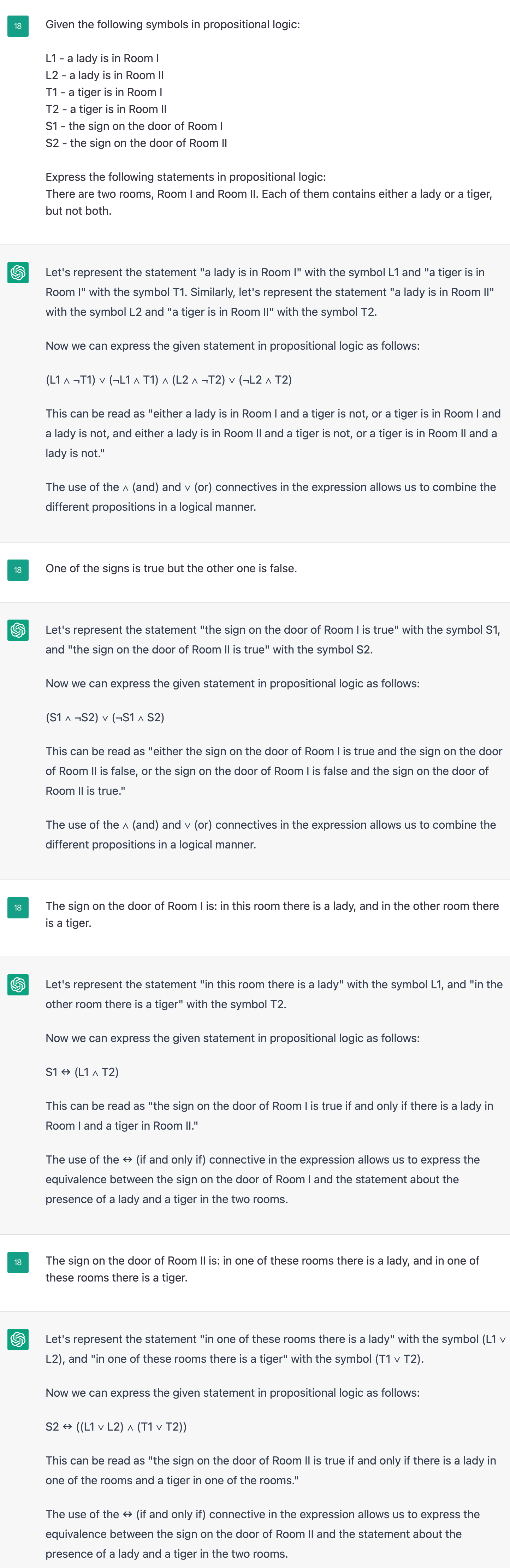}

\newpage
\subsection*{Trial II}
Direct query to ChatGPT:\\
\includegraphics[width=\columnwidth]{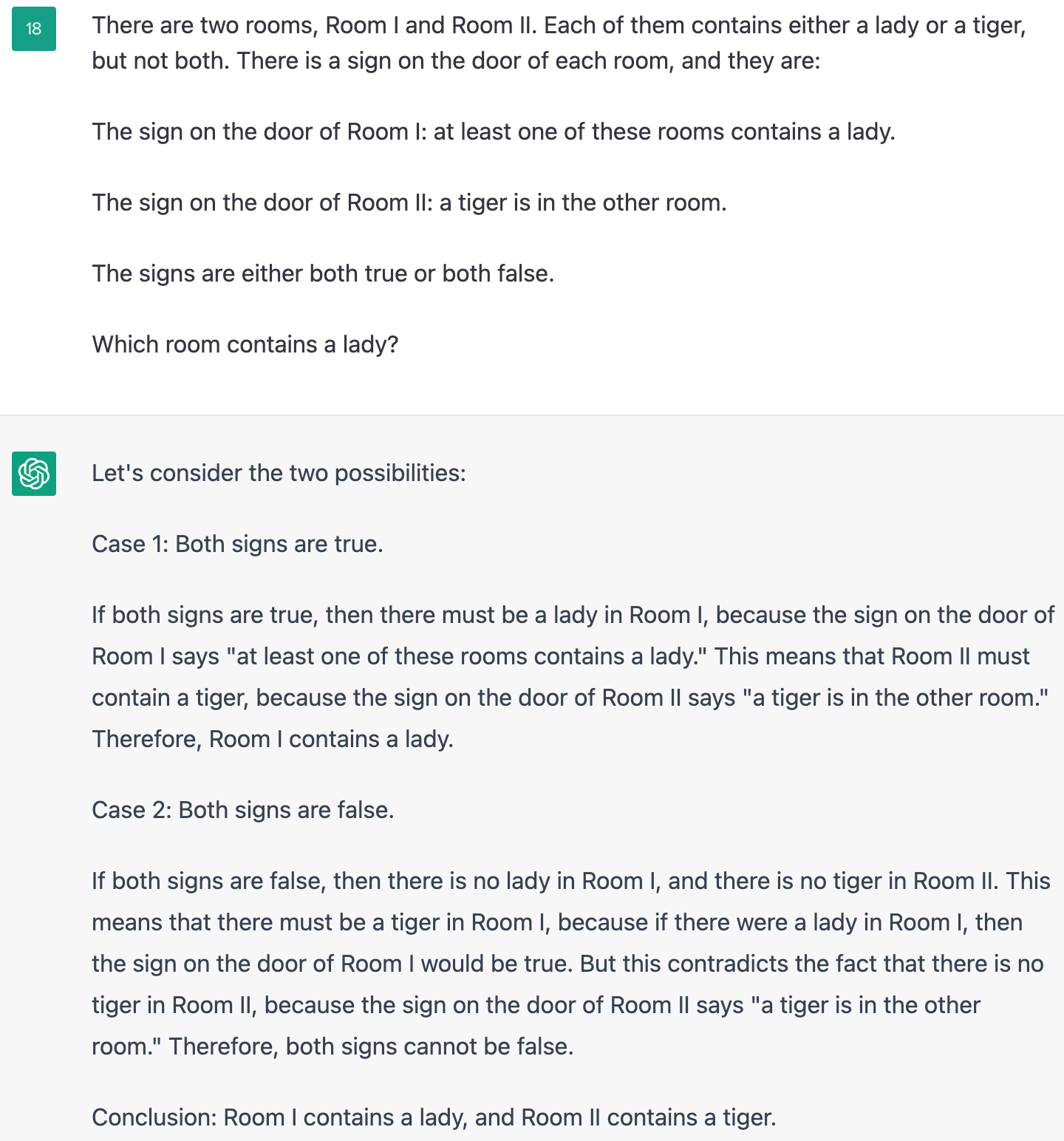}
\\
Asking it to translate the text to logic:\\
\includegraphics[width=\columnwidth]{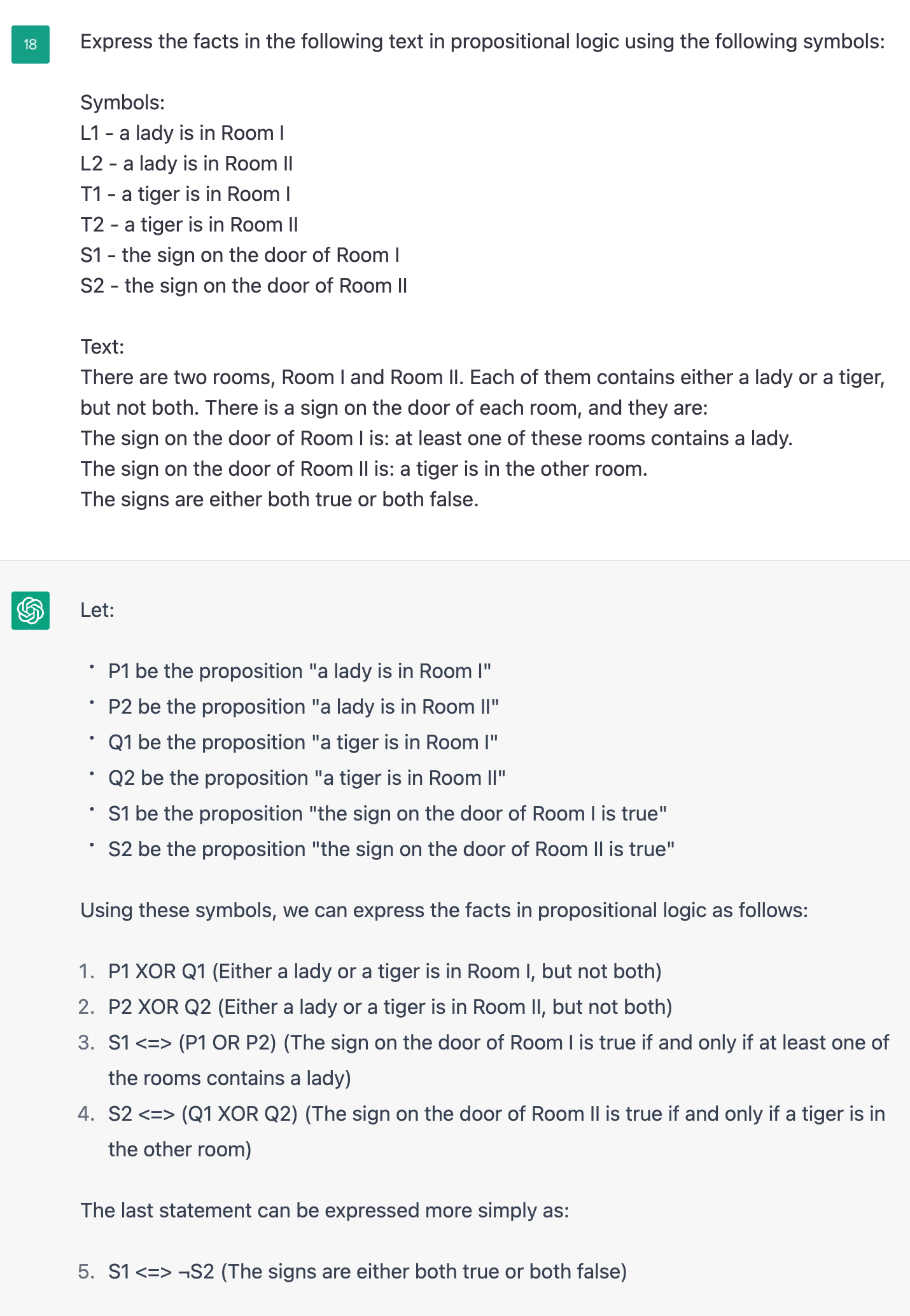}

\newpage
\noindent
Asking it to translate the statements in the text one at a time:\\[1ex]
\includegraphics[width=\columnwidth]{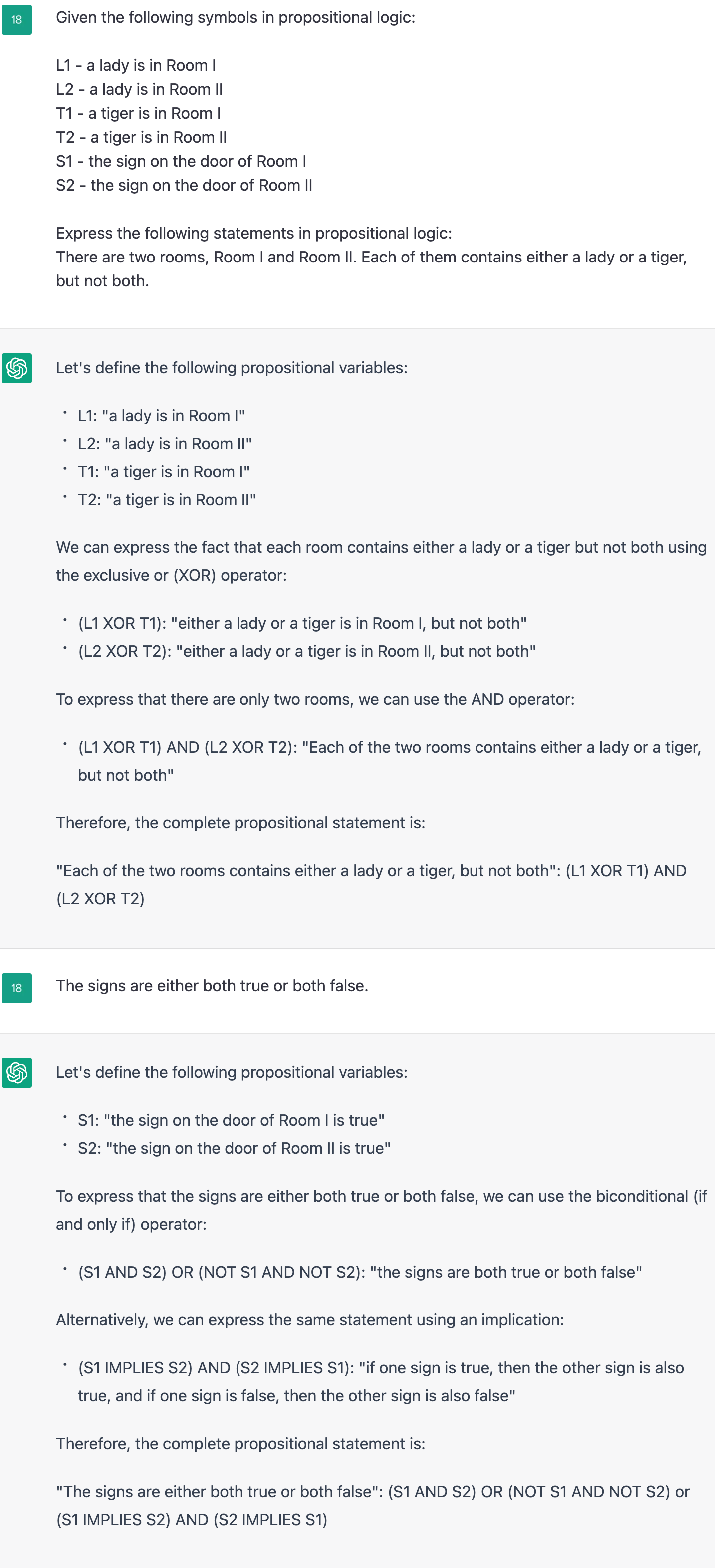}
\newpage
\noindent
Asking it to translate the statements in the text one at a time: (continued)\\
\includegraphics[width=\columnwidth]{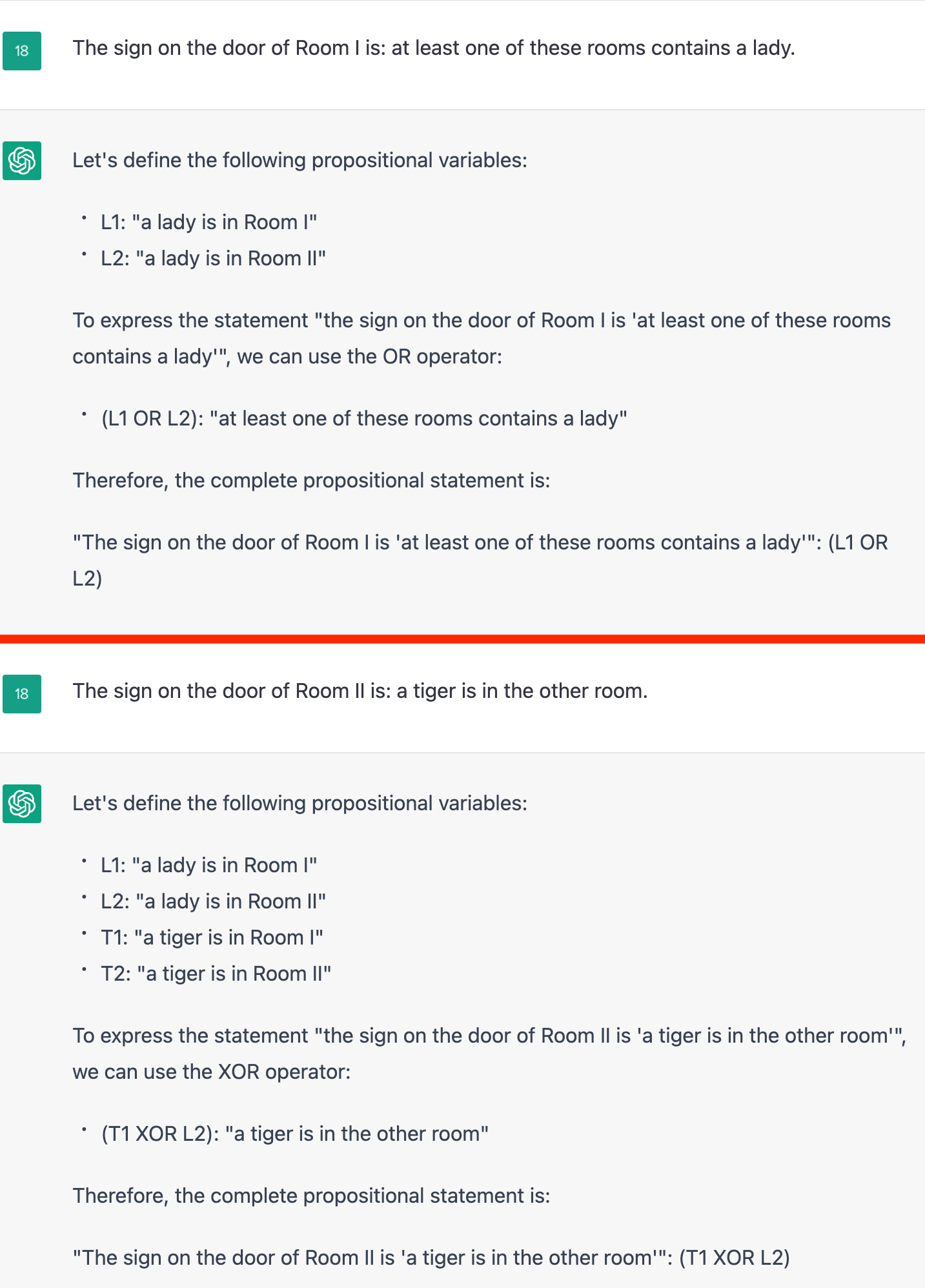}

\subsection*{Trial III}
Direct query to ChatGPT:\\[1ex]
\includegraphics[width=\columnwidth]{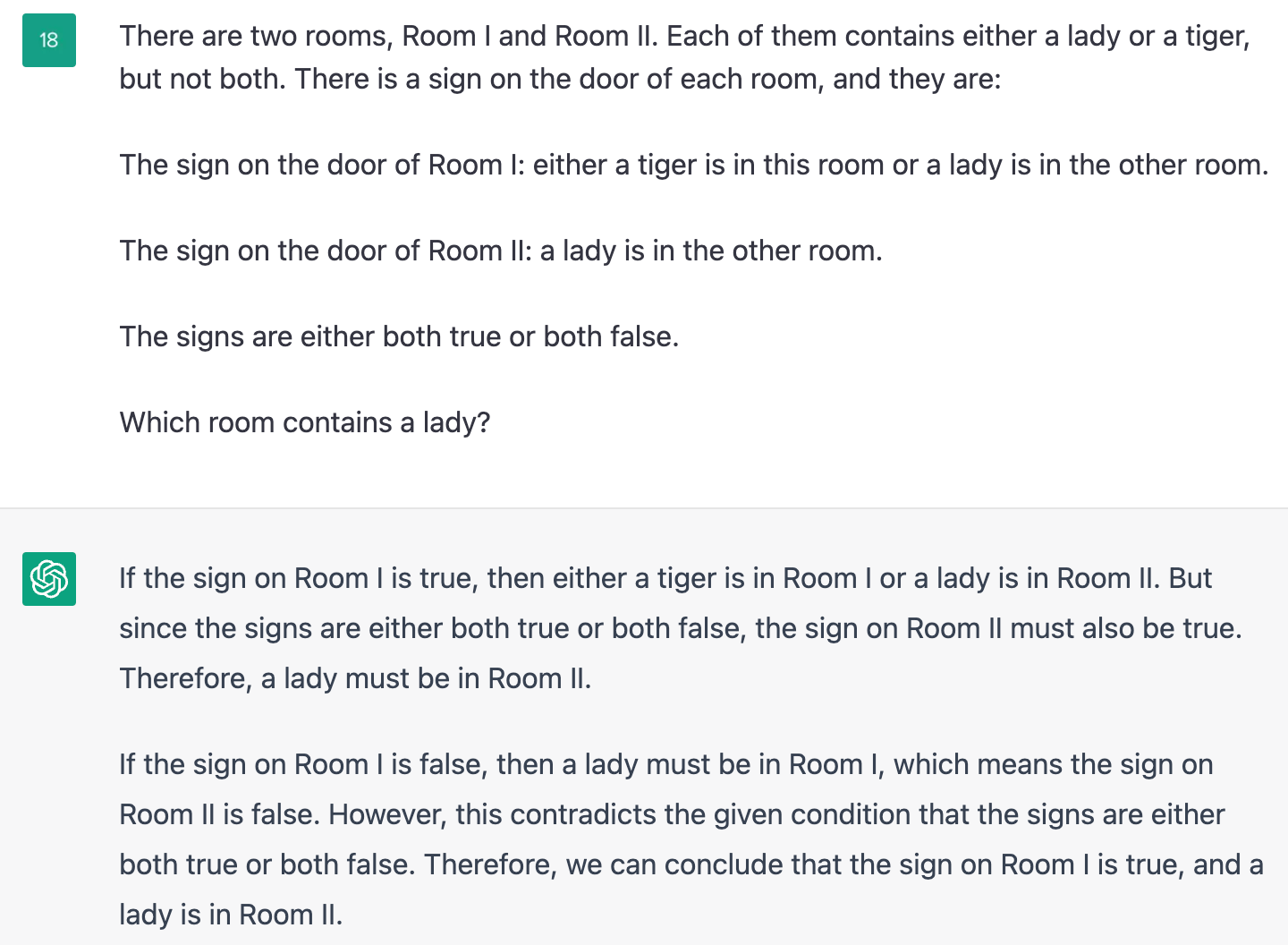}

\newpage
\noindent
Asking it to translate the text to logic:\\
\includegraphics[height=.97\textheight]{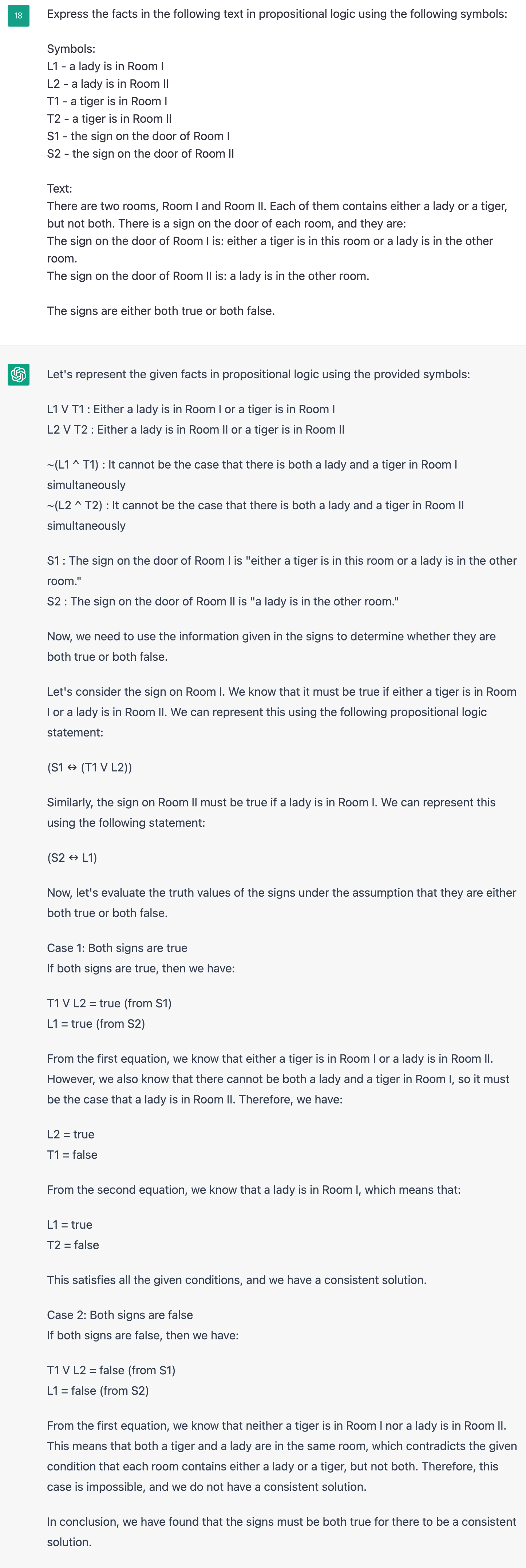}

\newpage
\noindent
Asking it to translate the statements in the text one at a time:\\
\includegraphics[width=.95\columnwidth]{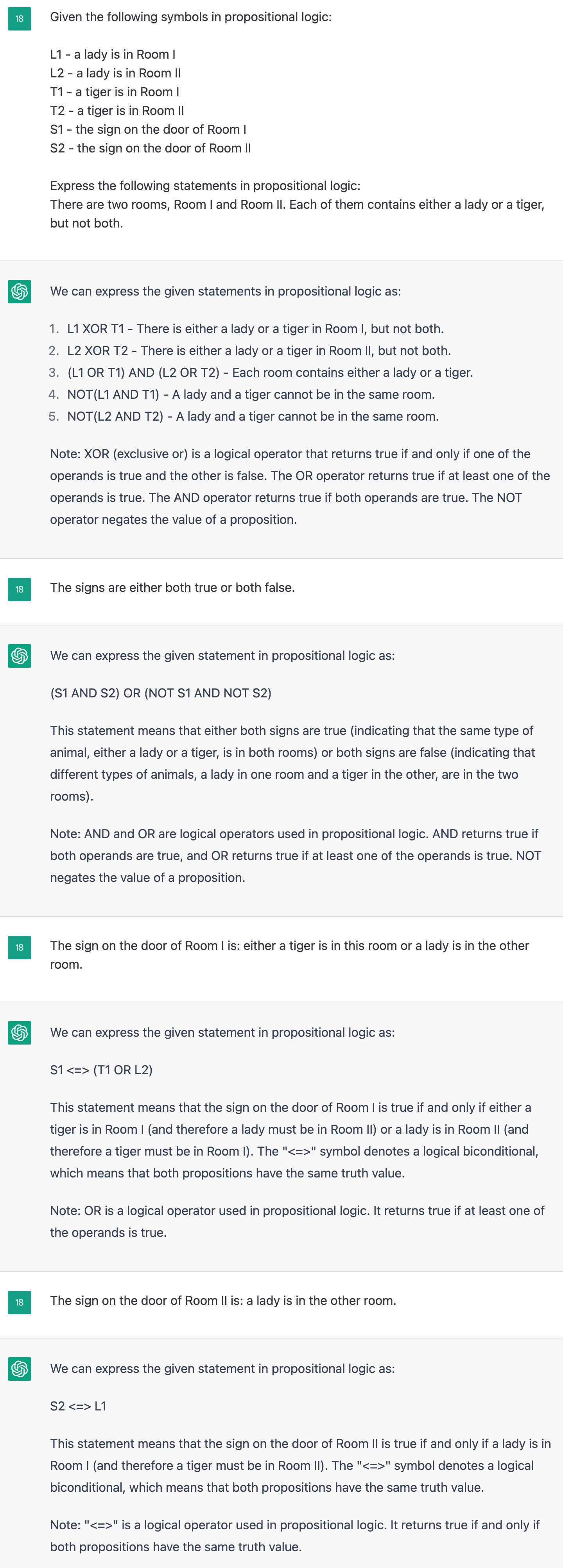}

\subsection*{Trial IV}
Direct query to ChatGPT:\\[1ex]
\includegraphics[width=\columnwidth]{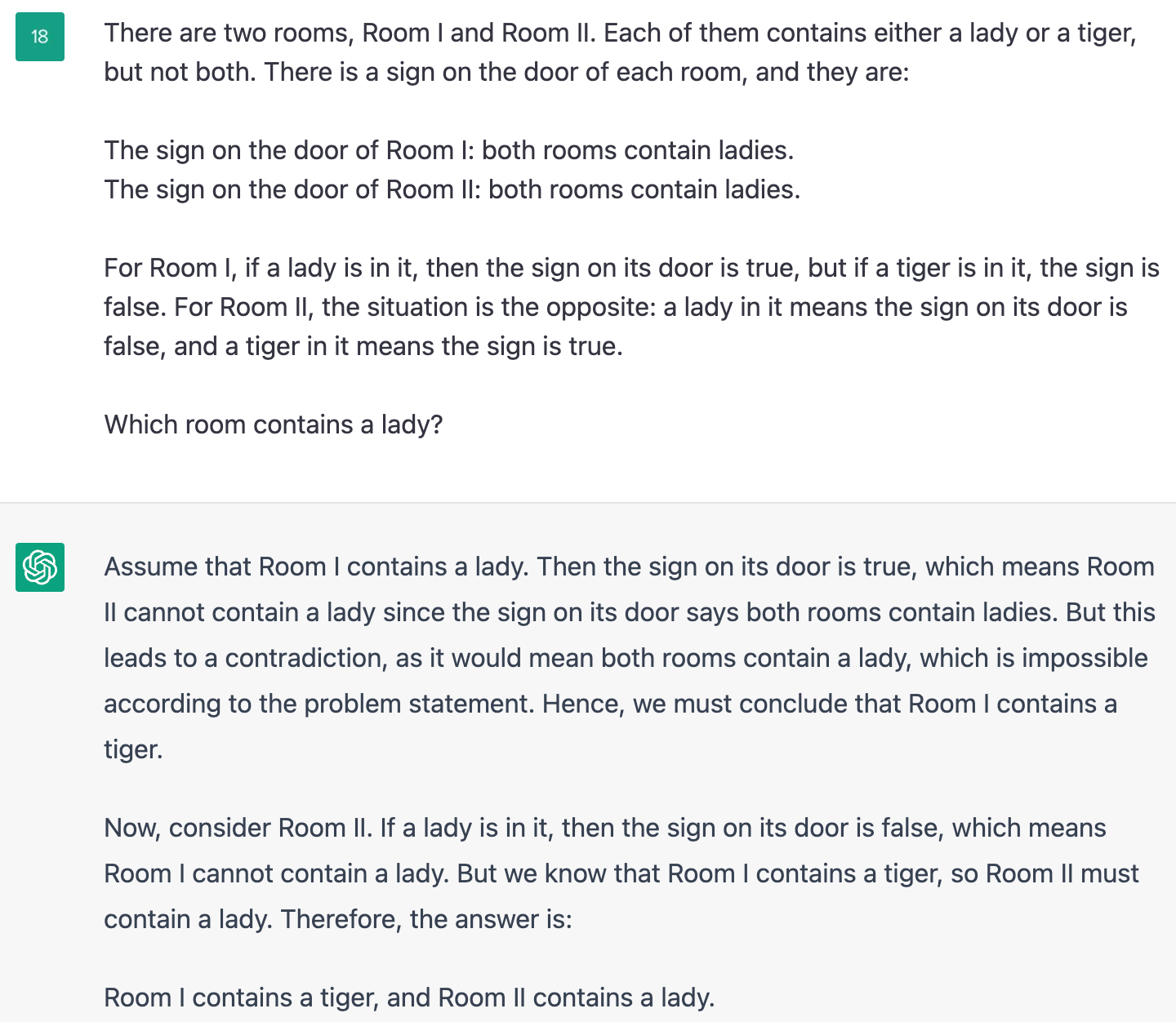}
Asking it to translate the text to logic:\\[1ex]
\includegraphics[width=\columnwidth]{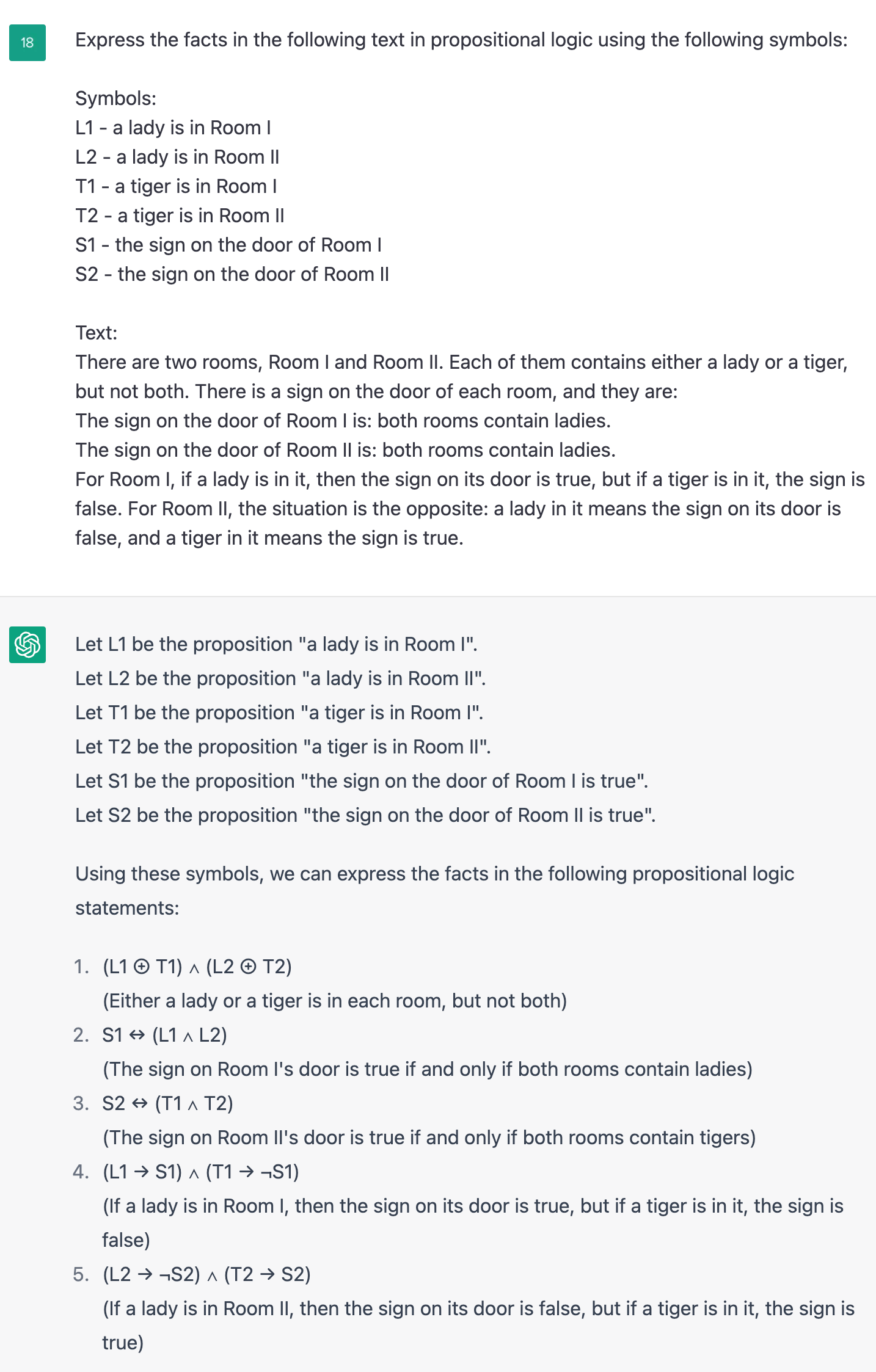}

\newpage
\noindent
Asking it to translate the statements in the text one at a time:
\includegraphics[width=\columnwidth]{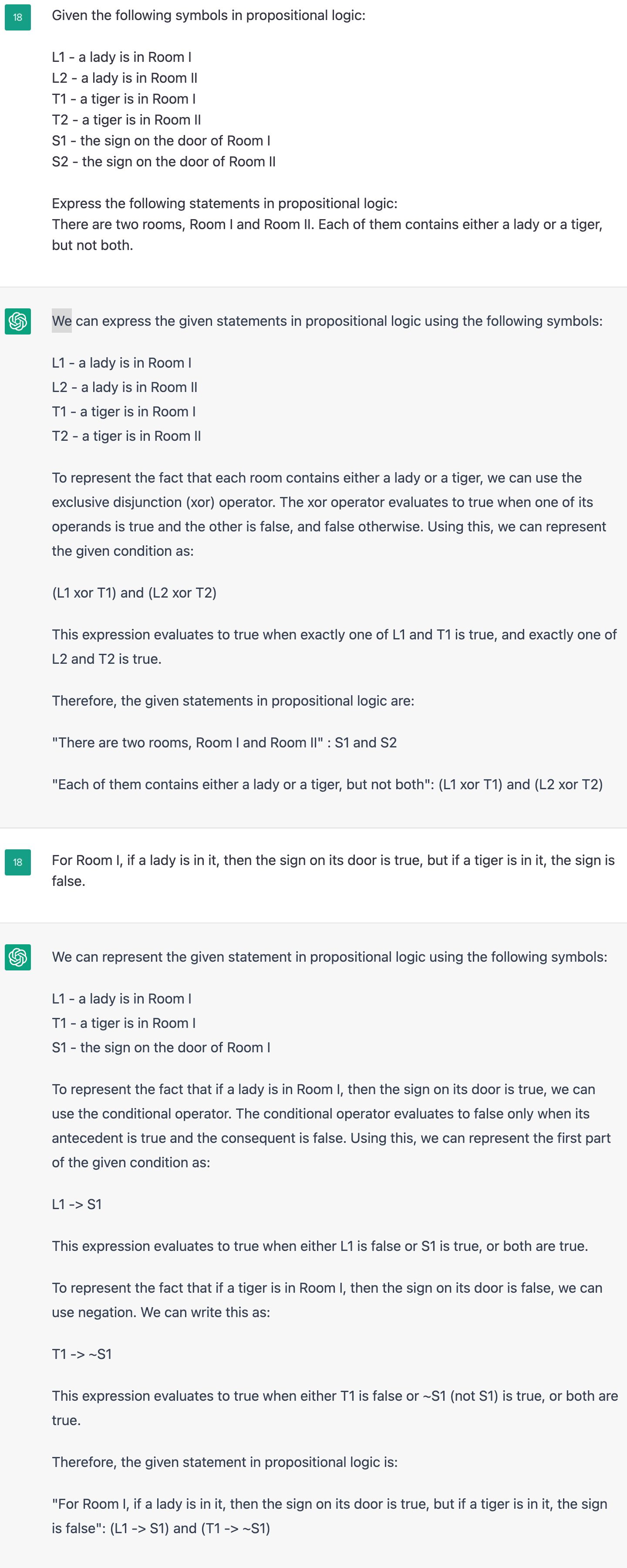}
\includegraphics[width=\columnwidth]{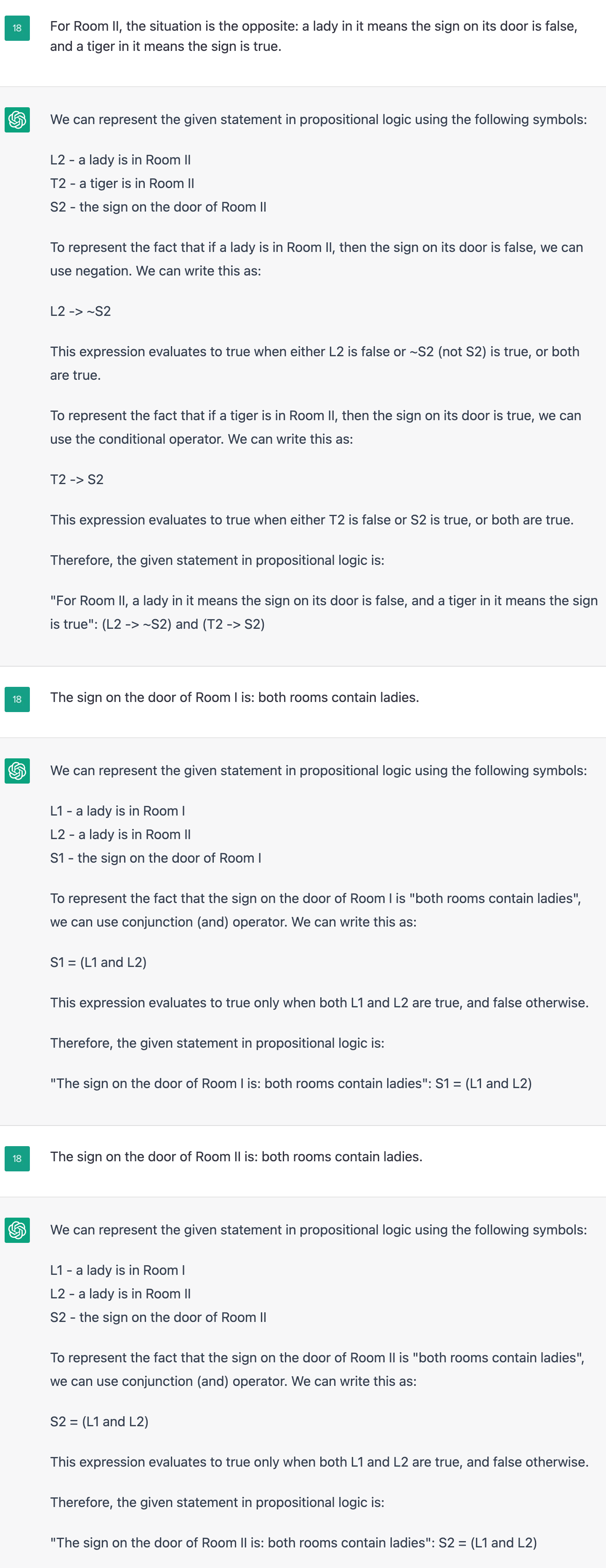}
\newpage
\subsection*{Trial V}
Direct query to ChatGPT:\\[1ex]
\includegraphics[width=\columnwidth]{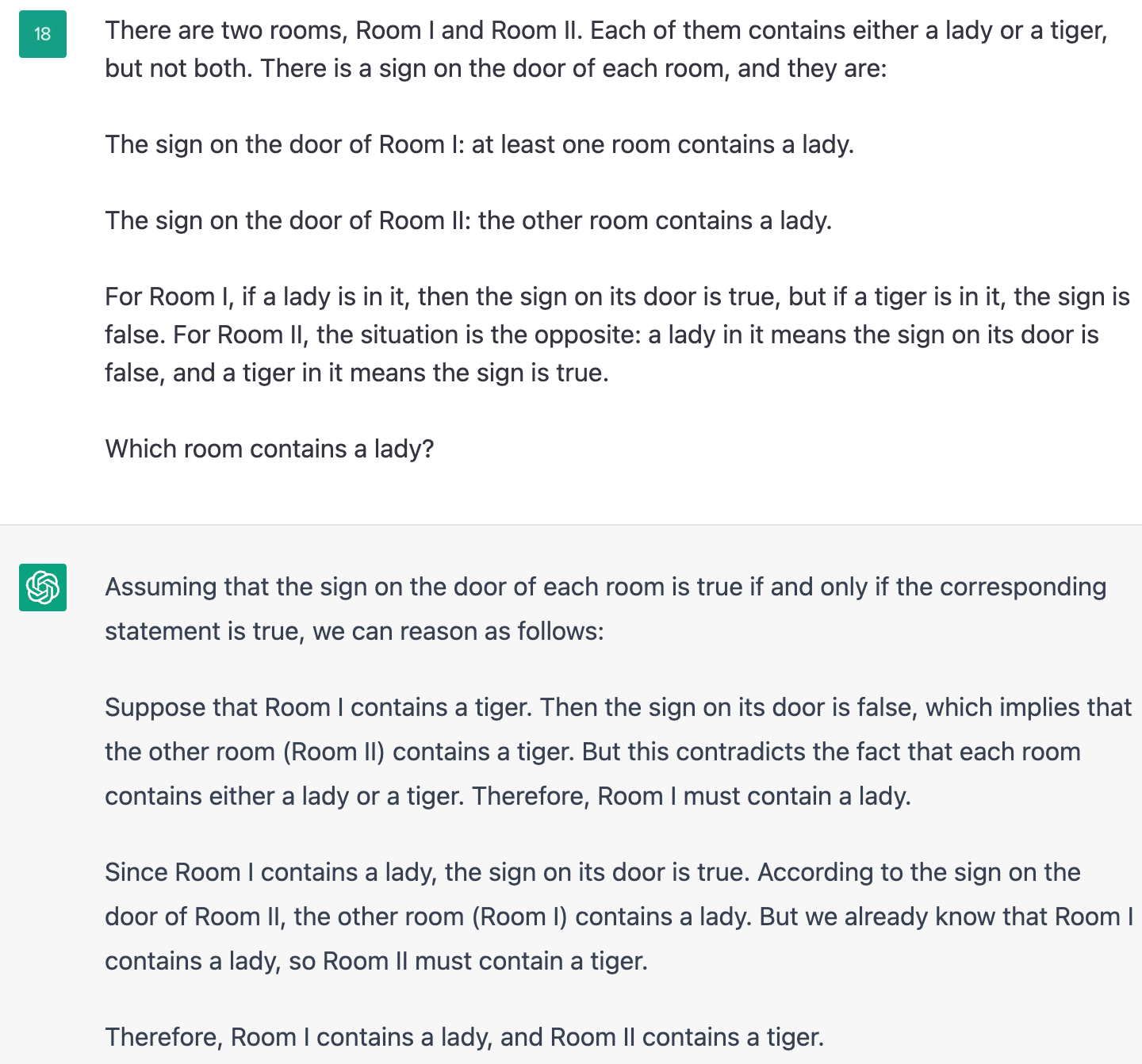}
\\
Asking it to translate the text to logic:\\[1ex]
\includegraphics[width=\columnwidth]{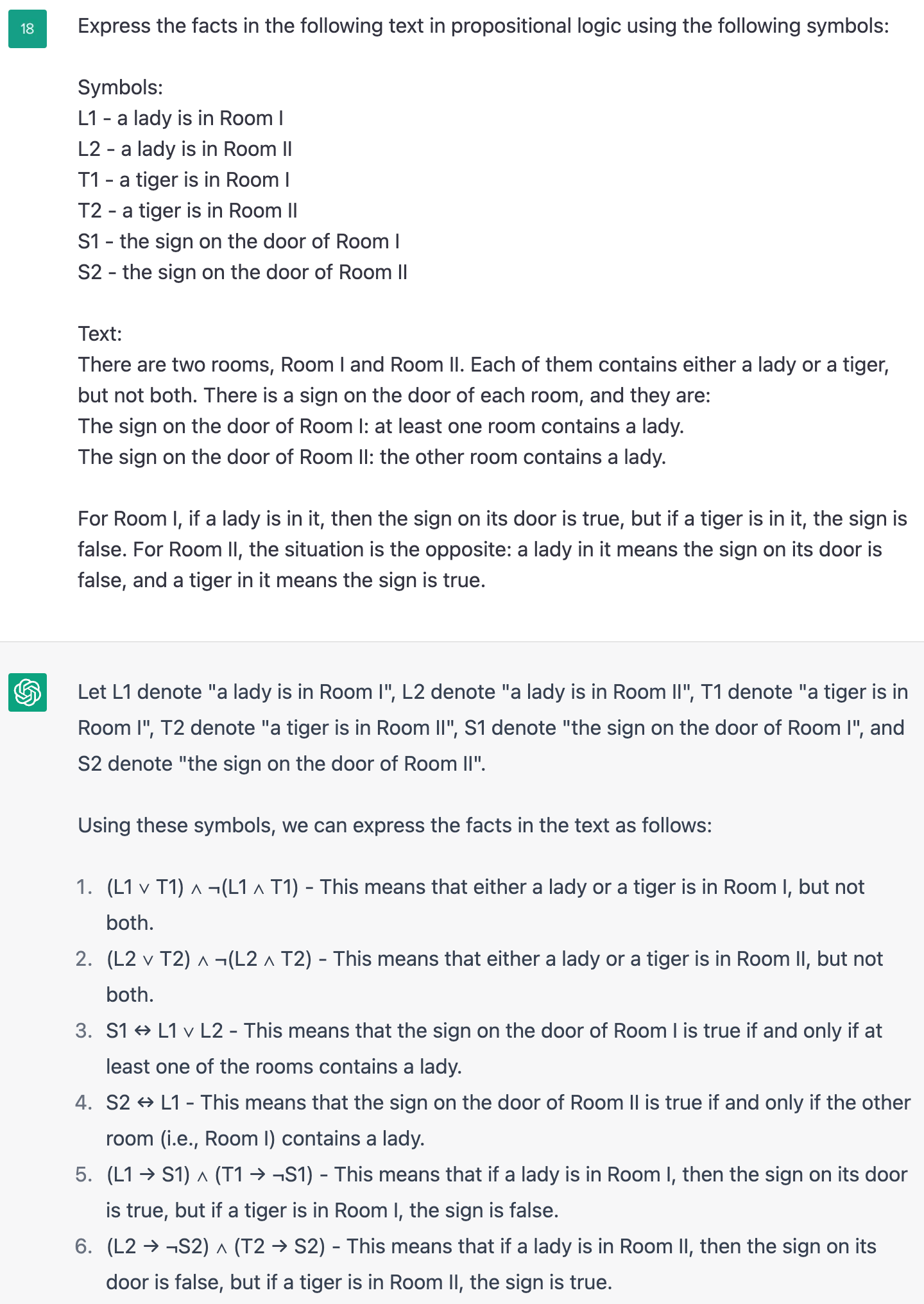}
\newpage
\noindent
Asking it to translate the statements in the text one at a time:
\\
\includegraphics[width=\columnwidth]{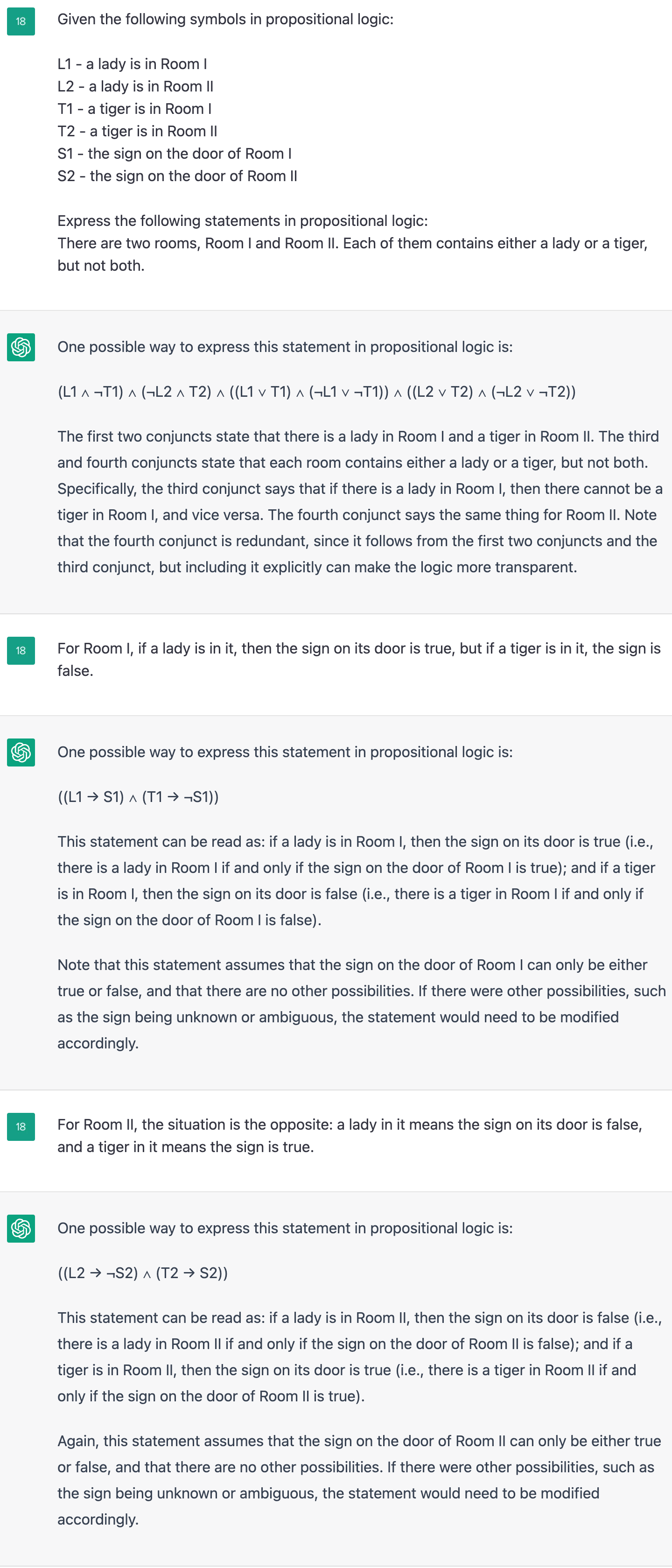}
\newpage
\noindent
Asking it to translate the statements in the text one at a time: (continued)
\\
\includegraphics[width=\columnwidth]{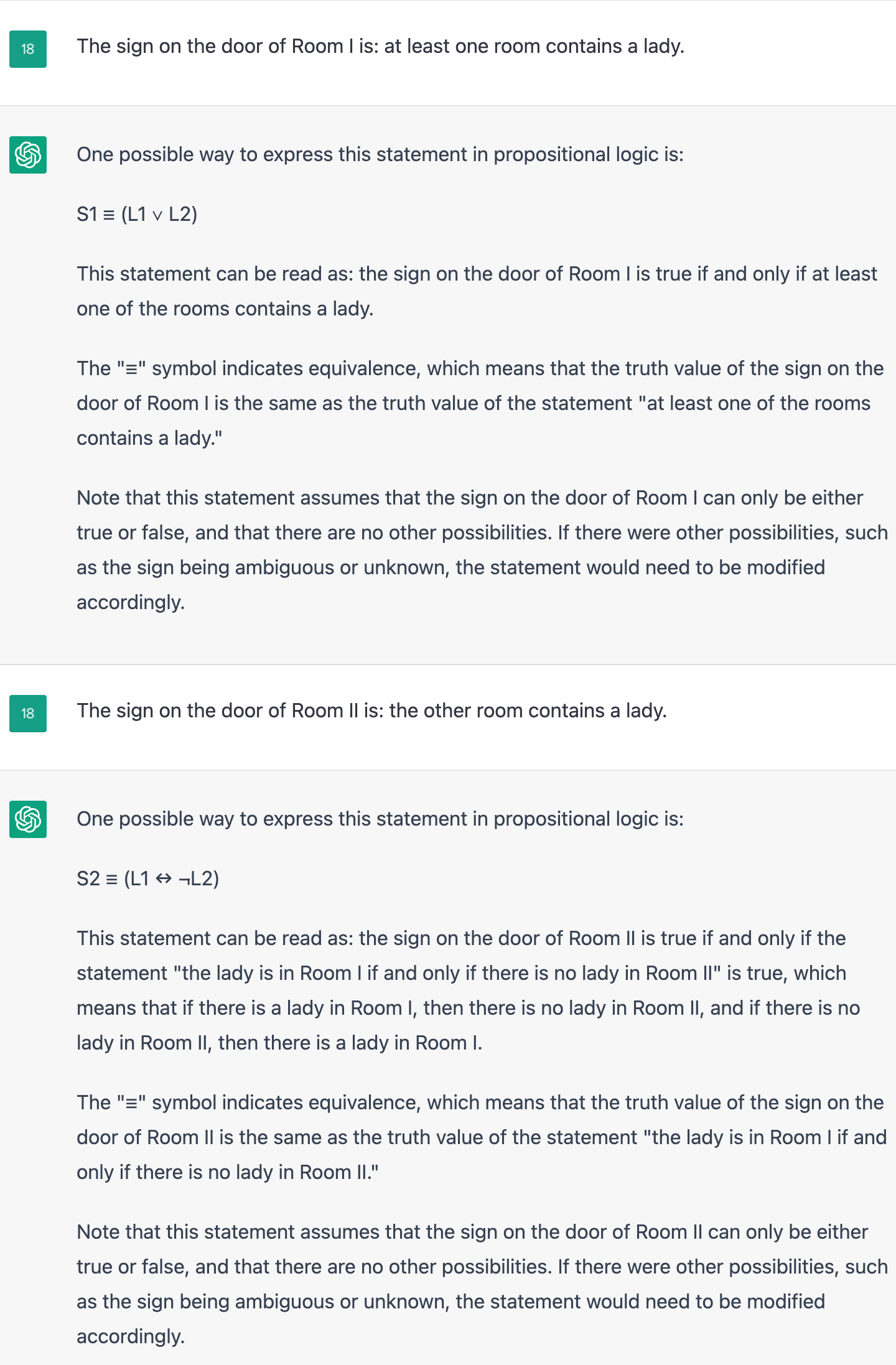}

\subsection*{Trial VI}
Direct query to ChatGPT:\\[1ex]
\includegraphics[width=\columnwidth]{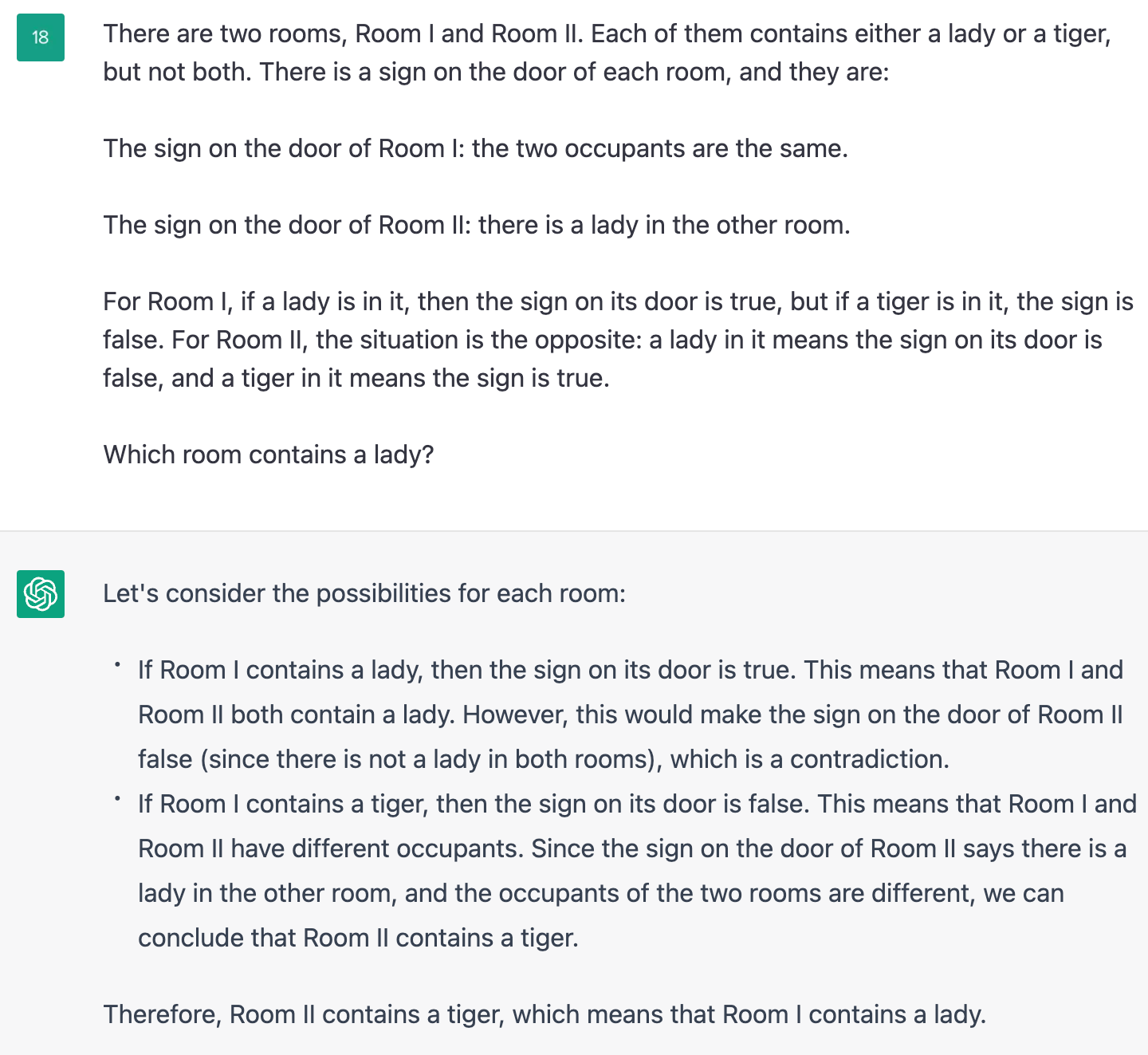}
\\
Asking it to translate the text to logic:\\
\includegraphics[width=\columnwidth]{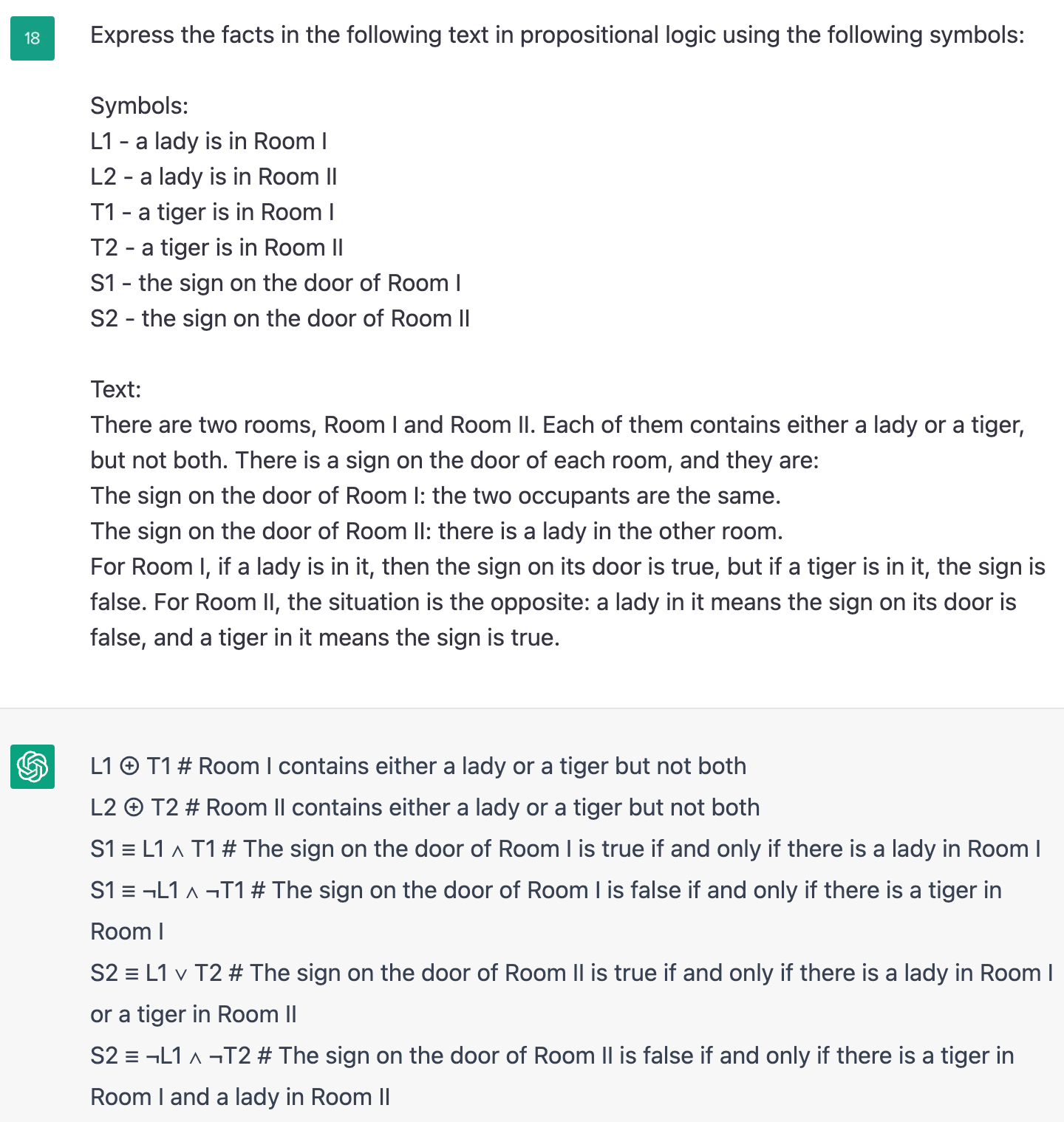}
\\
Asking it to translate the statements in the text one at a time:\\
\includegraphics[width=\columnwidth]{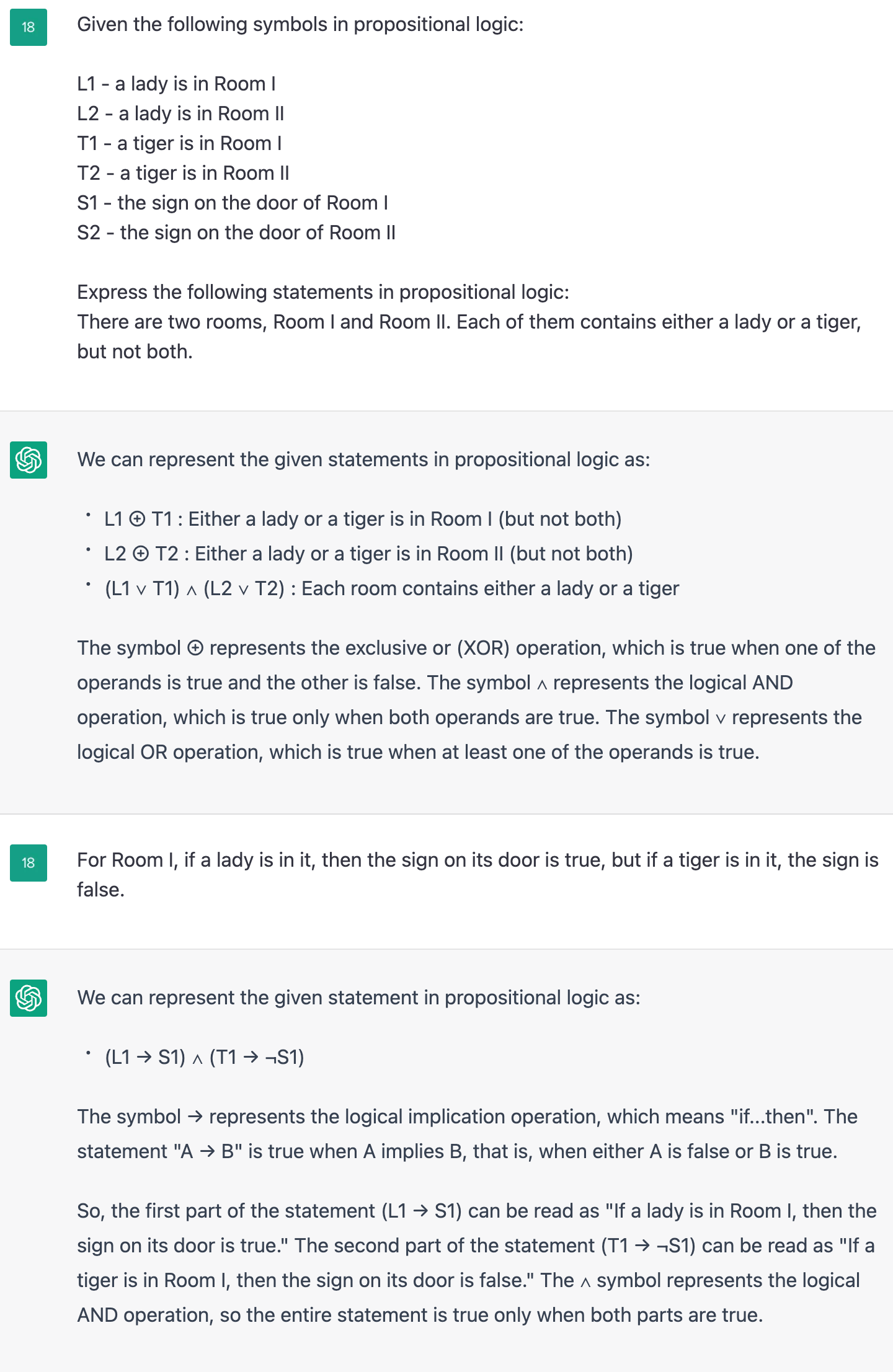}
Asking it to translate the statements in the text one at a time: (continued)\\
\includegraphics[width=\columnwidth]{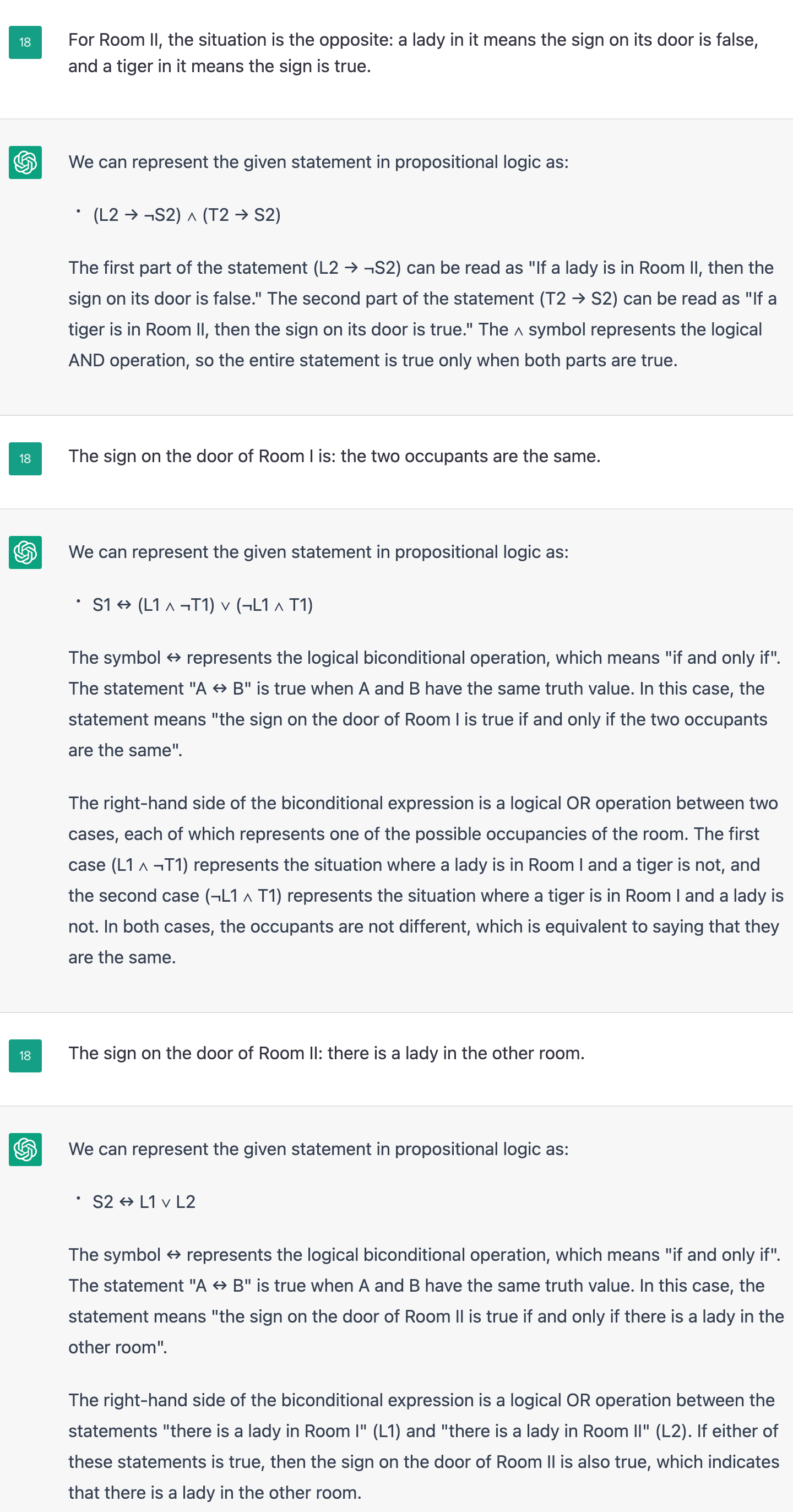}

\newpage
\subsection*{Trial VII}
Direct query to ChatGPT:\\[1ex]
\includegraphics[width=\columnwidth]{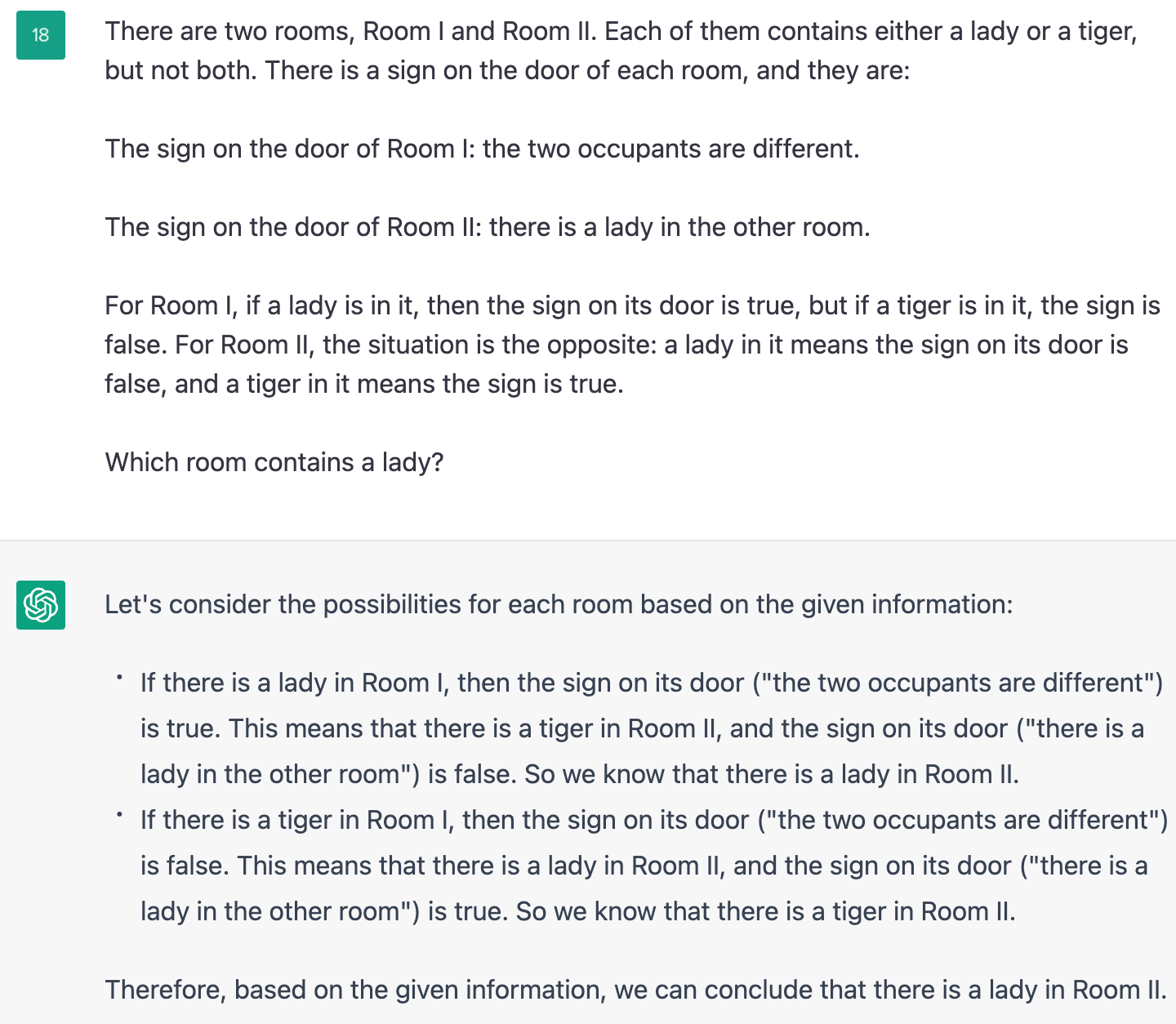}
Asking it to translate the text to logic:\\[1ex]
\includegraphics[width=\columnwidth]{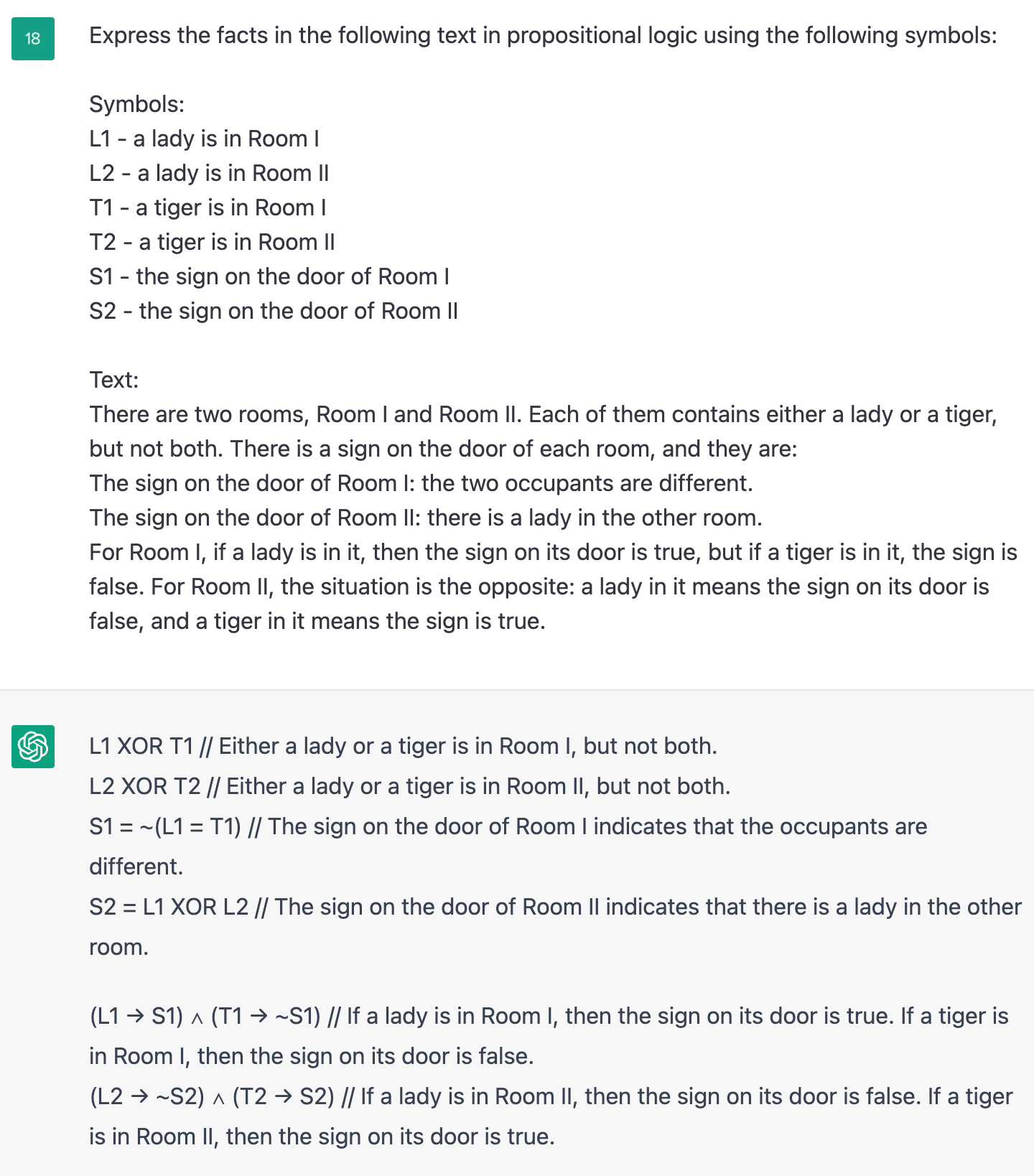}
\\
Asking it to translate the statements in the text one at a time:\\
\newpage
\begin{center}
    \includegraphics[width=.95\columnwidth]{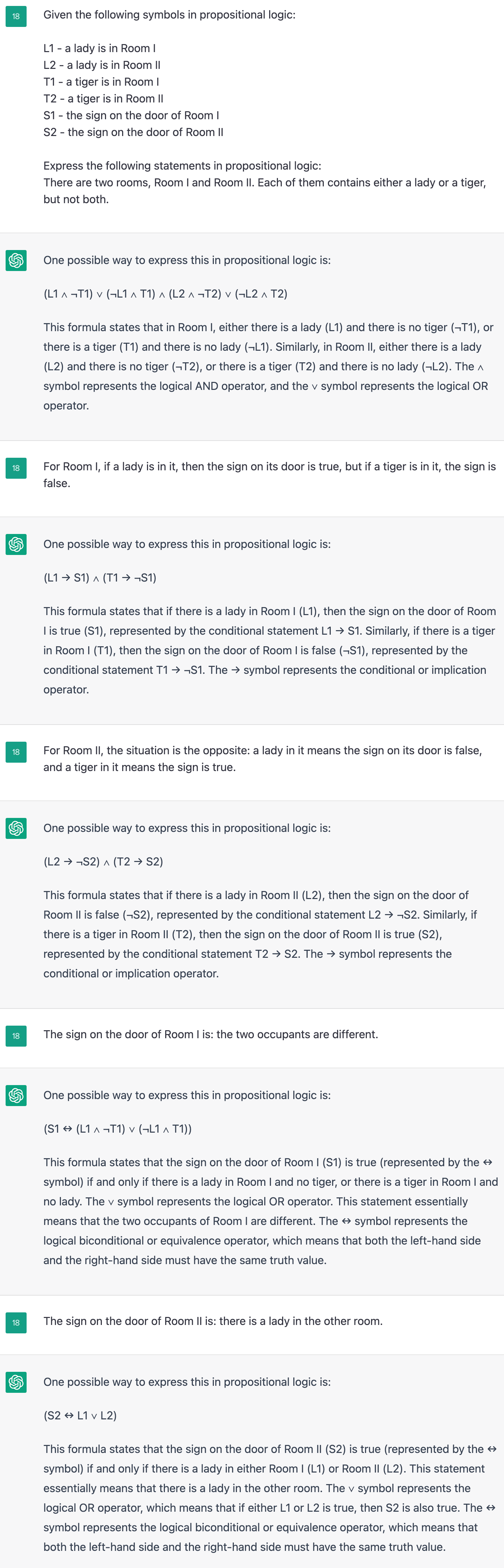}
\end{center}

\subsection*{Trial VIII}
Direct query to ChatGPT:\\[.5ex]
\includegraphics[width=.86\columnwidth]{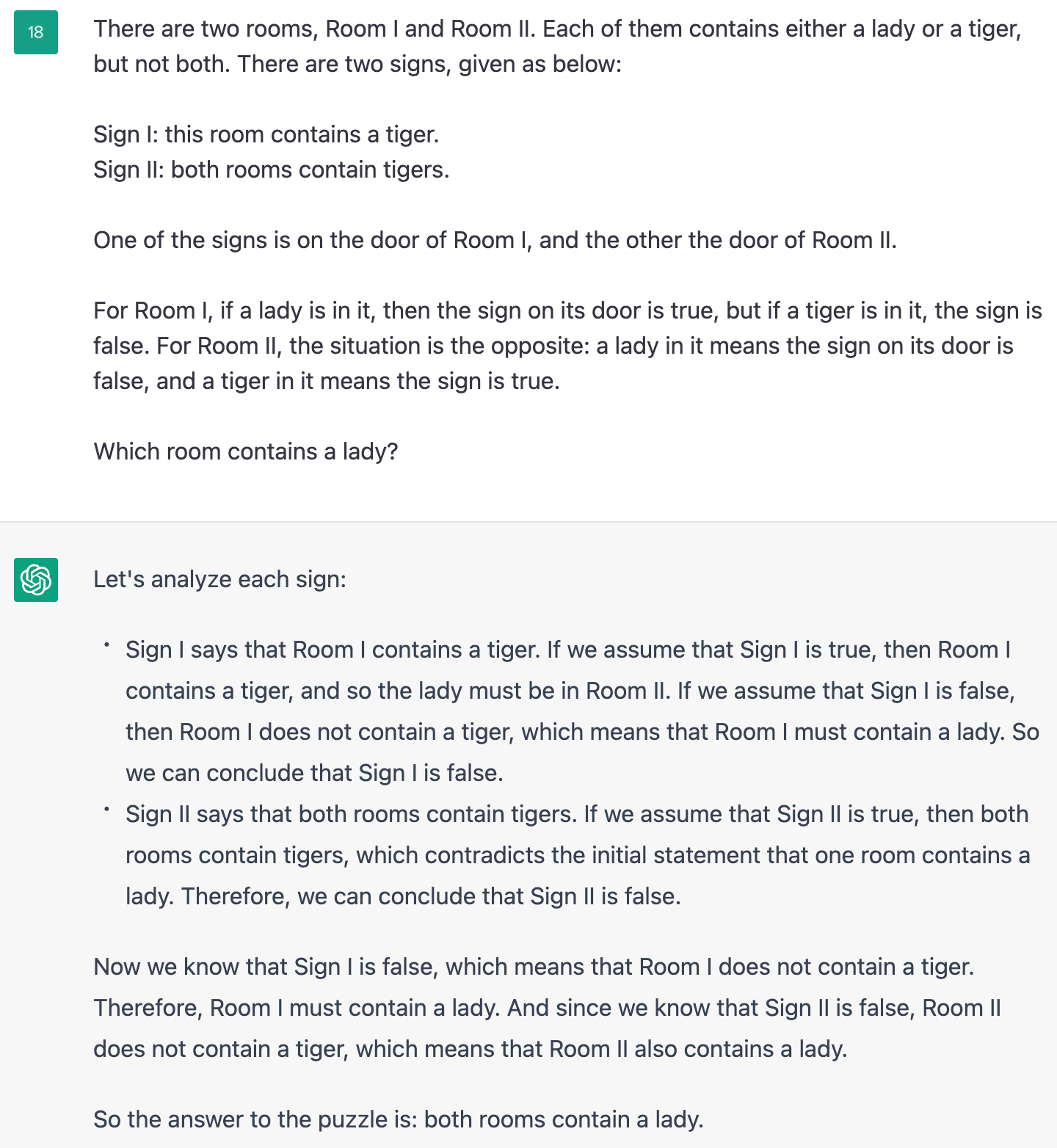}
\\
Asking it to translate the text to logic:\\[.5ex]
\includegraphics[width=.86\columnwidth]{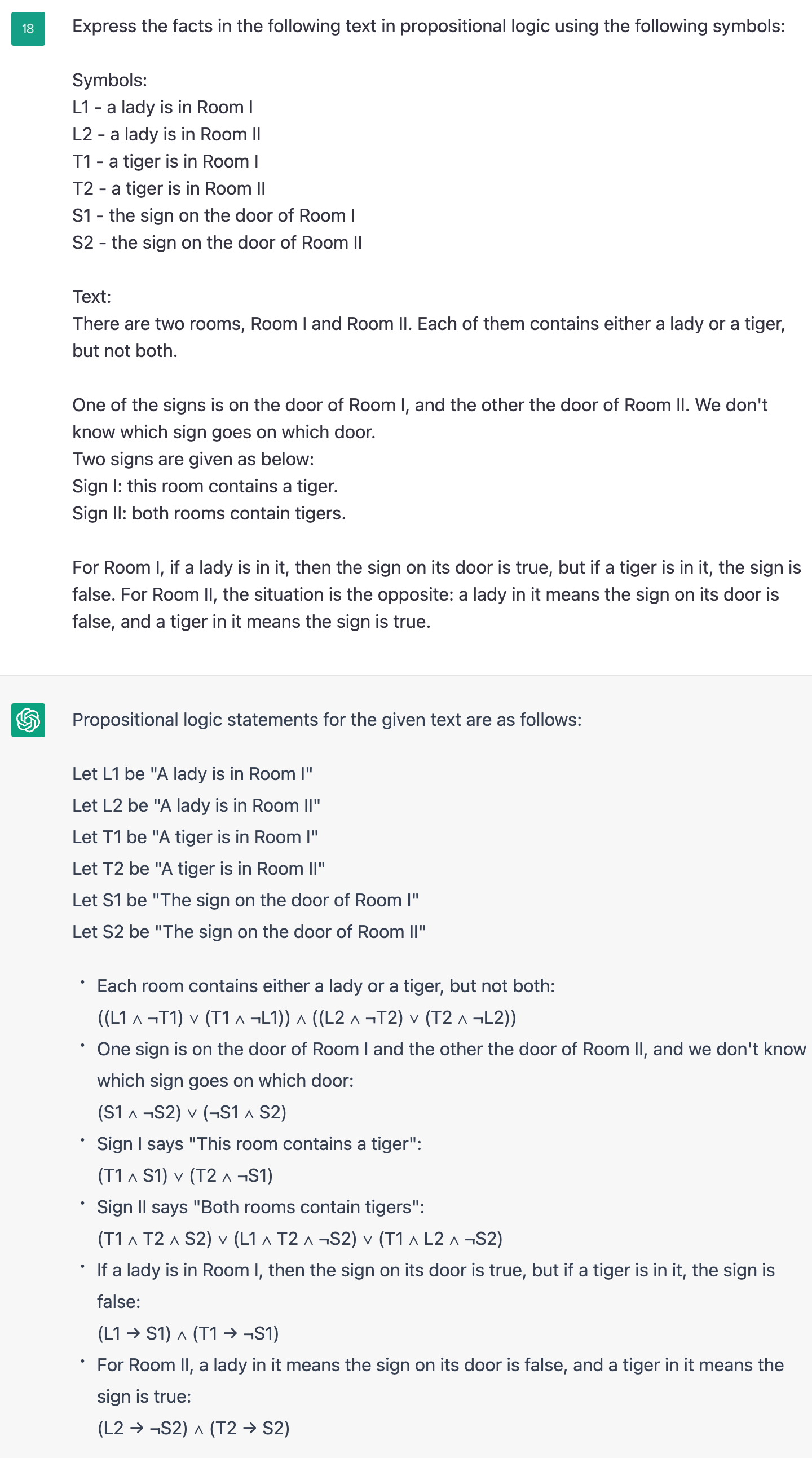}
\\
Asking it to translate the statements in the text one at a time:
\newpage
\begin{center}
\includegraphics[width=.92\columnwidth]{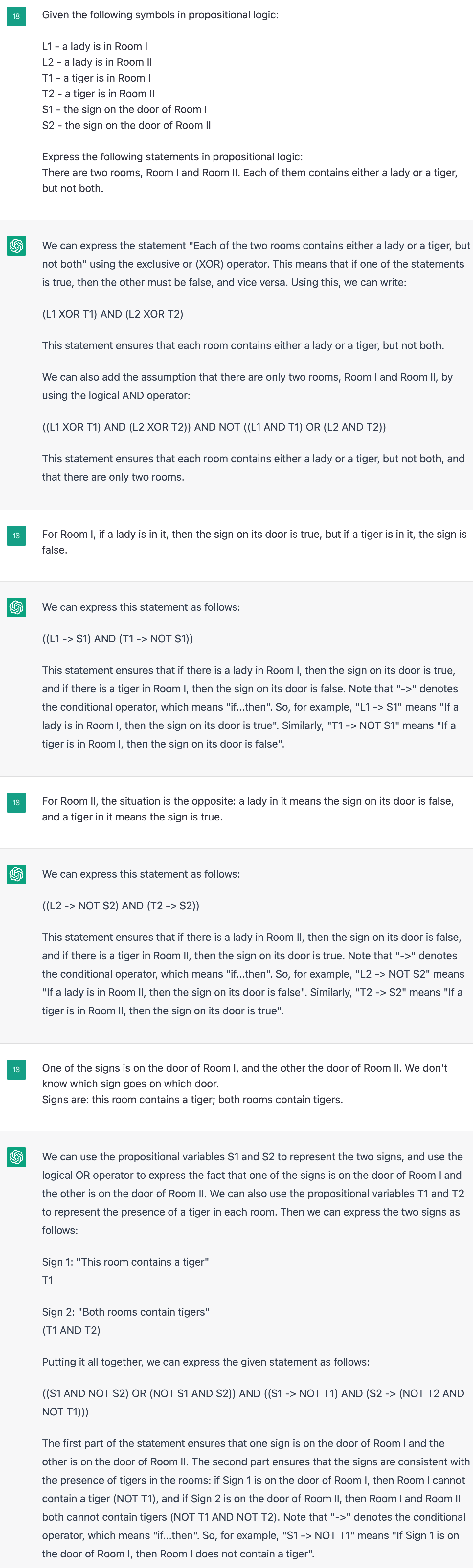}
\end{center}

\subsection*{Trial IX}
Direct query to ChatGPT:\\[1ex]
\includegraphics[width=\columnwidth]{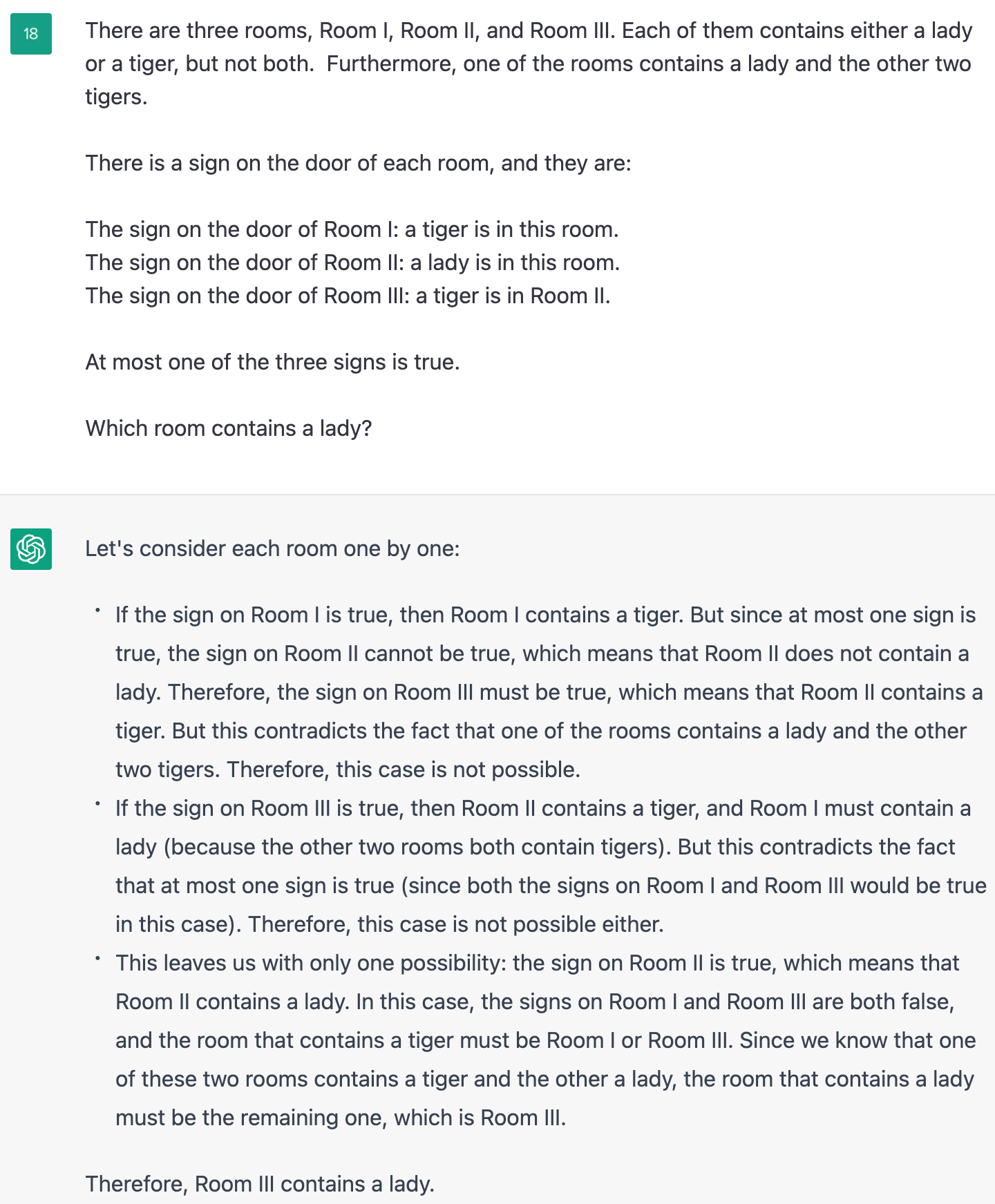}
Asking it to translate the text to logic:\\[1ex]
\includegraphics[width=\columnwidth]{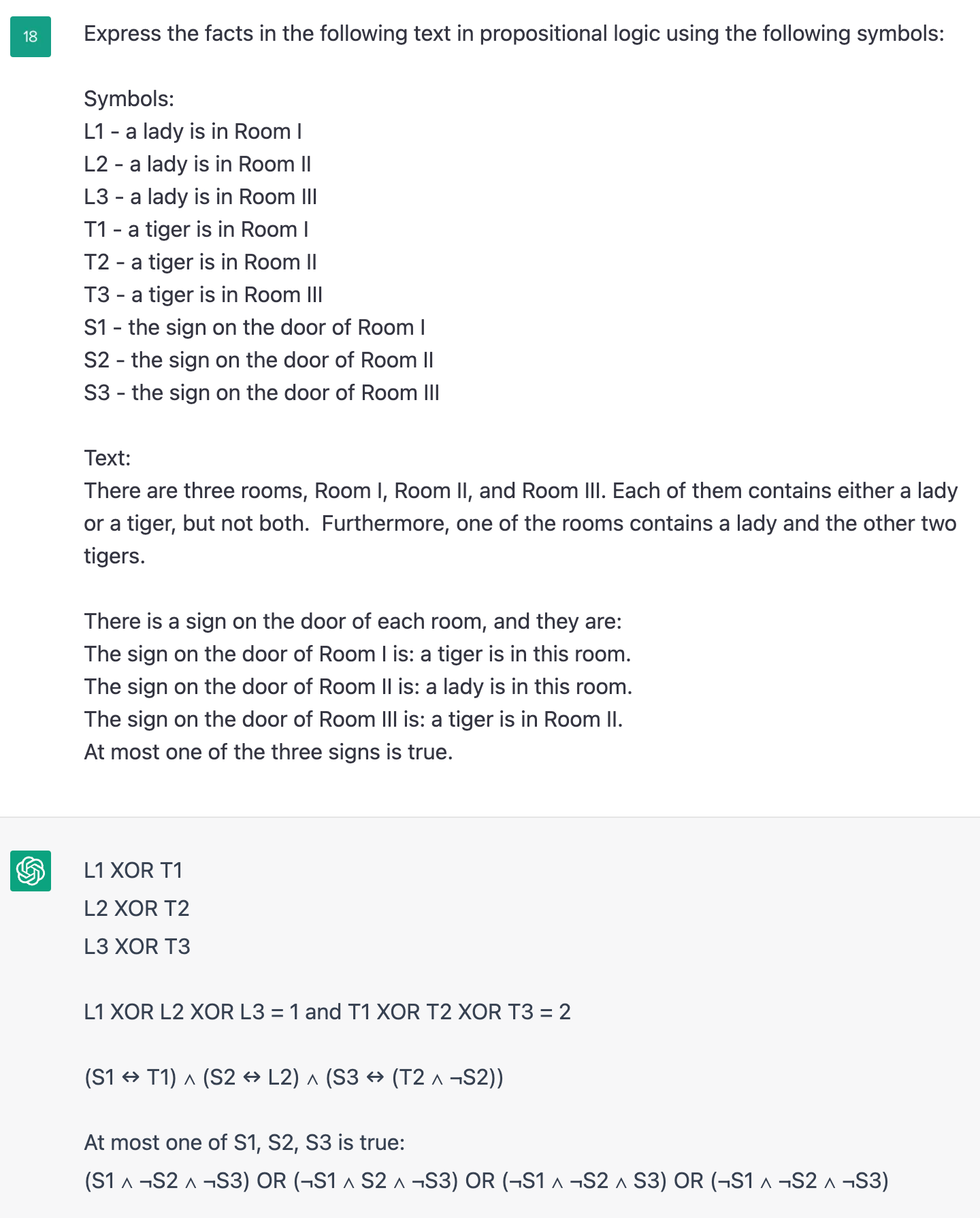}\\
Asking it to translate the statements in the text one at a time:
\newpage
\begin{center}
\includegraphics[width=.98\columnwidth]{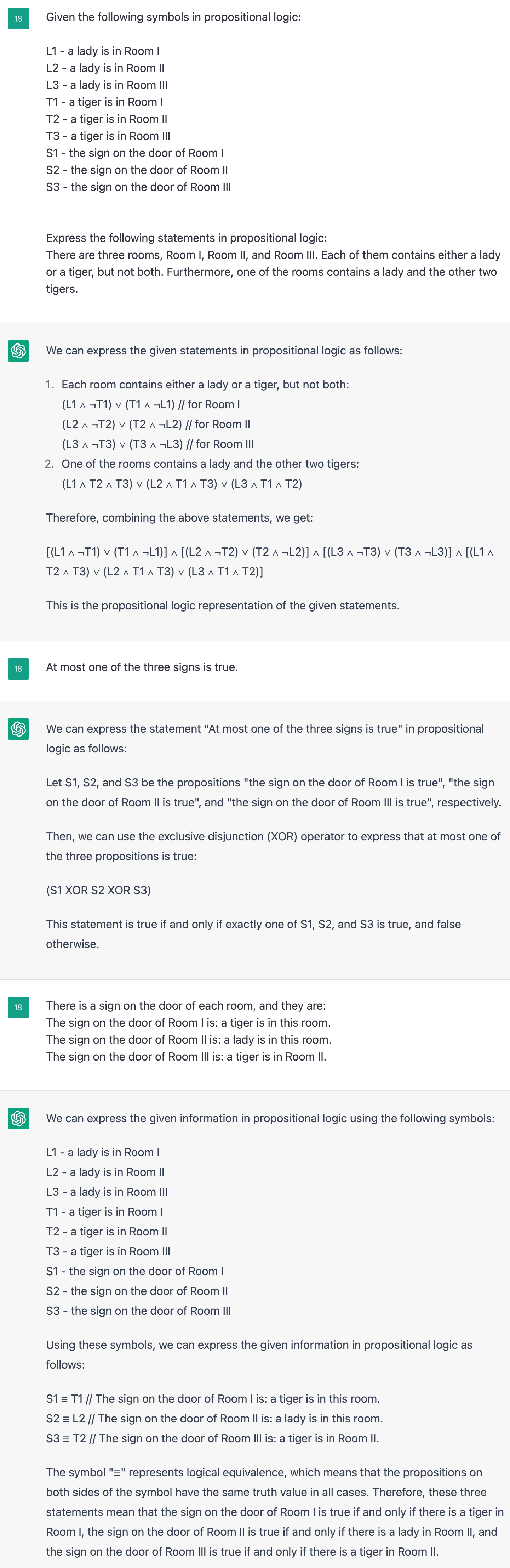}
\end{center}

\subsection*{Trial X}
Direct query to ChatGPT:\\
\includegraphics[width=.84\columnwidth]{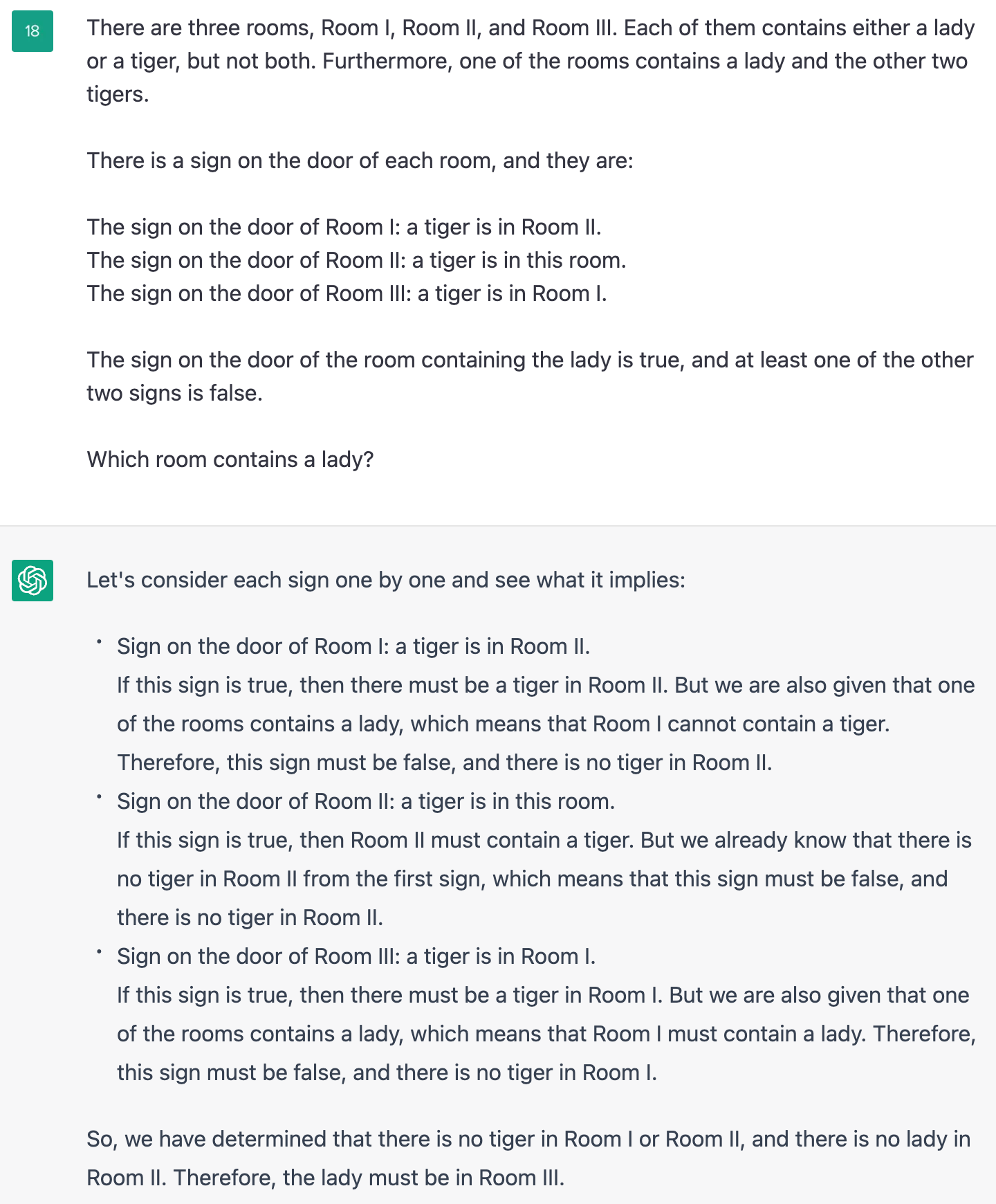}
\\
Asking it to translate the text to logic:\\
\includegraphics[width=.85\columnwidth]{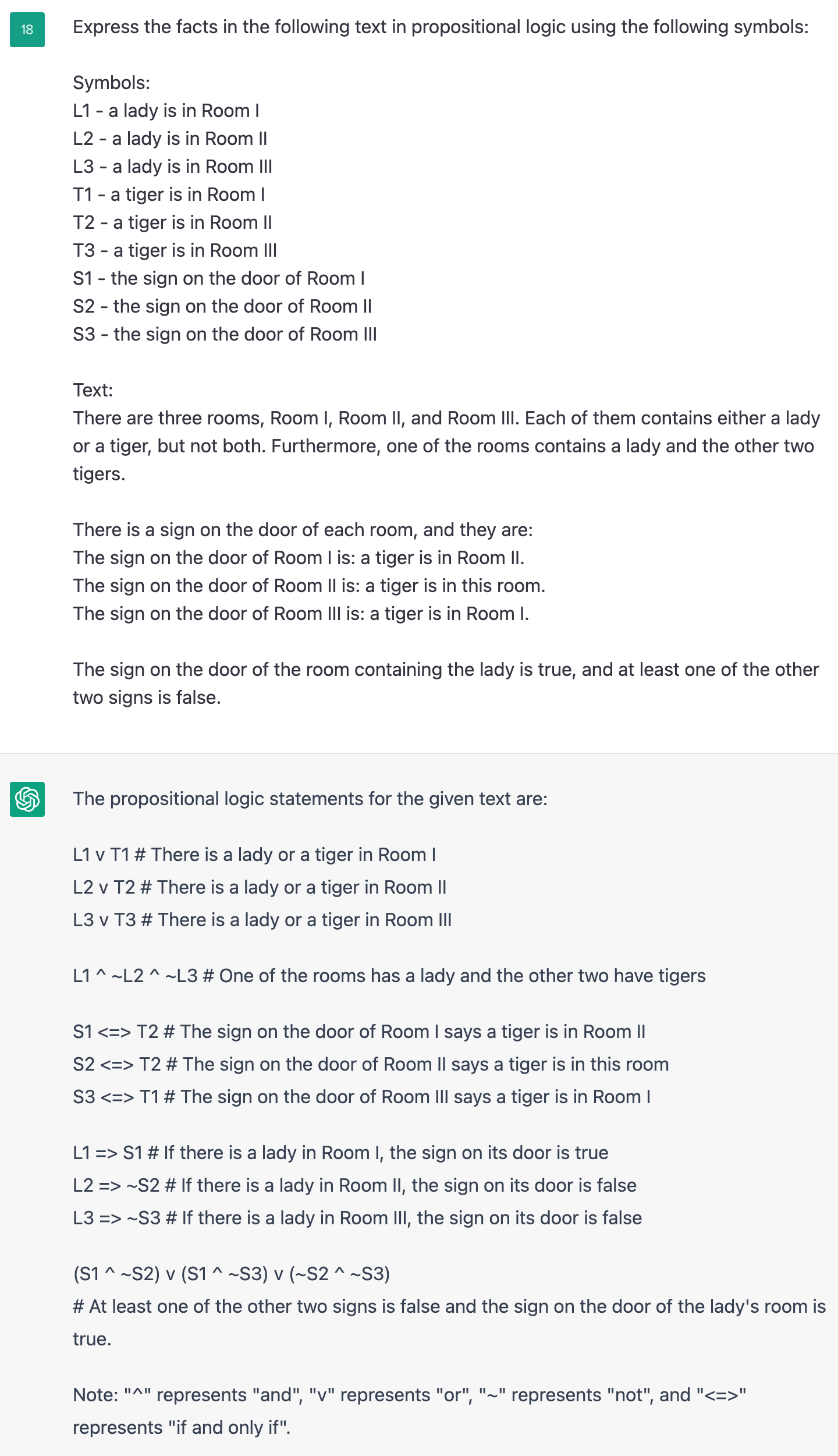}
\\
Asking it to translate the statements in the text one at a time:
\newpage
\begin{center}
    \includegraphics[width=.78\columnwidth]{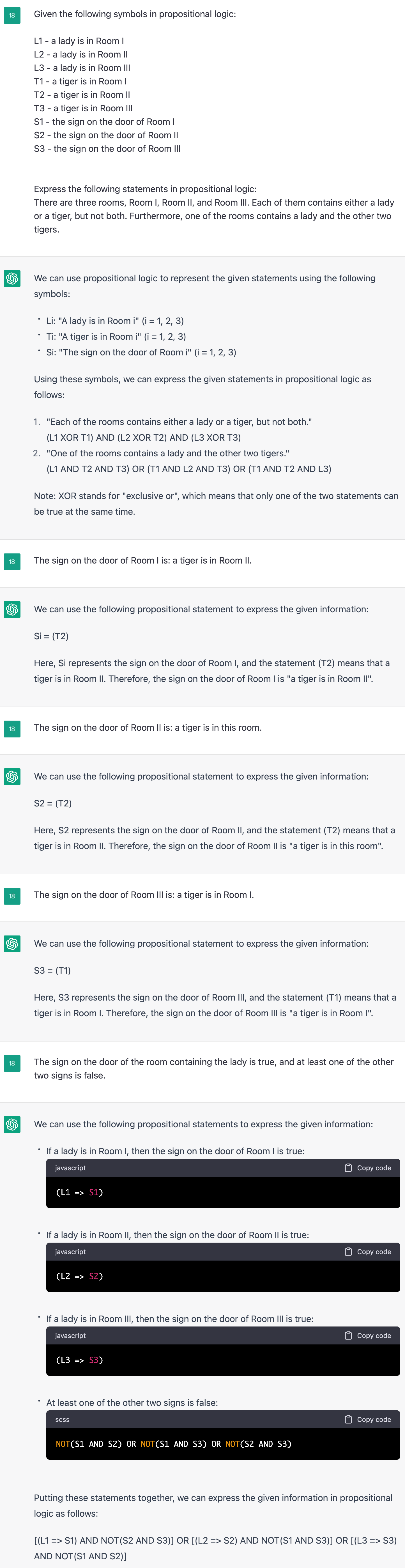}
\end{center}

\subsection*{Trial XI}
Direct query to ChatGPT:\\[1ex]
\includegraphics[width=.8\columnwidth]{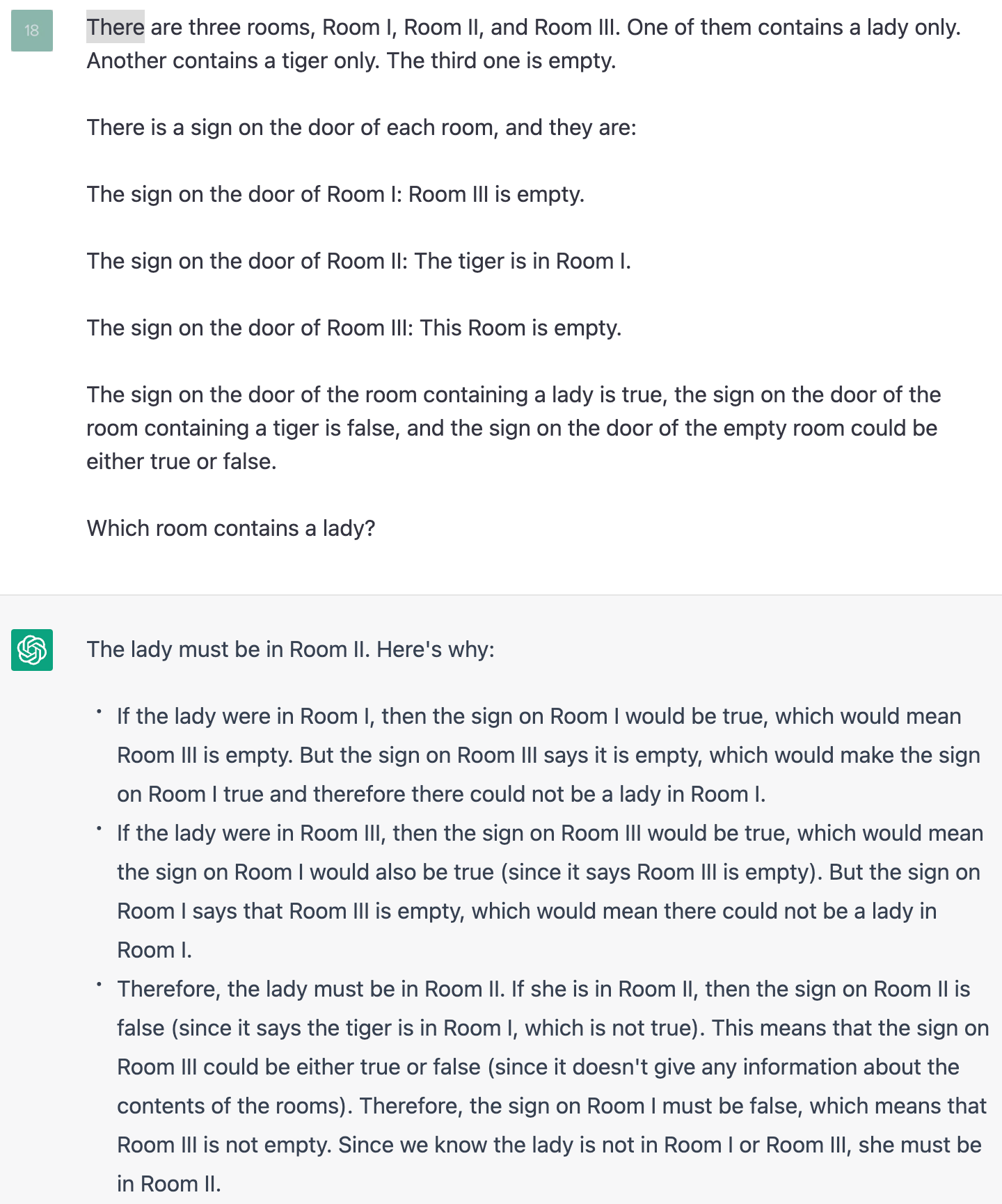}
\\
Asking it to translate the text to logic:\\[1ex]
\includegraphics[width=.8\columnwidth]{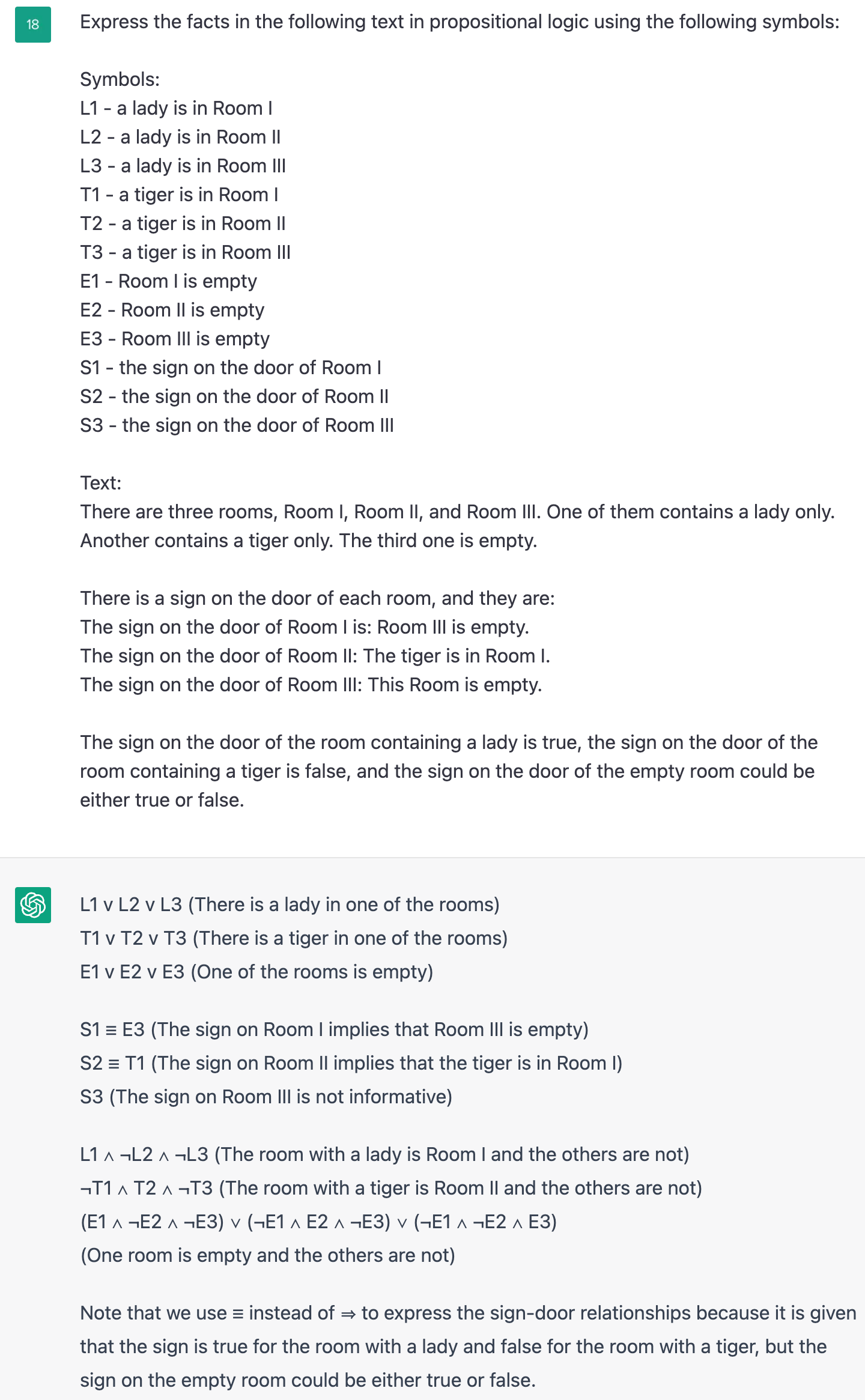}
\\
Asking it to translate the statements in the text one at a time:
\newpage
\includegraphics[width=.94\columnwidth]{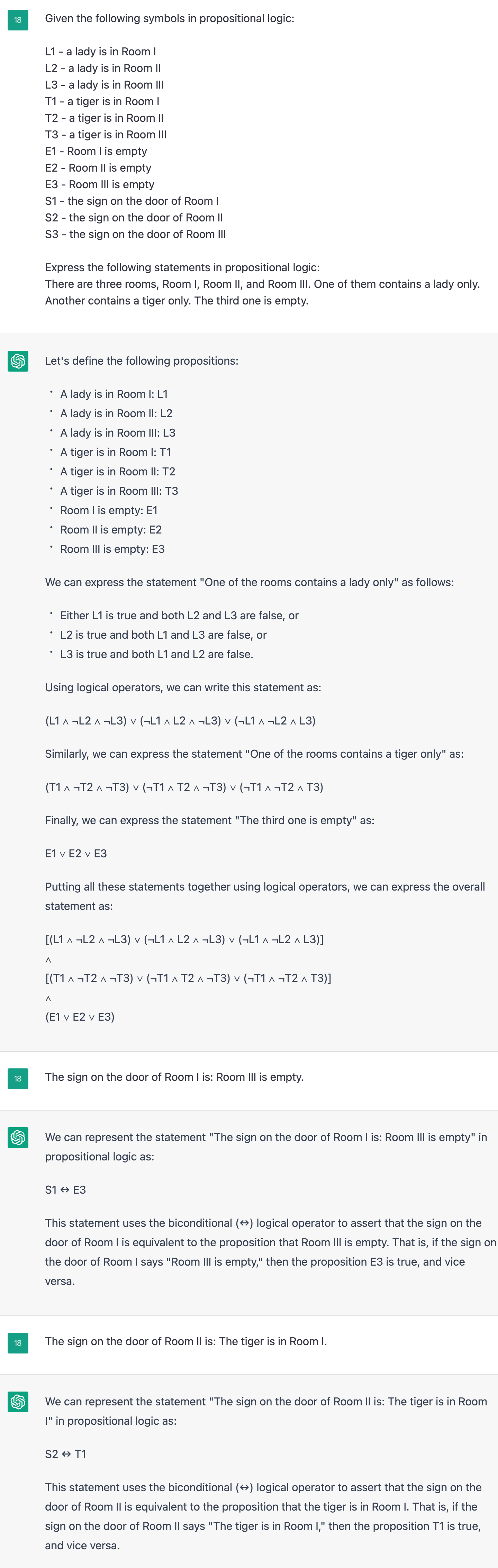}
\includegraphics[width=.94\columnwidth]{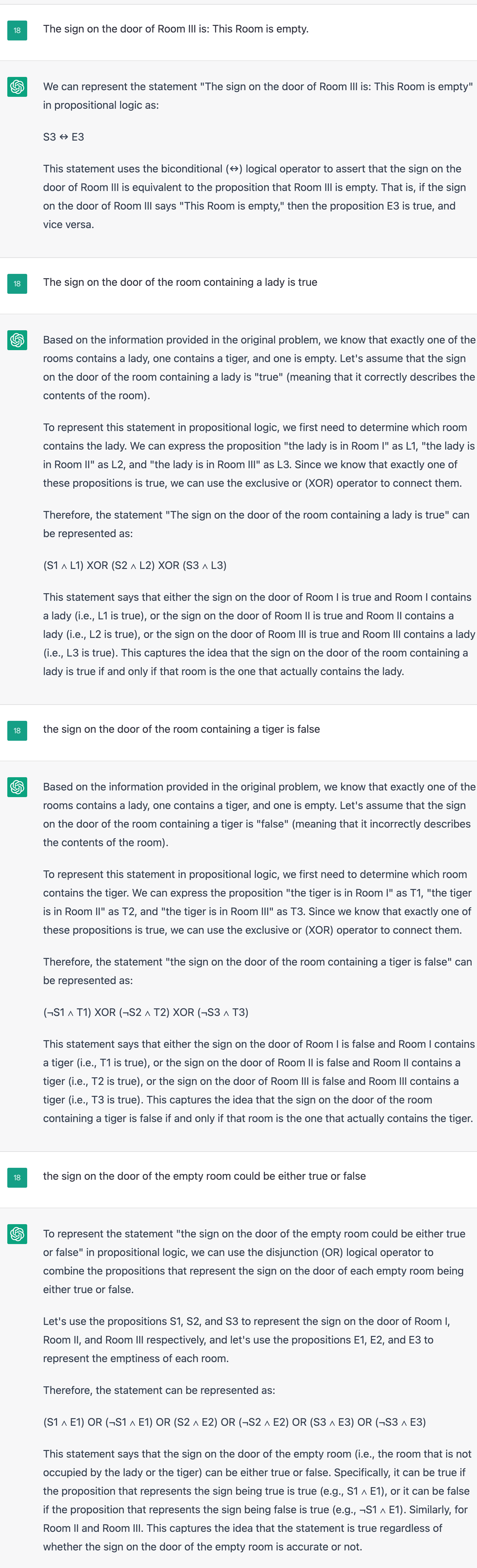}

\subsection*{Trial XII}
Direct query to ChatGPT:\\[1ex]
\includegraphics[width=\columnwidth]{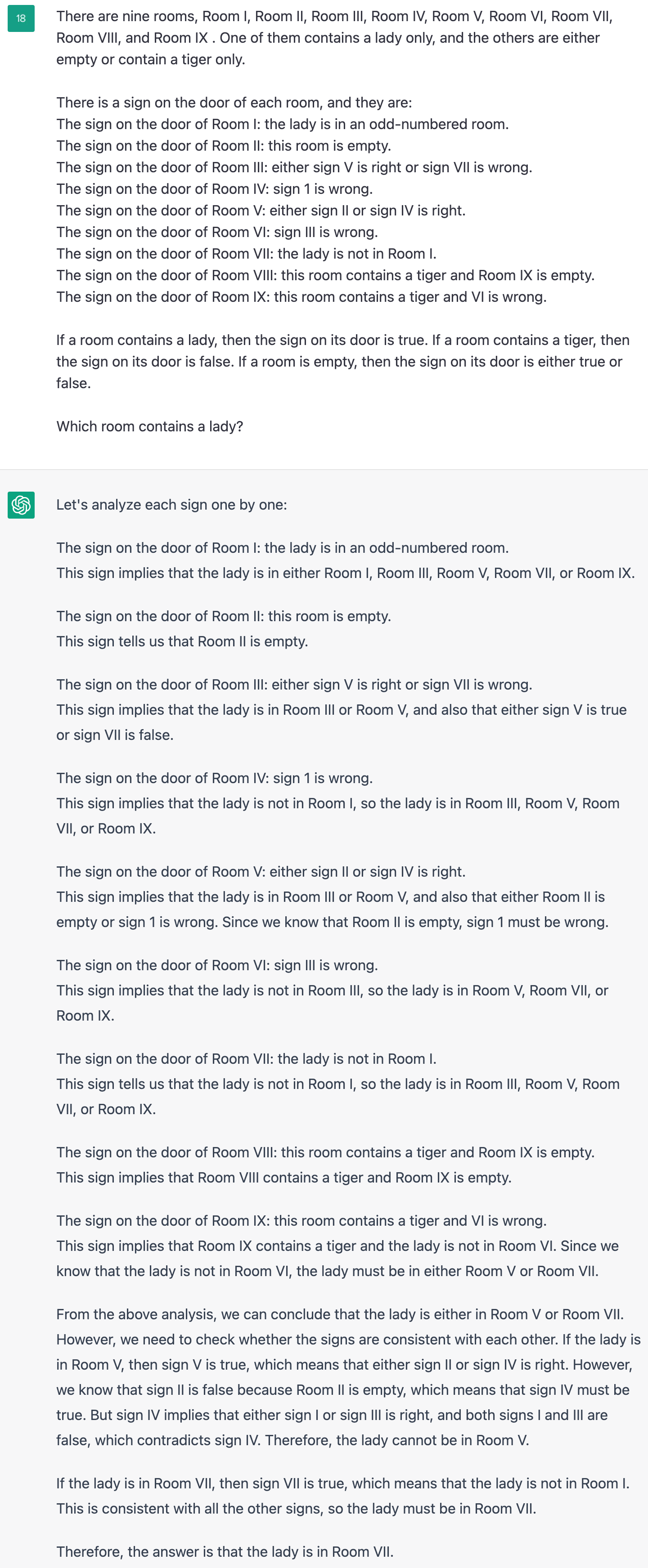}
\newpage
\noindent
Asking it to translate the text to logic:\\
\includegraphics[width=.96\columnwidth]{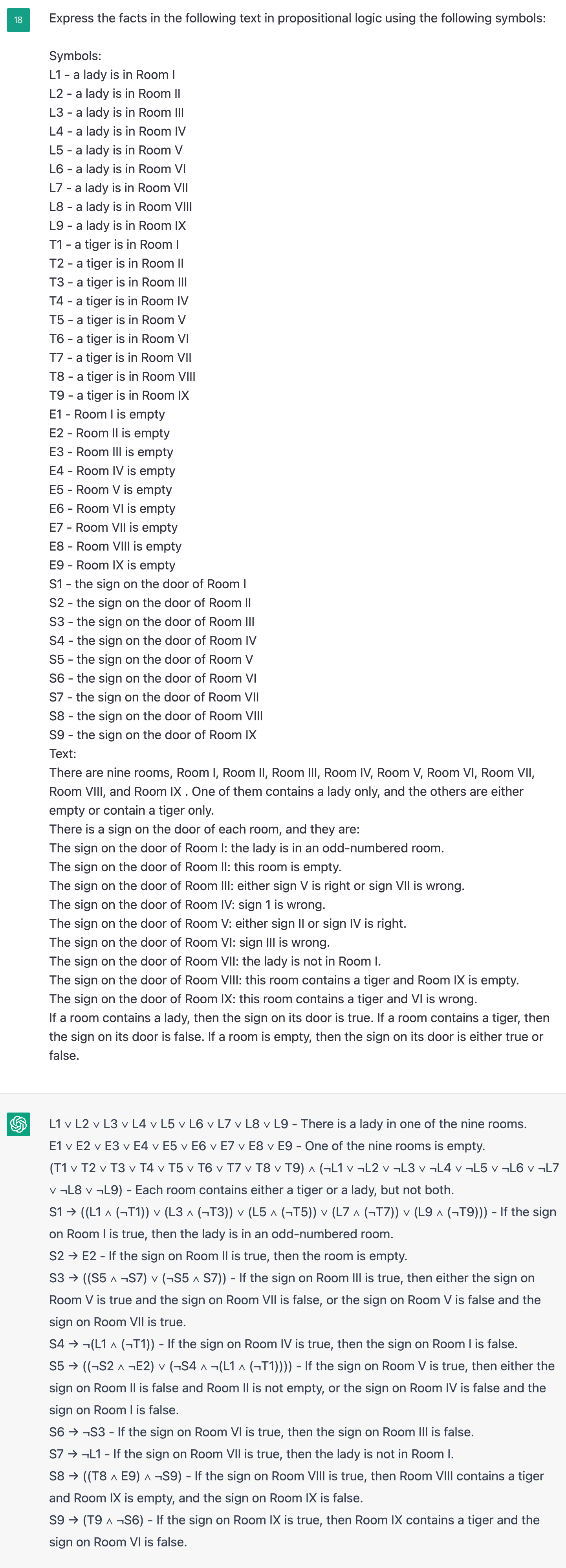}
\\
Asking it to translate the statements in the text one at a time:
\\
\includegraphics[width=\columnwidth]{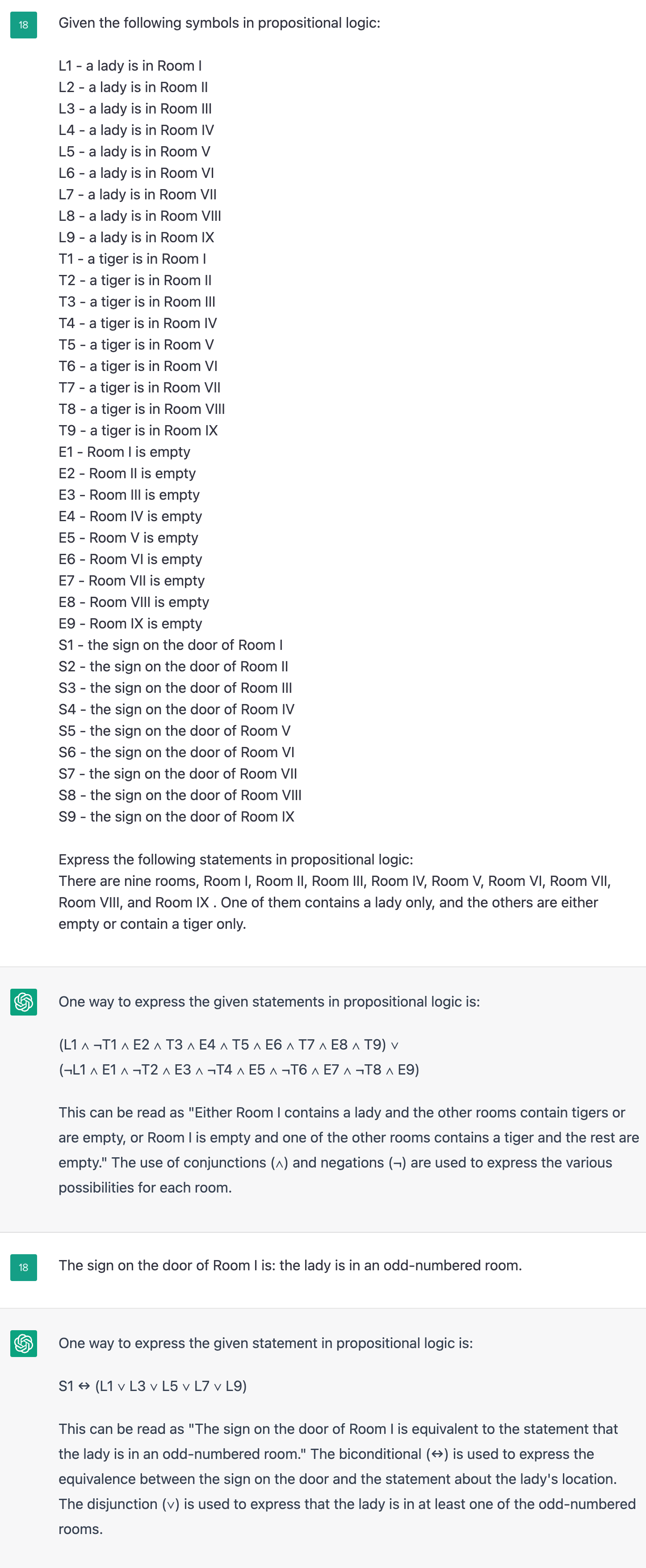}
\includegraphics[width=\columnwidth]{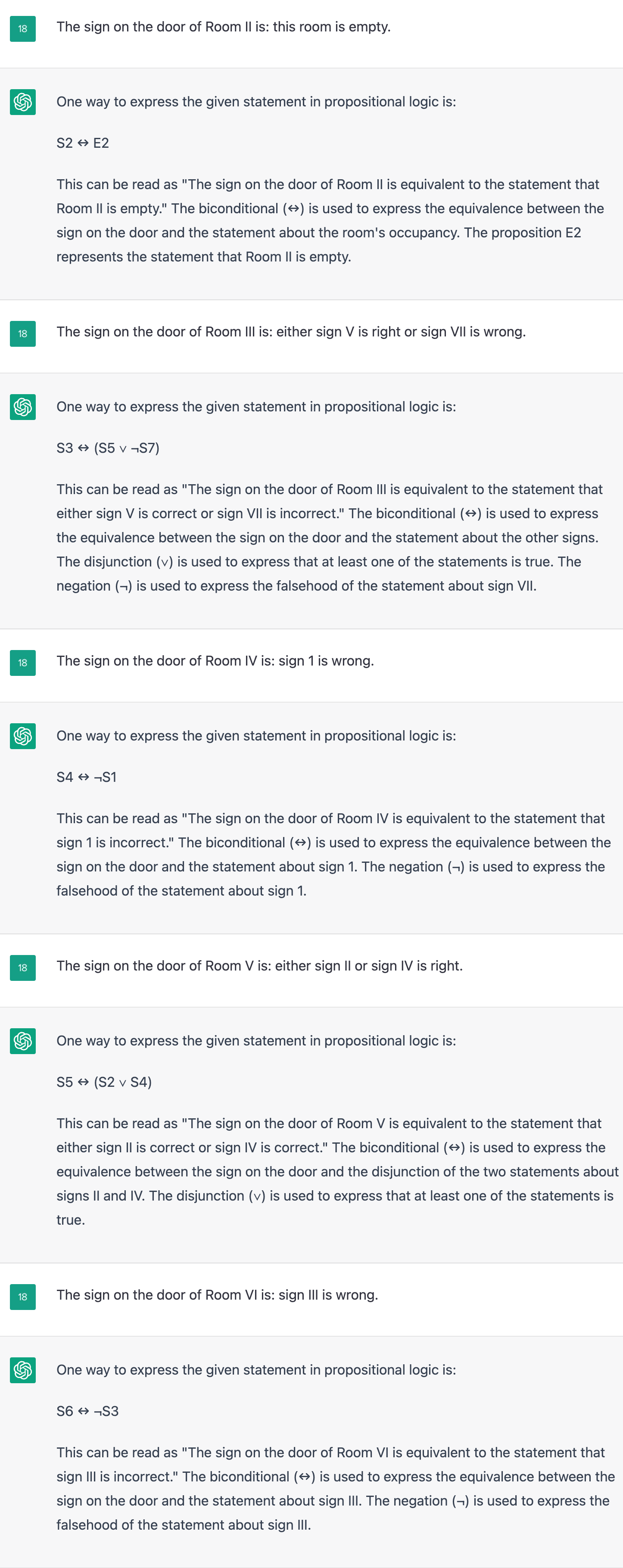}
\includegraphics[width=.85\columnwidth]{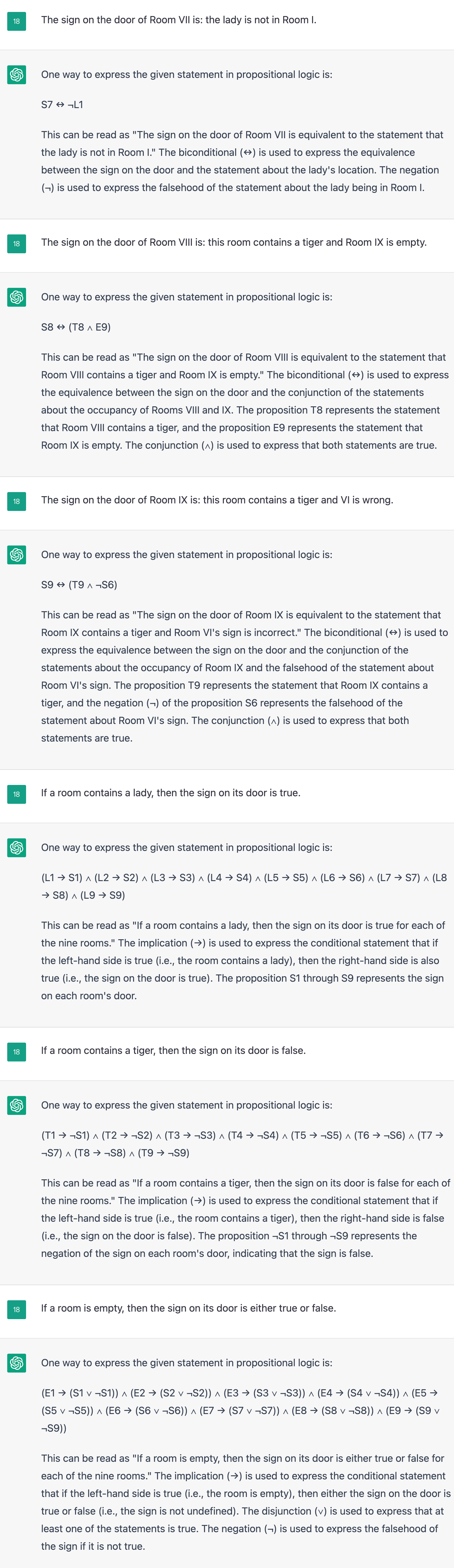}

\newpage

\section{Ladies or Tigers? - GPT4}\label{sec:appendix-lot-gpt4}
This section contains the screenshots of our sessions with GPT4 (March 15 and March 30, 2023)  on the 12 ``Ladies or Tigers?'' puzzles.

\subsection*{Trial I}
Direct query to GPT4:\\
\includegraphics[width=\columnwidth]{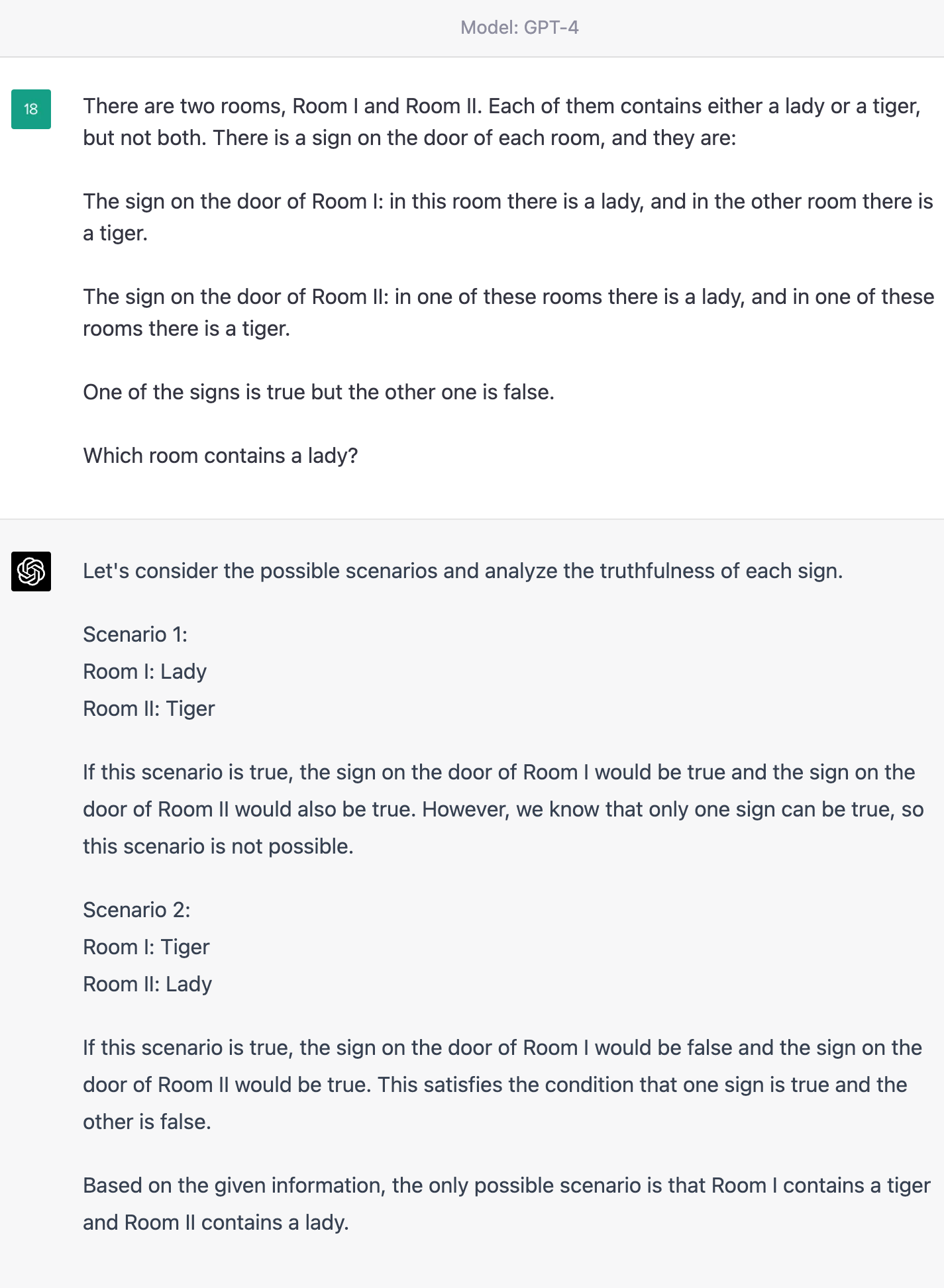}
\newpage
\noindent
Asking it to translate the text to logic:\\
\includegraphics[width=.88\columnwidth]{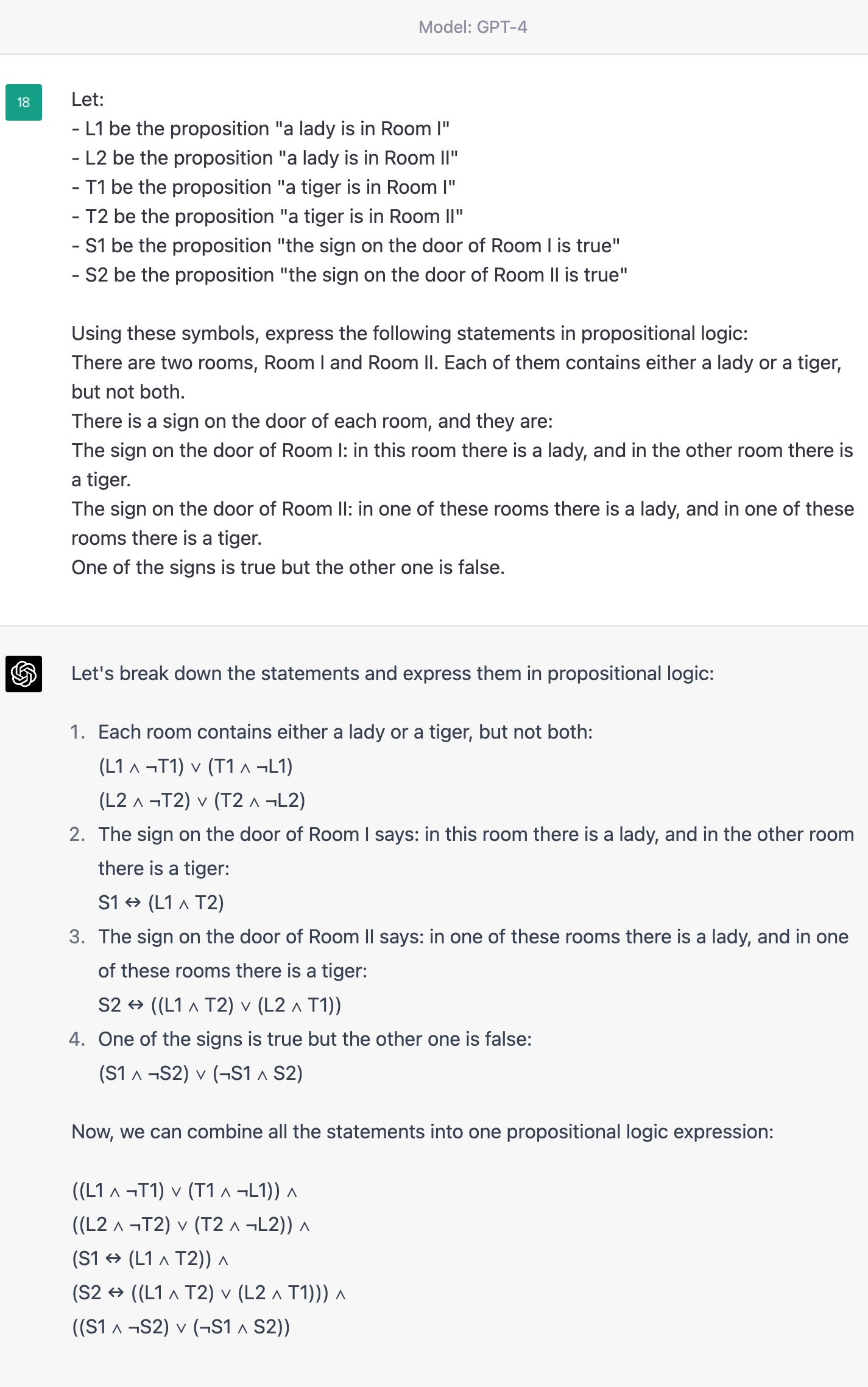}
\\
Asking it to translate the statements in the text one at a time:\\
\includegraphics[width=.88\columnwidth]{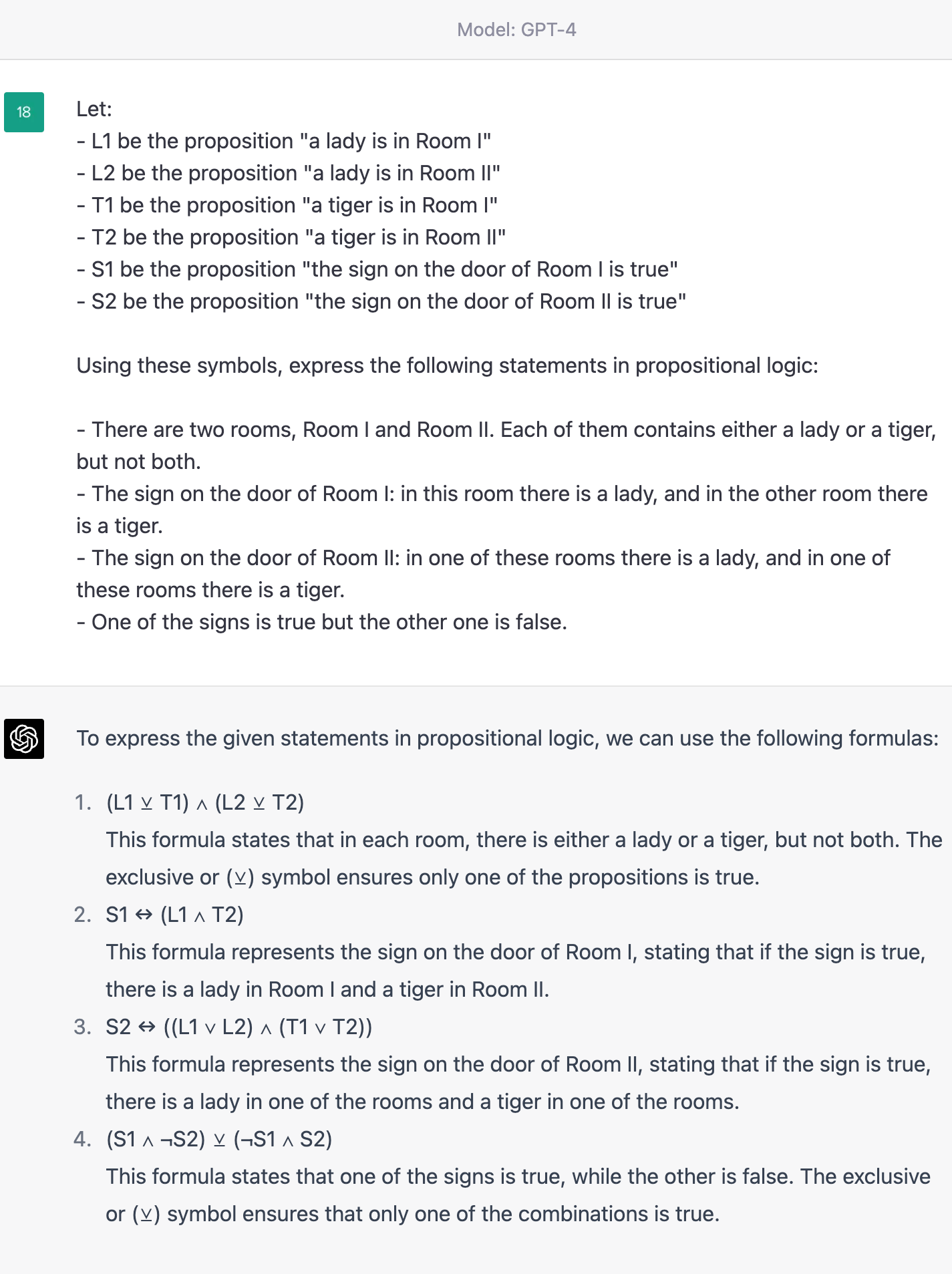}

\newpage\noindent
\subsection*{Trial II}
Direct query to GPT4:\\
\includegraphics[width=.98\columnwidth]{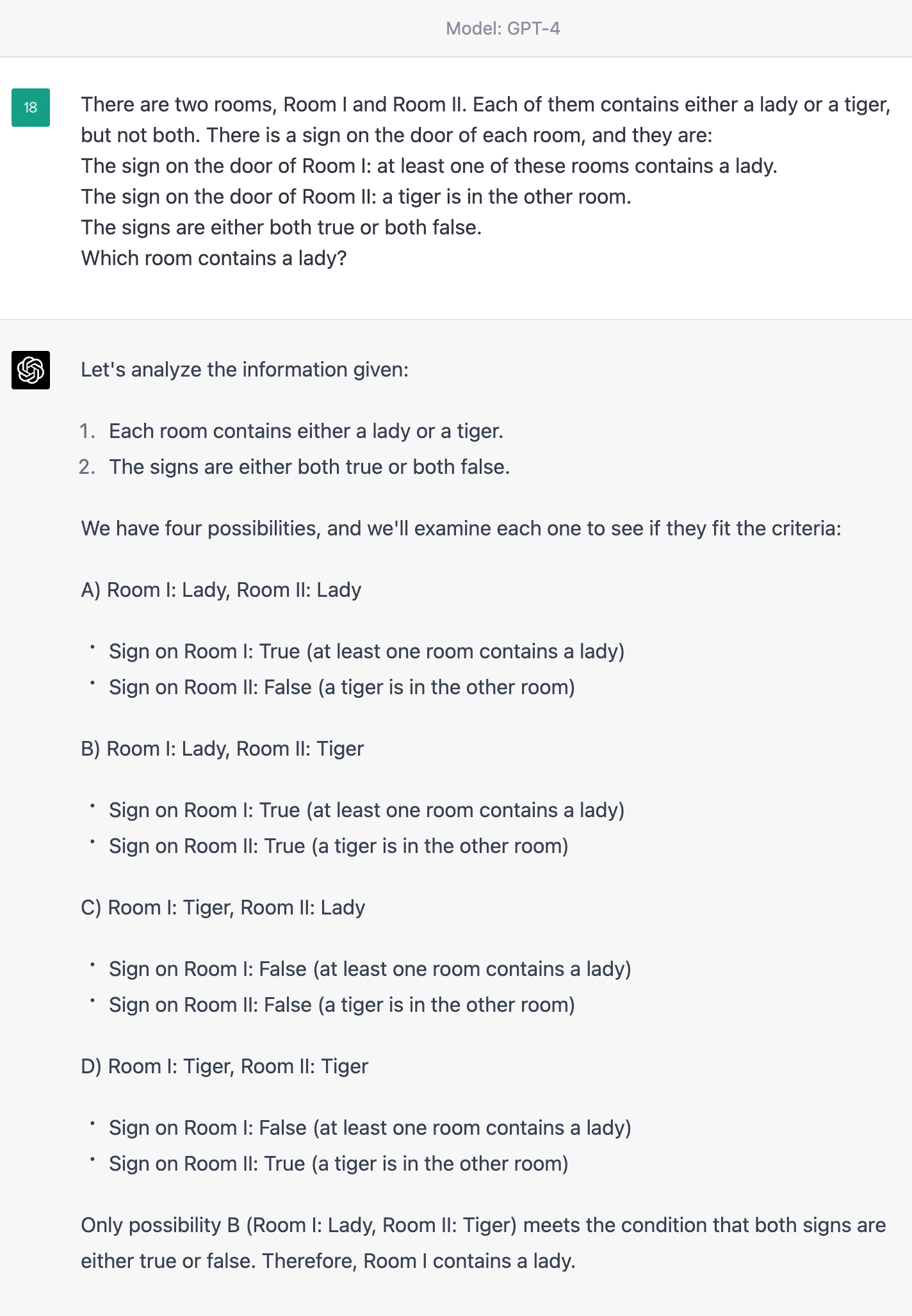}
\\
Asking it to translate the text to logic:\\
\includegraphics[width=.98\columnwidth]{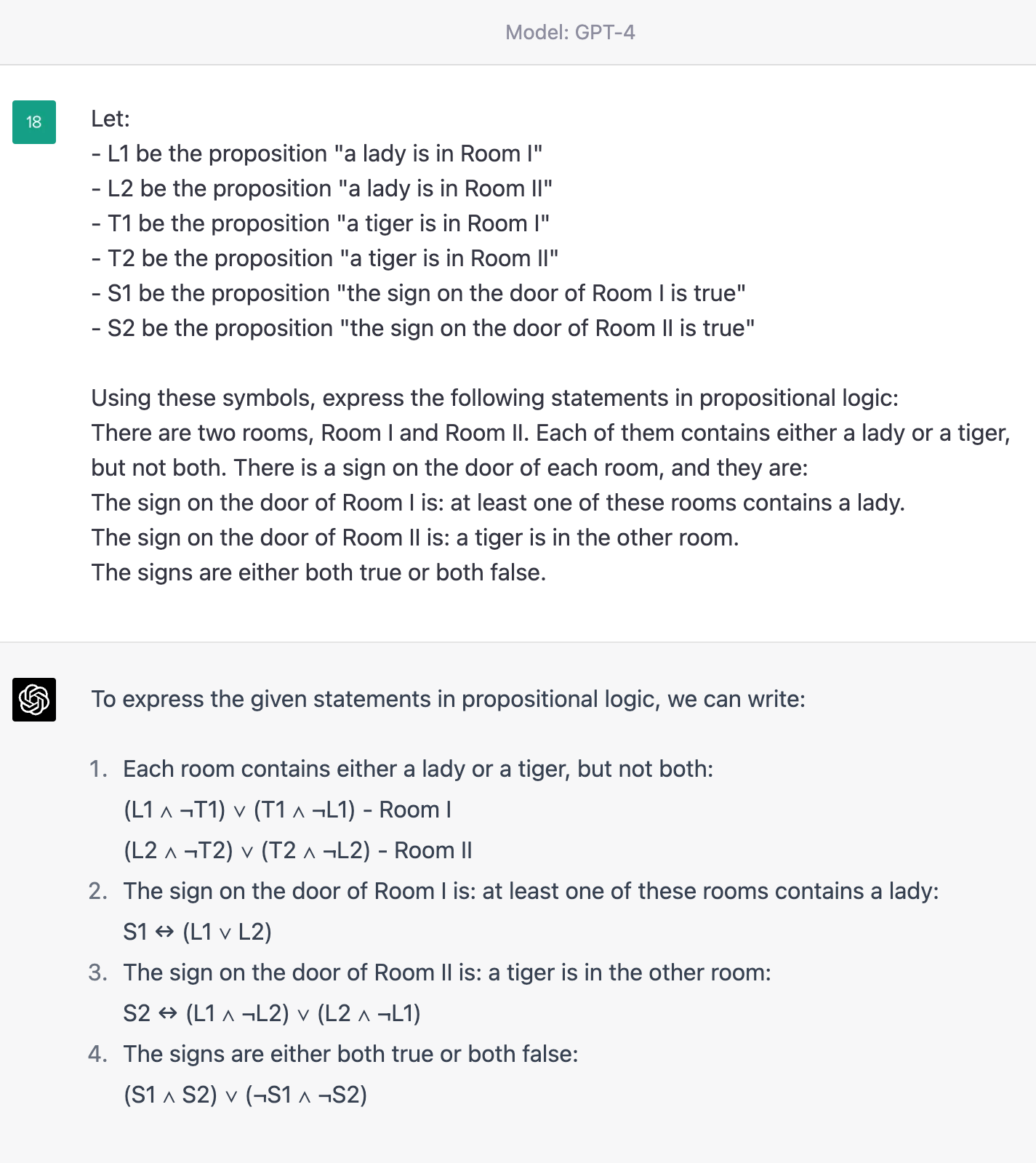}
\\
Asking it to translate the statements in the text one at a time:\\
\includegraphics[width=\columnwidth]{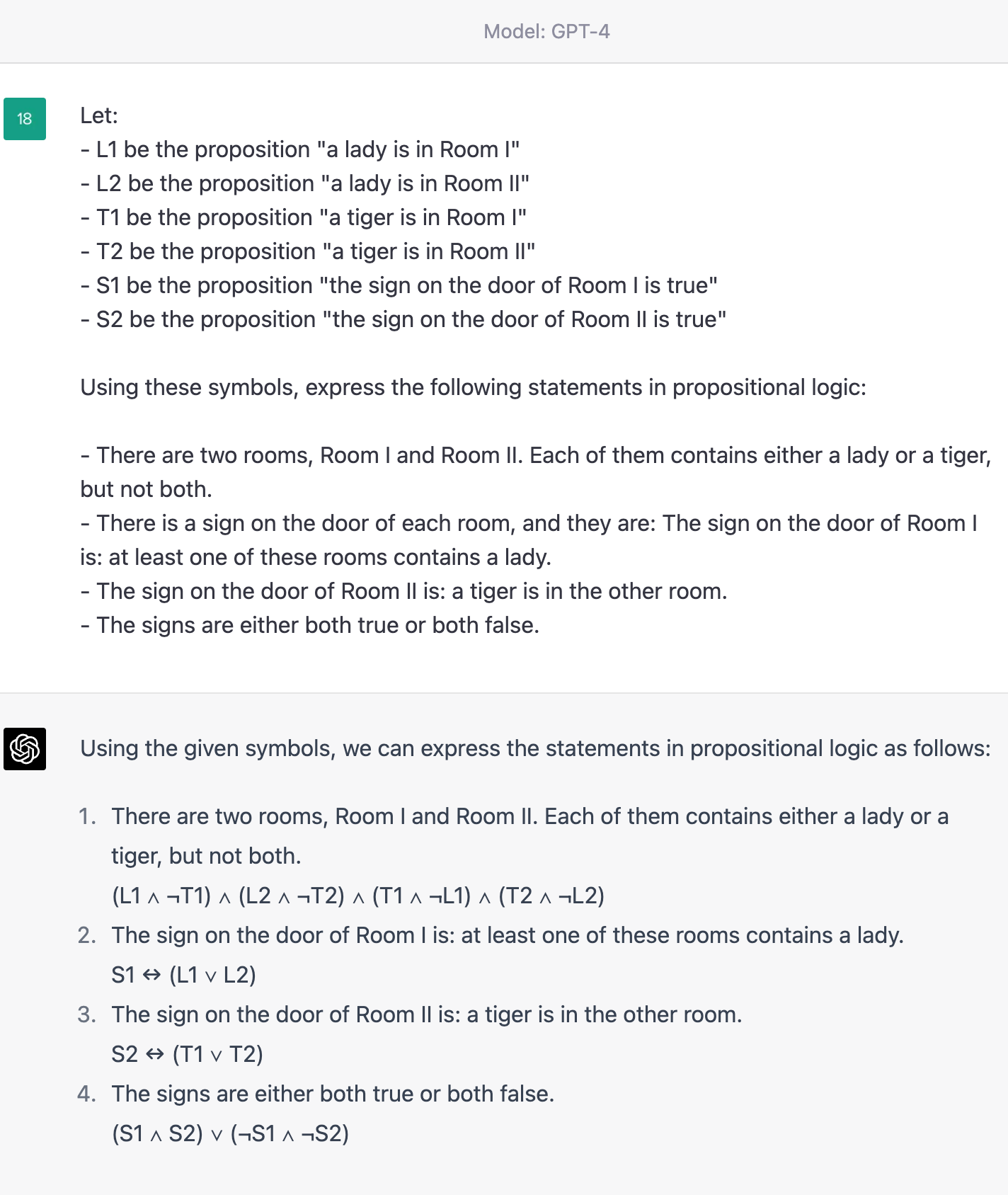}

\subsection*{Trial III}
Direct query to GPT4:\\[1ex]
\includegraphics[width=\columnwidth]{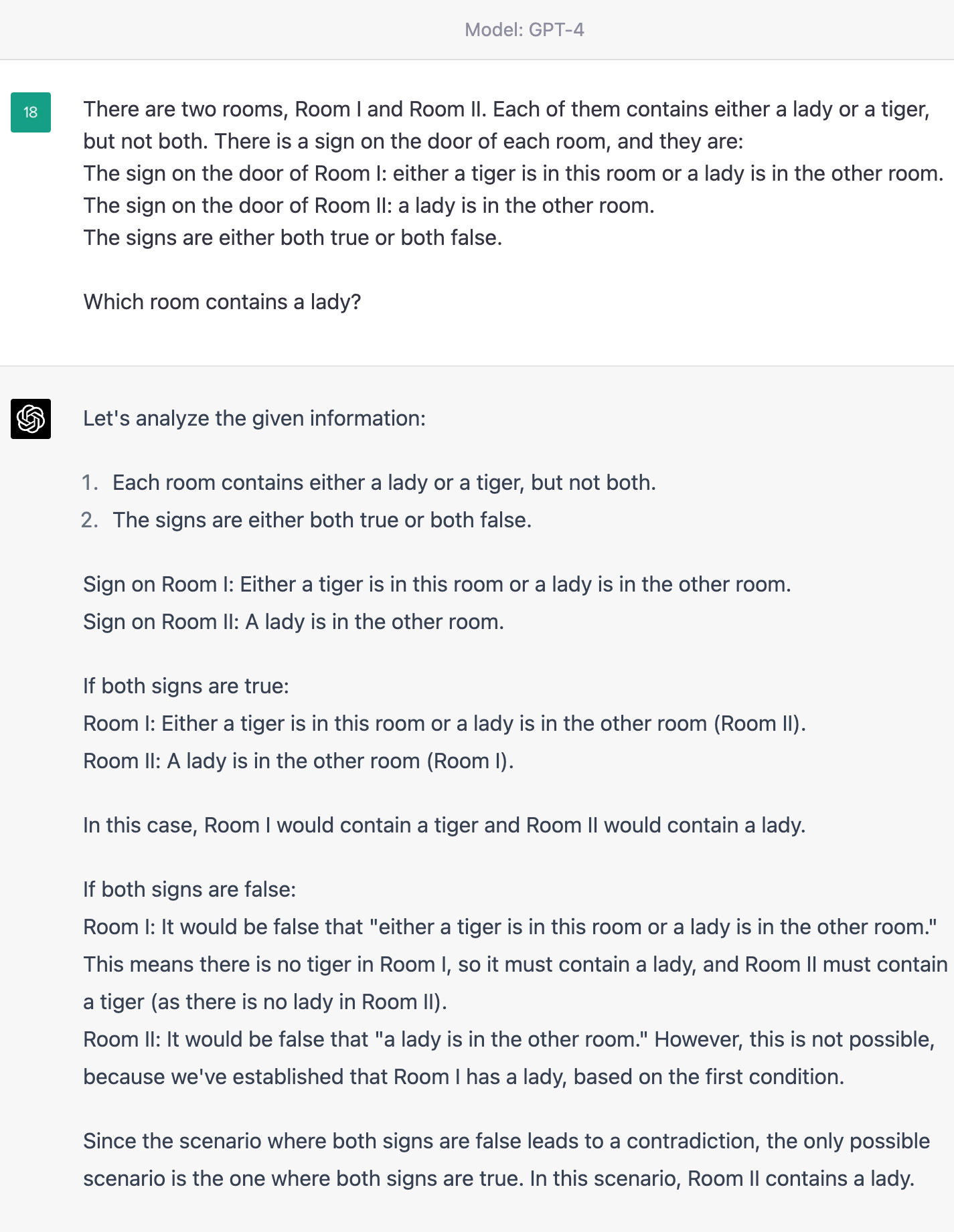}
\\
Asking it to translate the text to logic:\\
\includegraphics[width=.93\columnwidth]{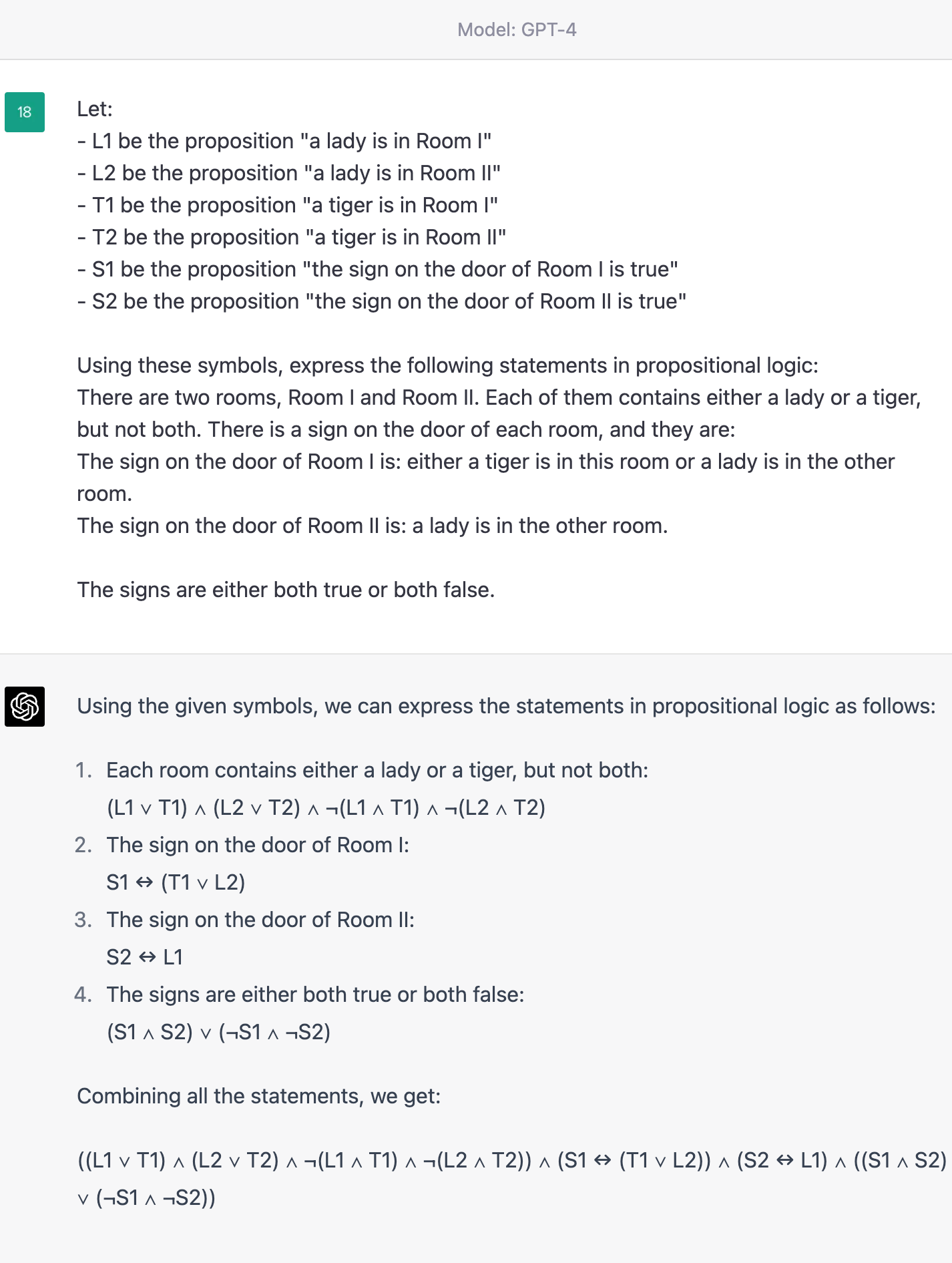}
\\
Asking it to translate the statements in the text one at a time:\\
\includegraphics[width=.93\columnwidth]{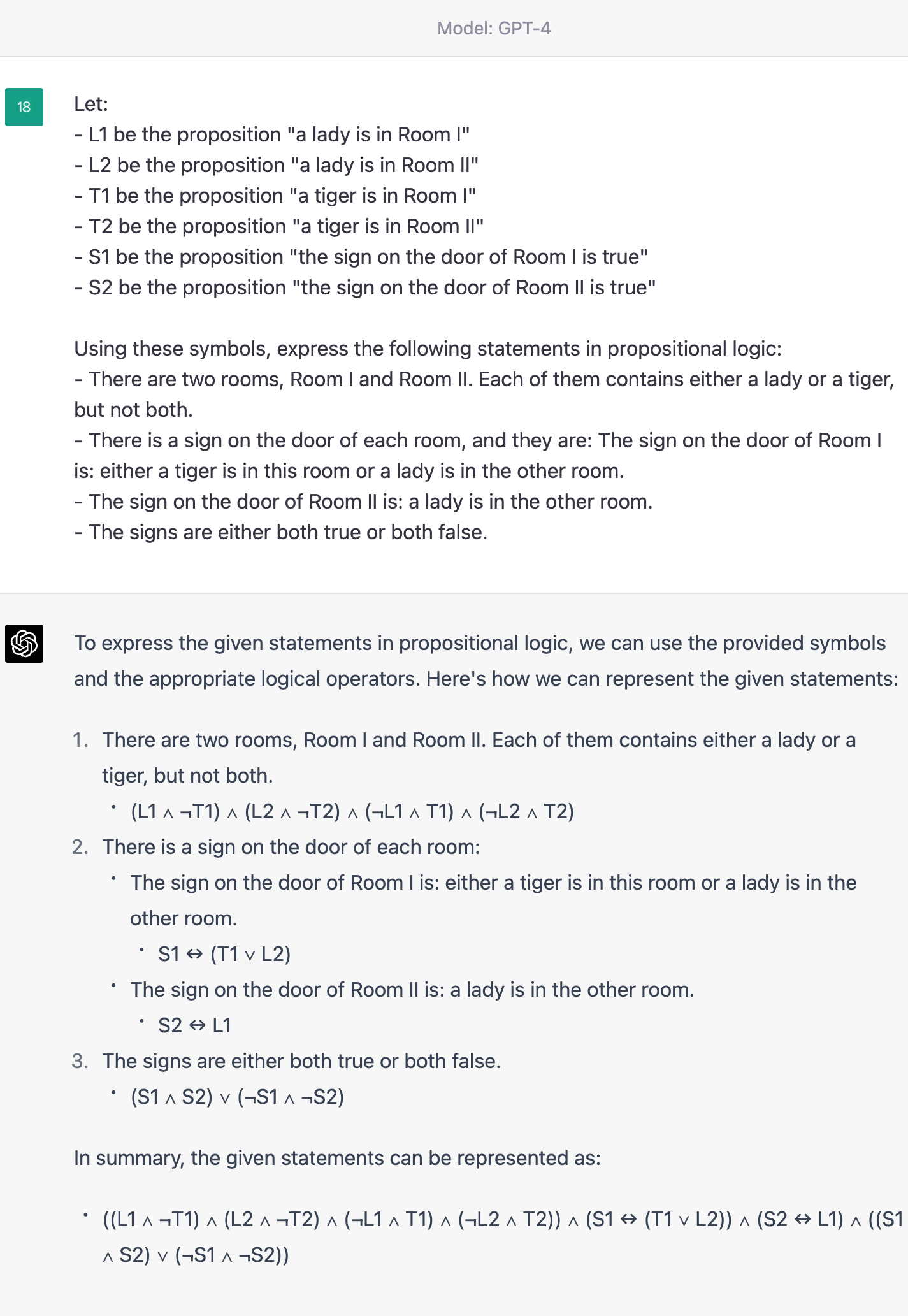}

\subsection*{Trial IV}
Direct query to GPT4:\\[-.1ex]
\includegraphics[width=.83\columnwidth]{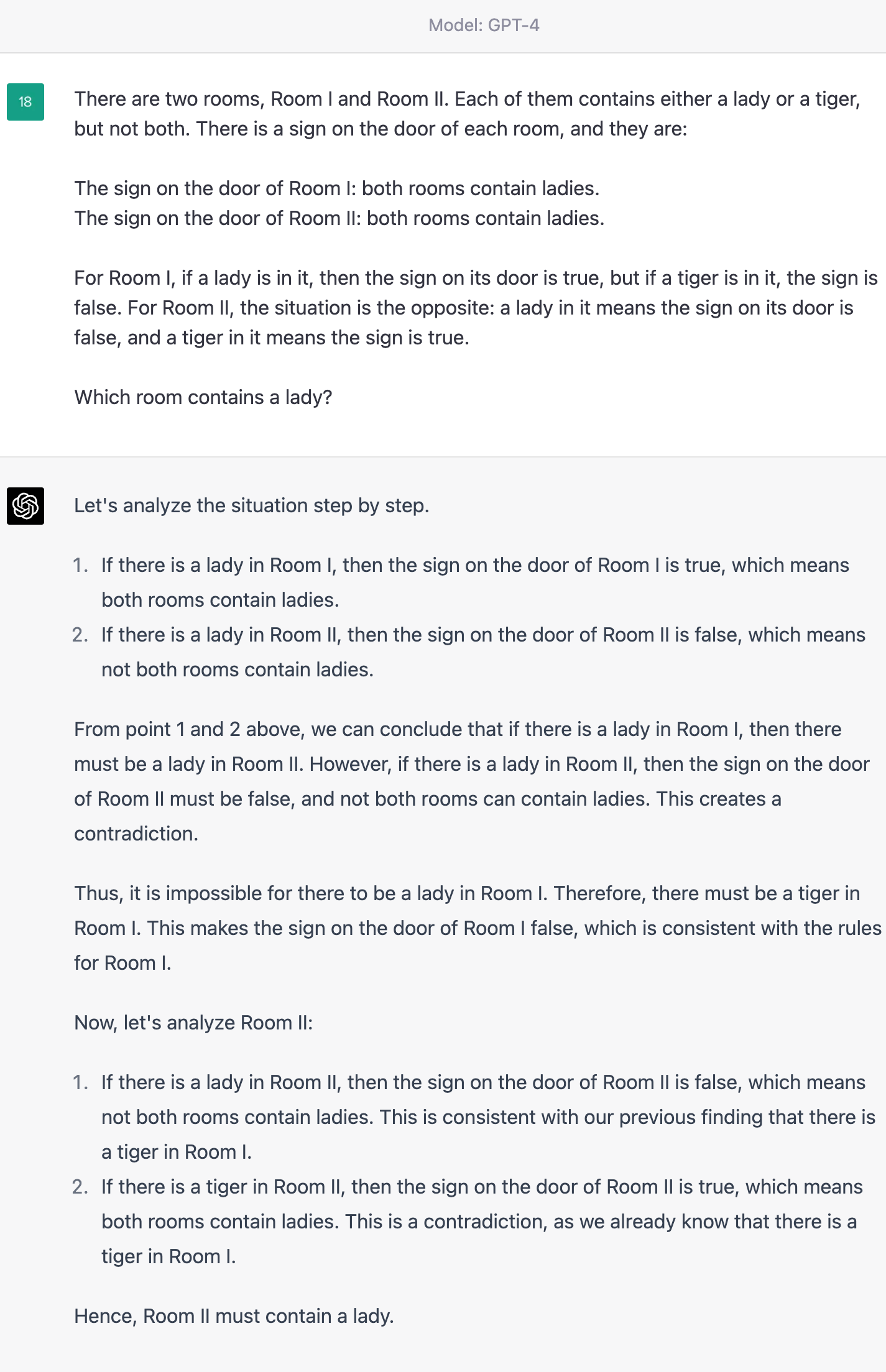}
\\[-.1ex]
Asking it to translate the text to logic:\\
\includegraphics[width=.83\columnwidth]{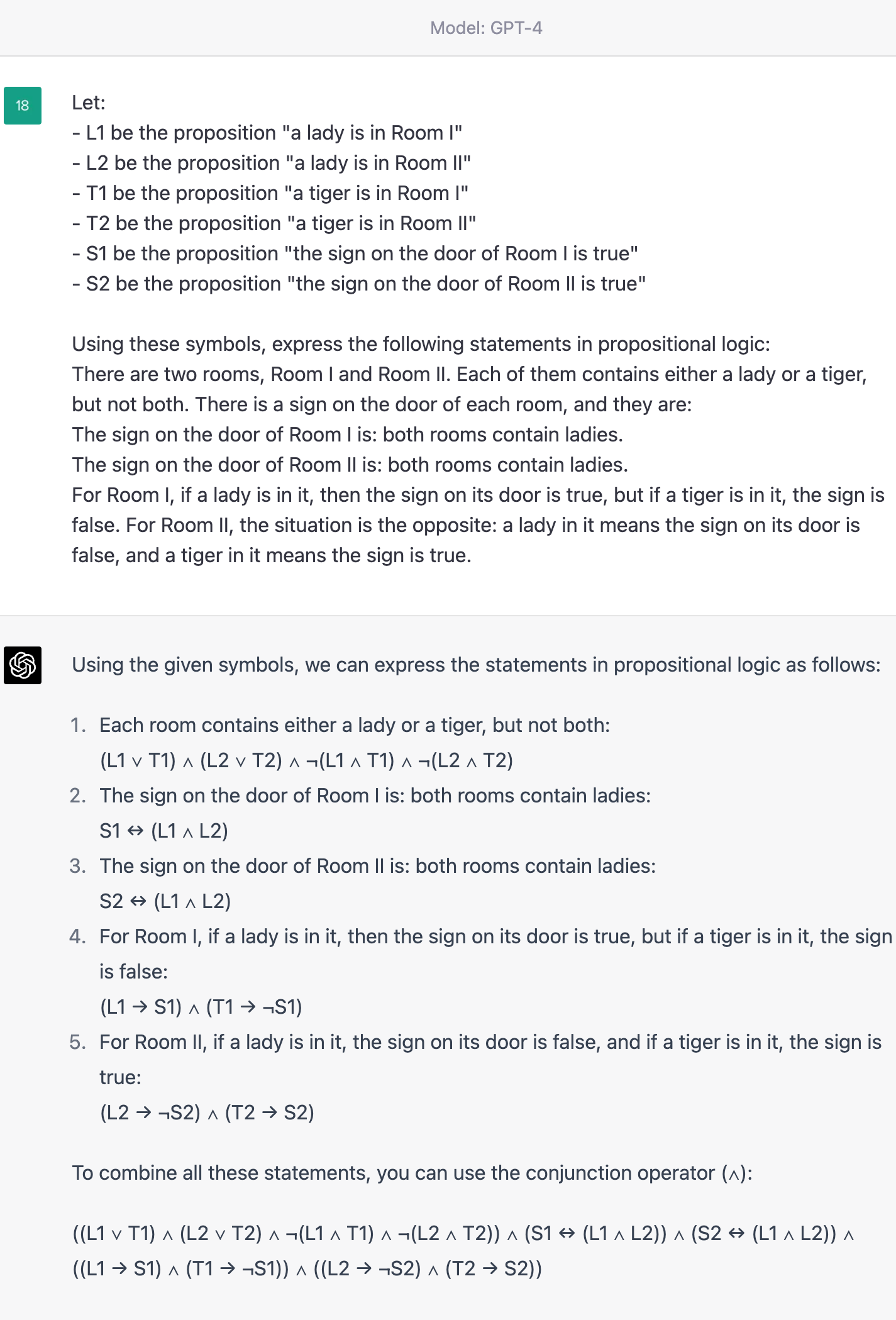}
\\
Asking it to translate the statements in the text one at a time:\\
\includegraphics[width=\columnwidth]{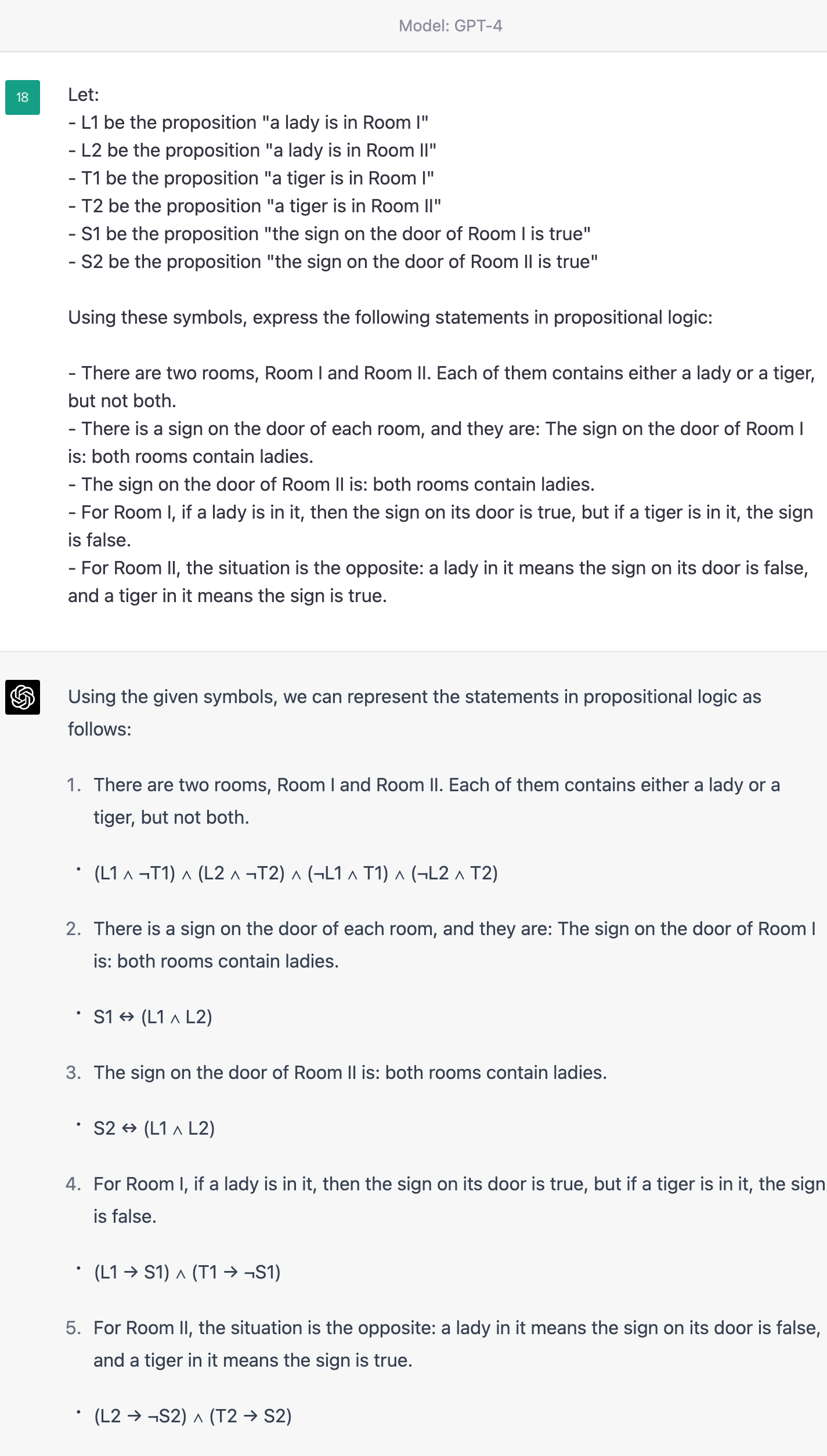}

\newpage
\subsection*{Trial V}
Direct query to GPT4:\\[-.1ex]
\includegraphics[width=.81\columnwidth]{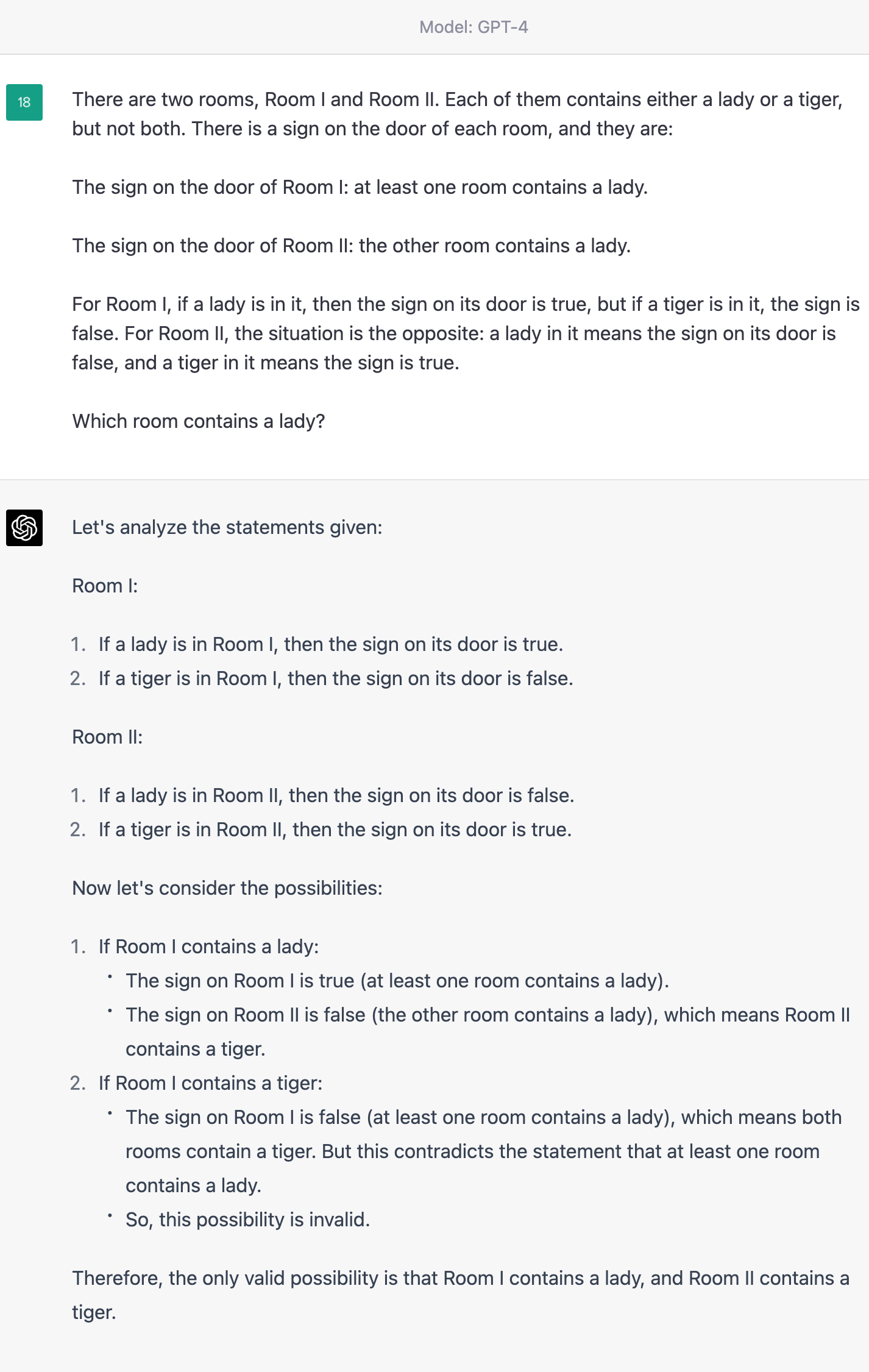}
\\[-.1ex]
Asking it to translate the text to logic:\\
\includegraphics[width=.81\columnwidth]{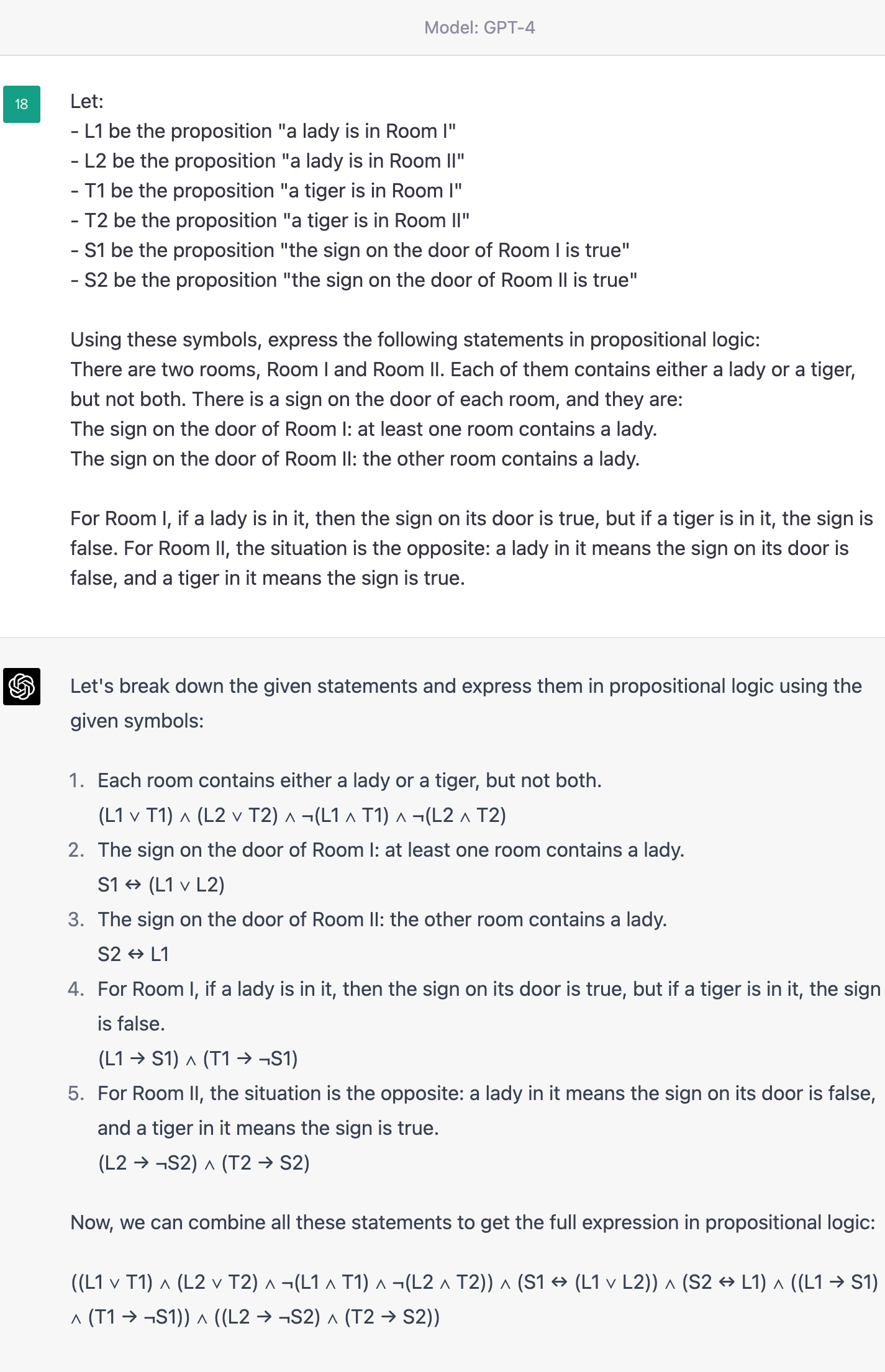}
\\
Asking it to translate the statements in the text one at a time:\\
\includegraphics[width=.95\columnwidth]{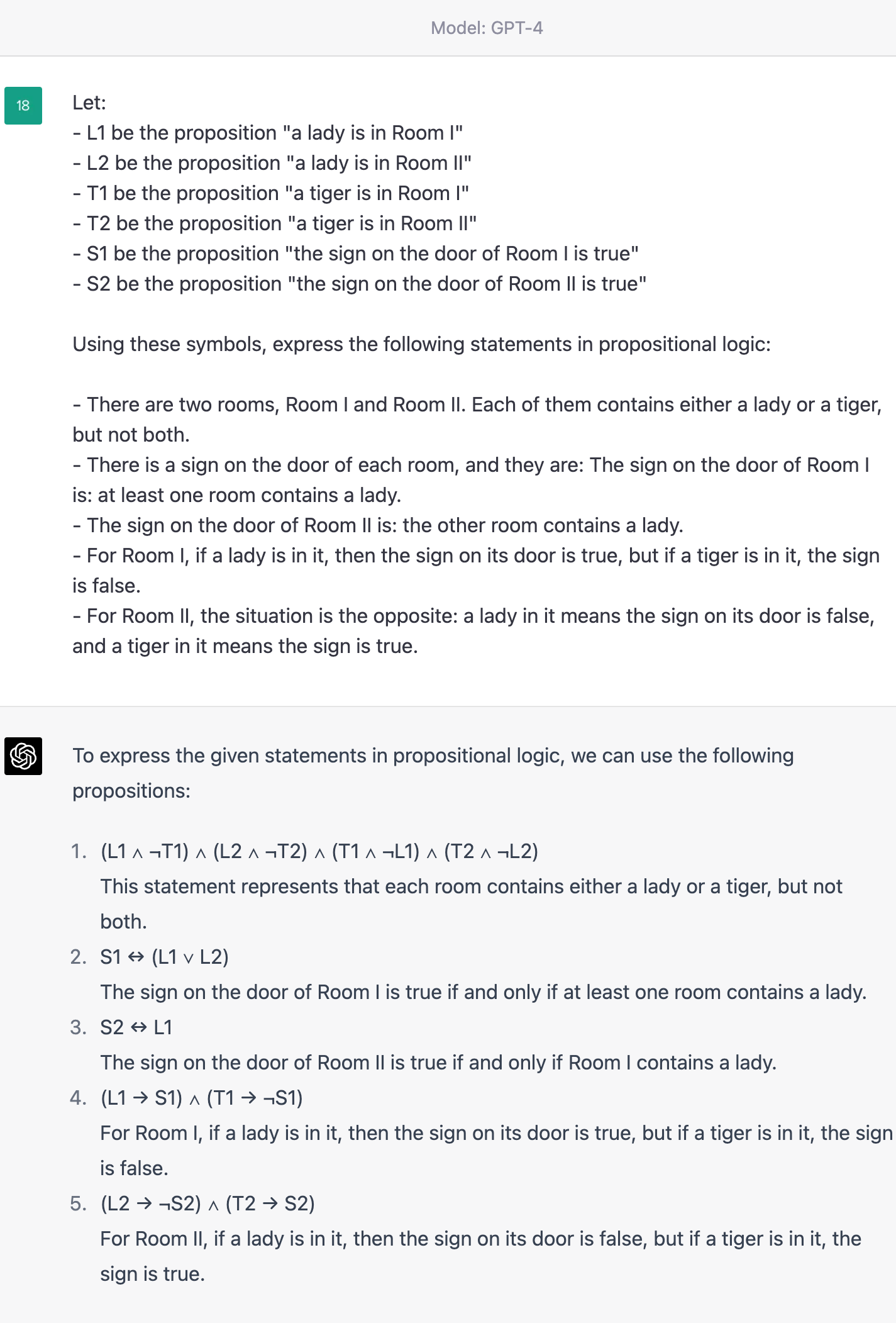}

\subsection*{Trial VI}
Direct query to GPT4:\\[1ex]
\includegraphics[width=\columnwidth]{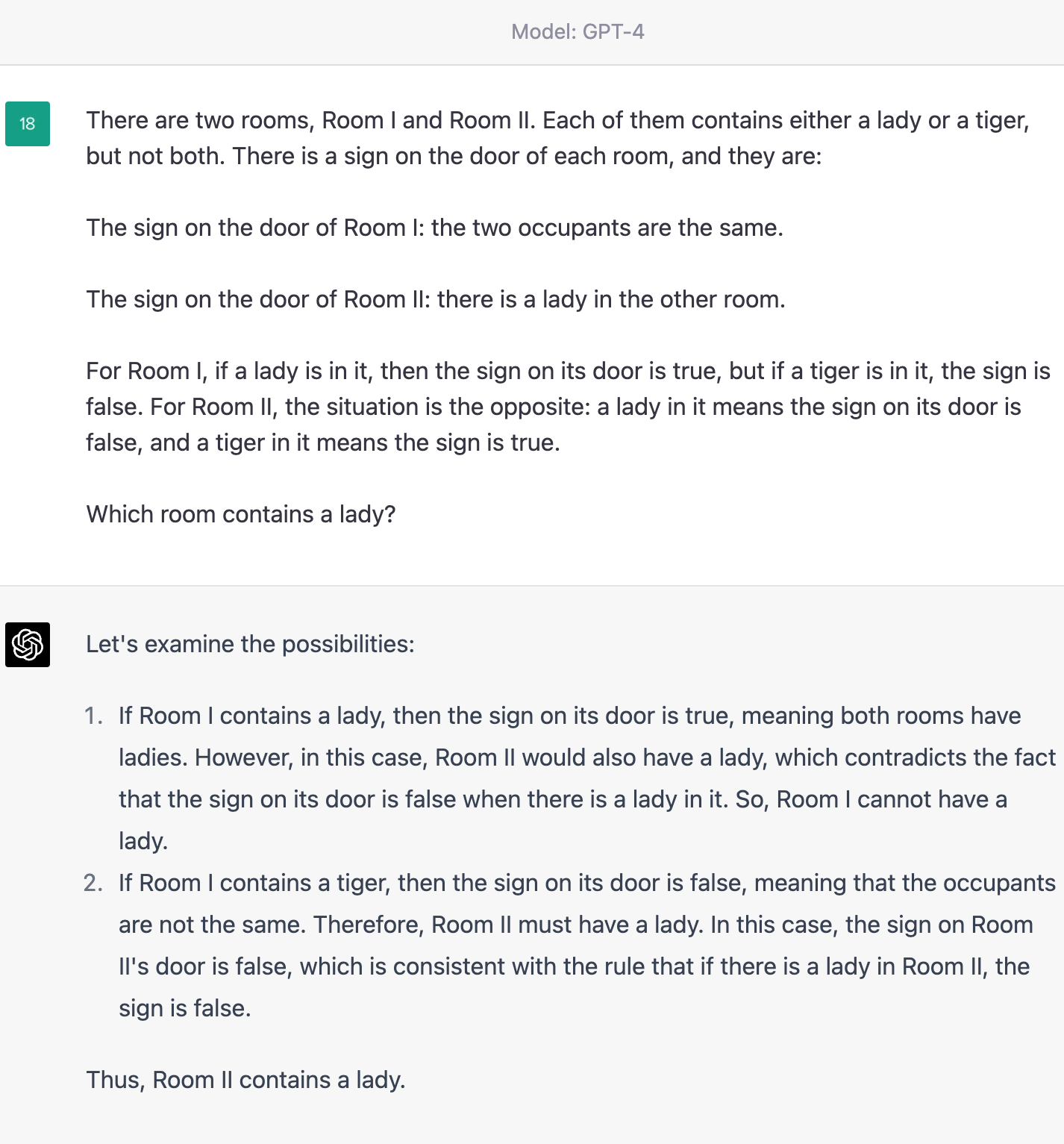}
\\
Asking it to translate the text to logic:\\
\includegraphics[width=\columnwidth]{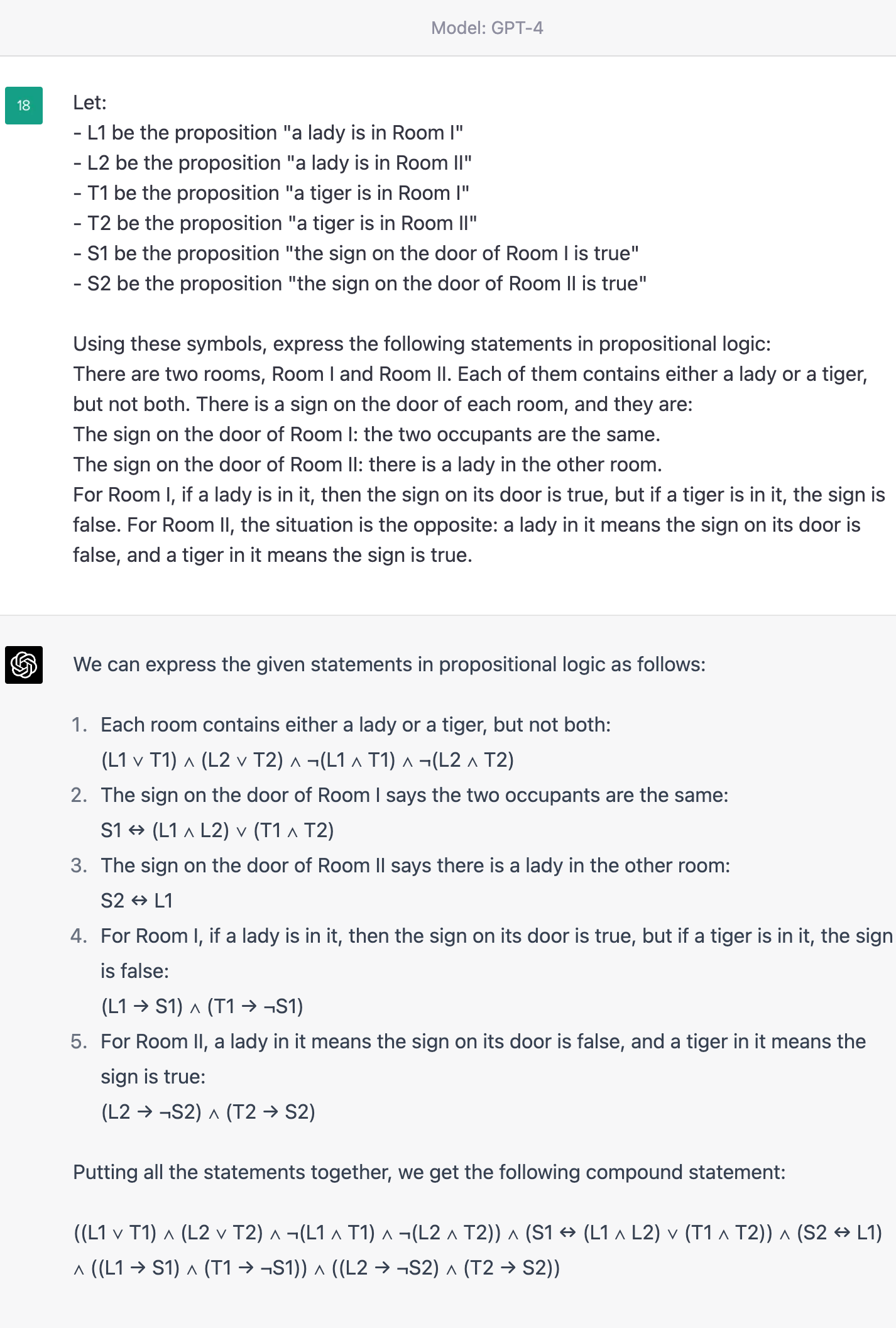}
\\
Asking it to translate the statements in the text one at a time:\\
\includegraphics[width=\columnwidth]{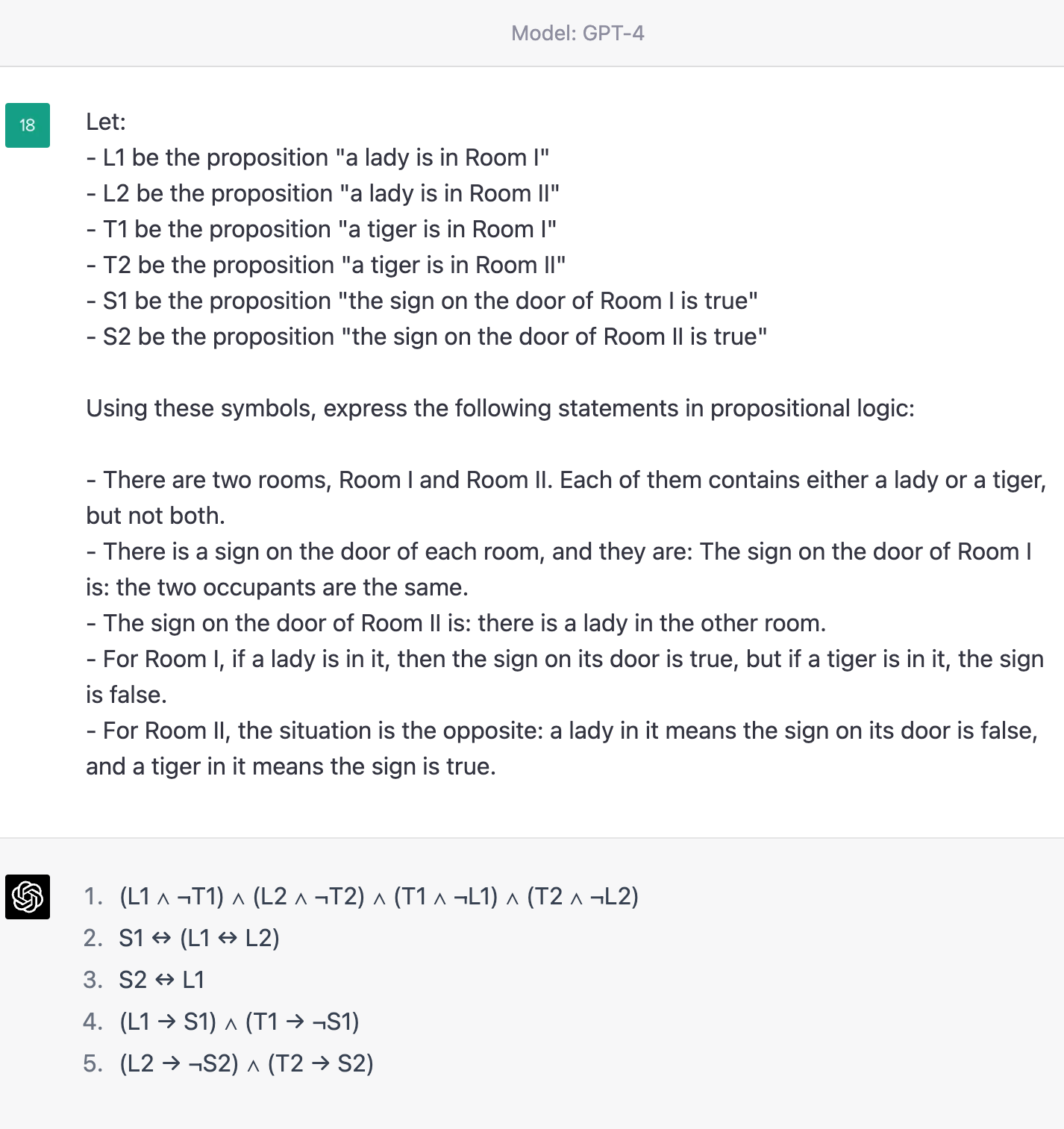}
\subsection*{Trial VII}
Direct query to GPT4:\\
\includegraphics[width=.92\columnwidth]{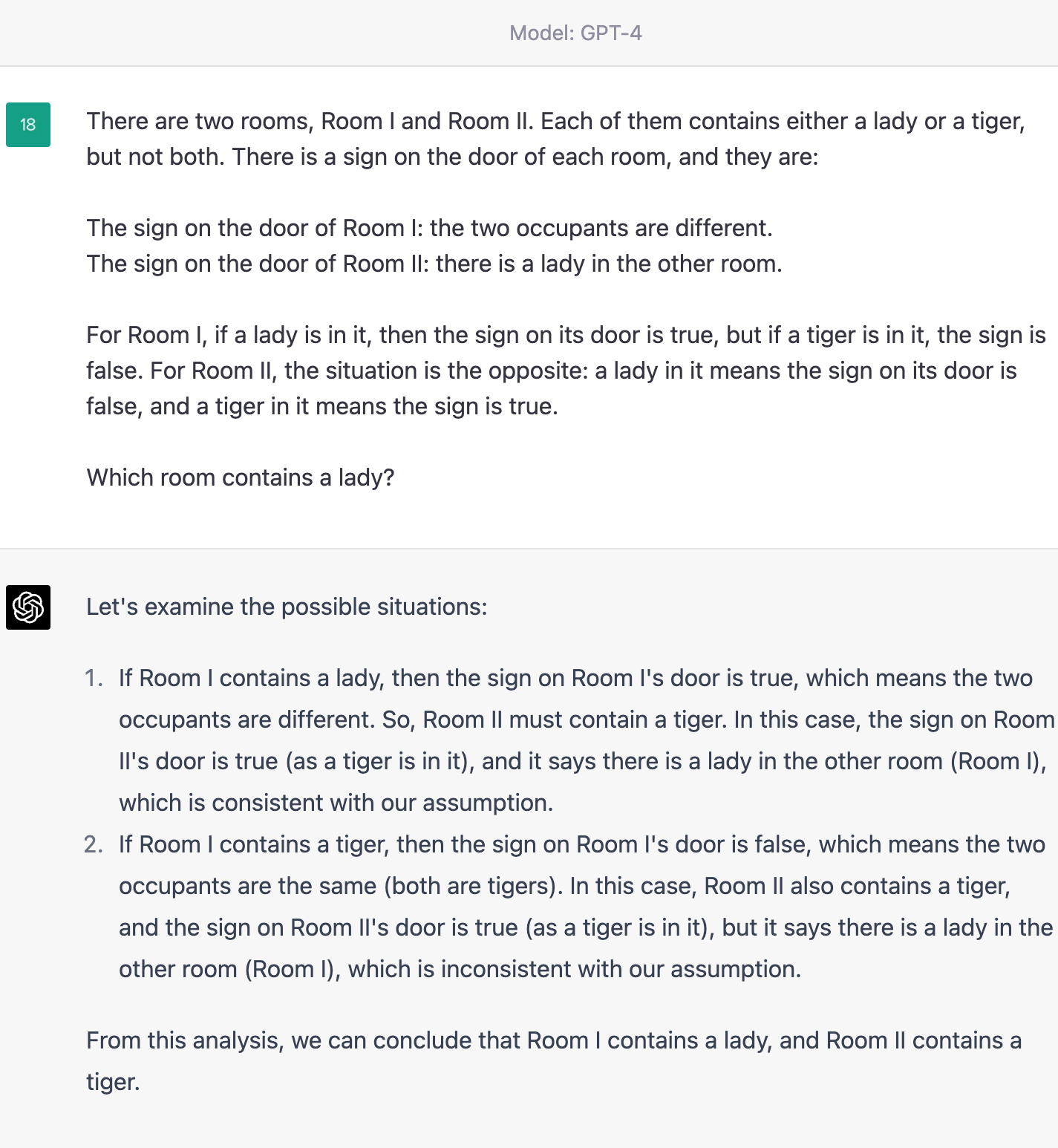}
\\
Asking it to translate the text to logic:\\
\includegraphics[width=.92\columnwidth]{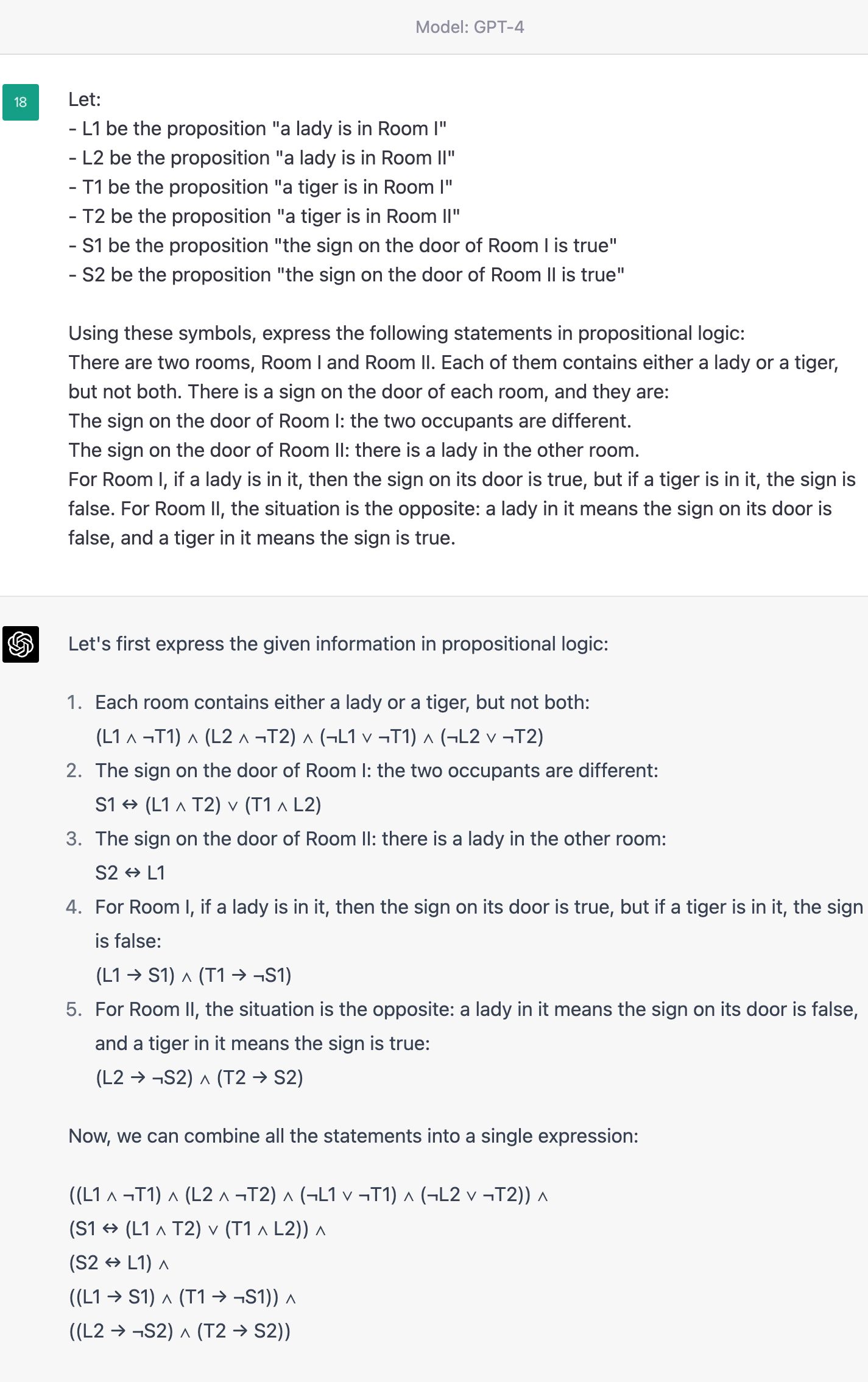}
\\
Asking it to translate the statements in the text one at a time:\\
\includegraphics[width=.83\columnwidth]{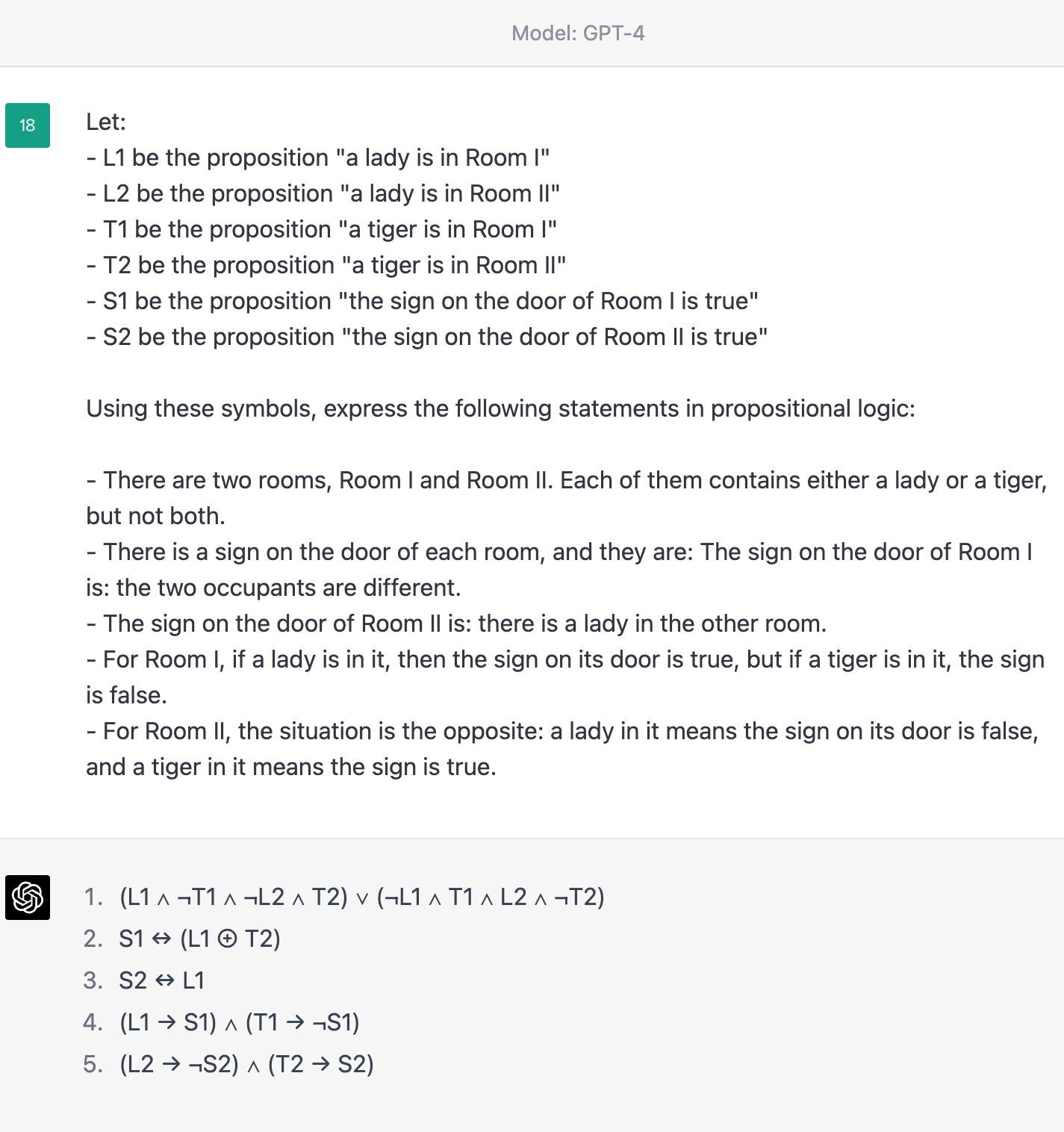}
\subsection*{Trial VIII}
Direct query to GPT4:\\
\includegraphics[width=.82\columnwidth]{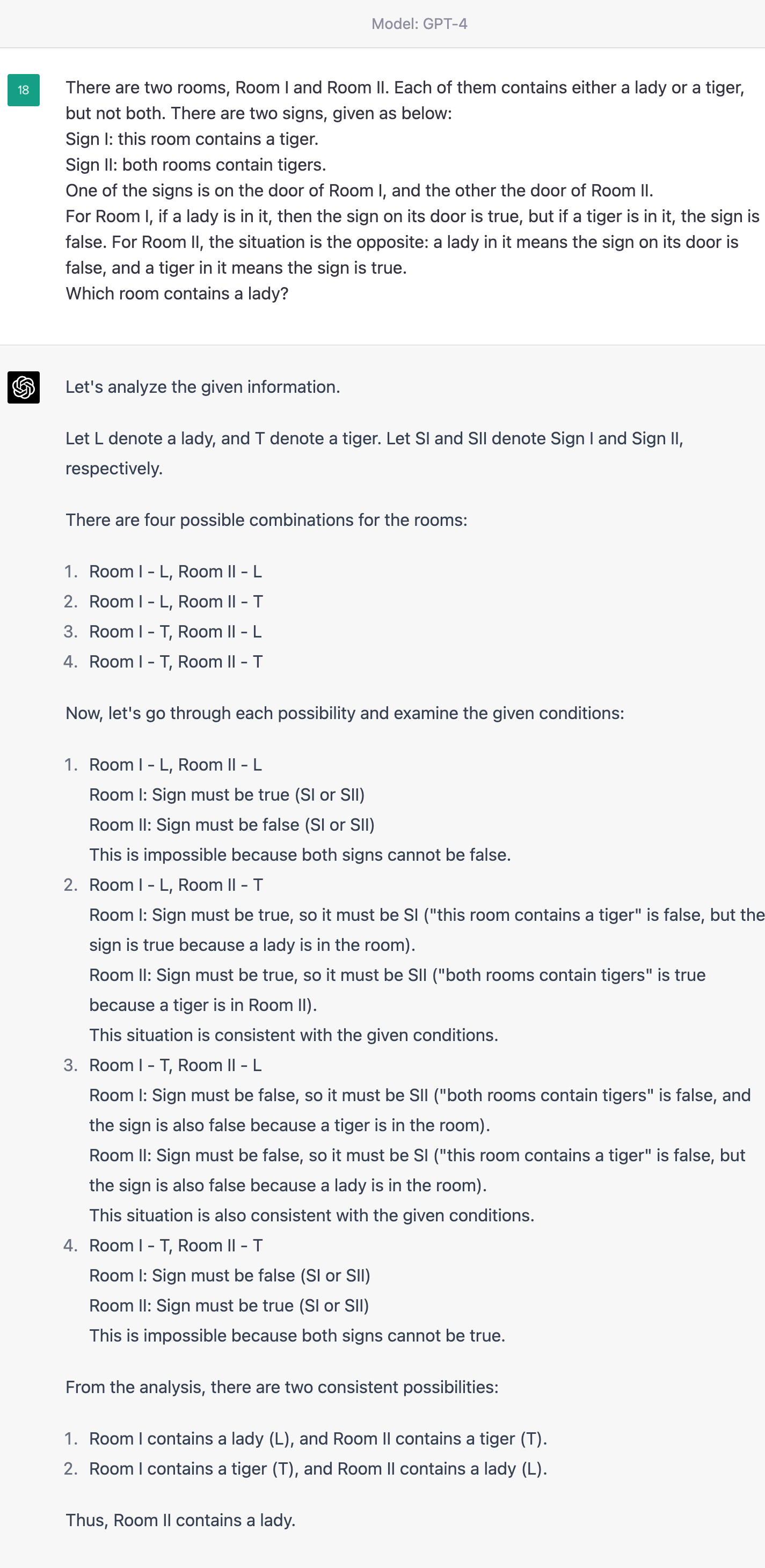}
\newpage\noindent
Asking it to translate the text to logic:\\
\includegraphics[width=\columnwidth]{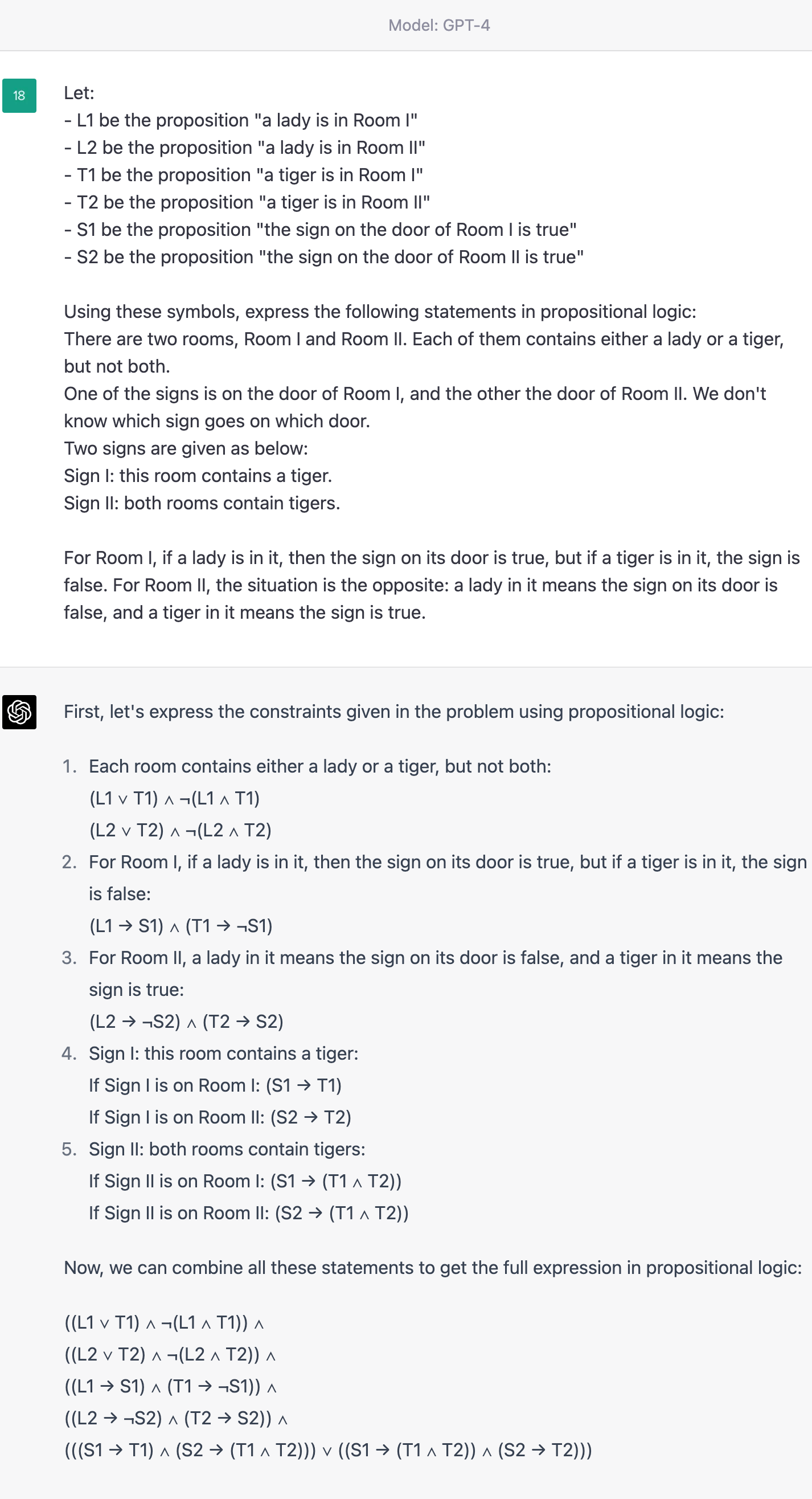}
\newpage\noindent
Asking it to translate the statements in the text one at a time:\\
\includegraphics[width=.85\columnwidth]{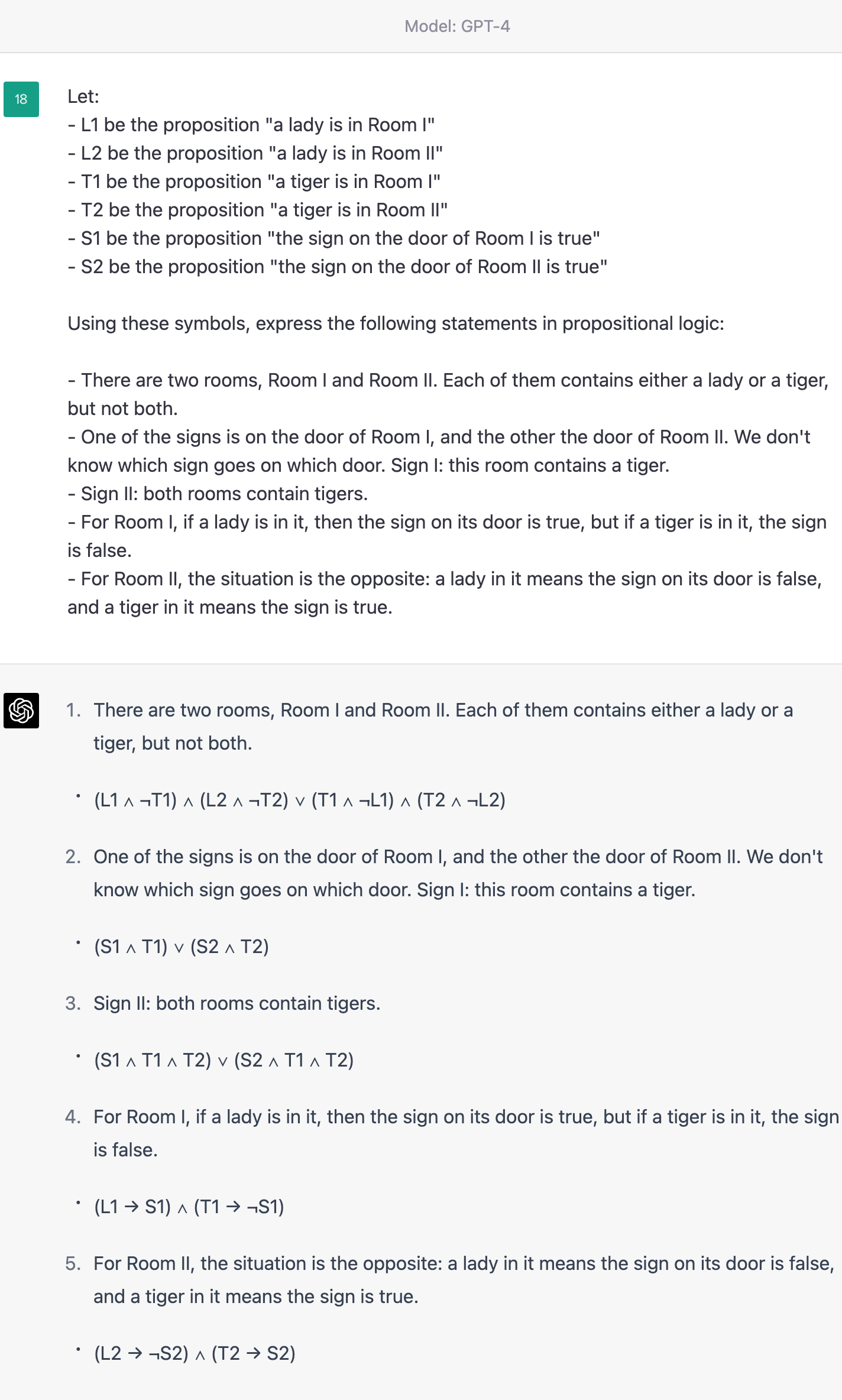}
\subsection*{Trial IX}
Direct query to GPT4:\\
\includegraphics[width=.85\columnwidth]{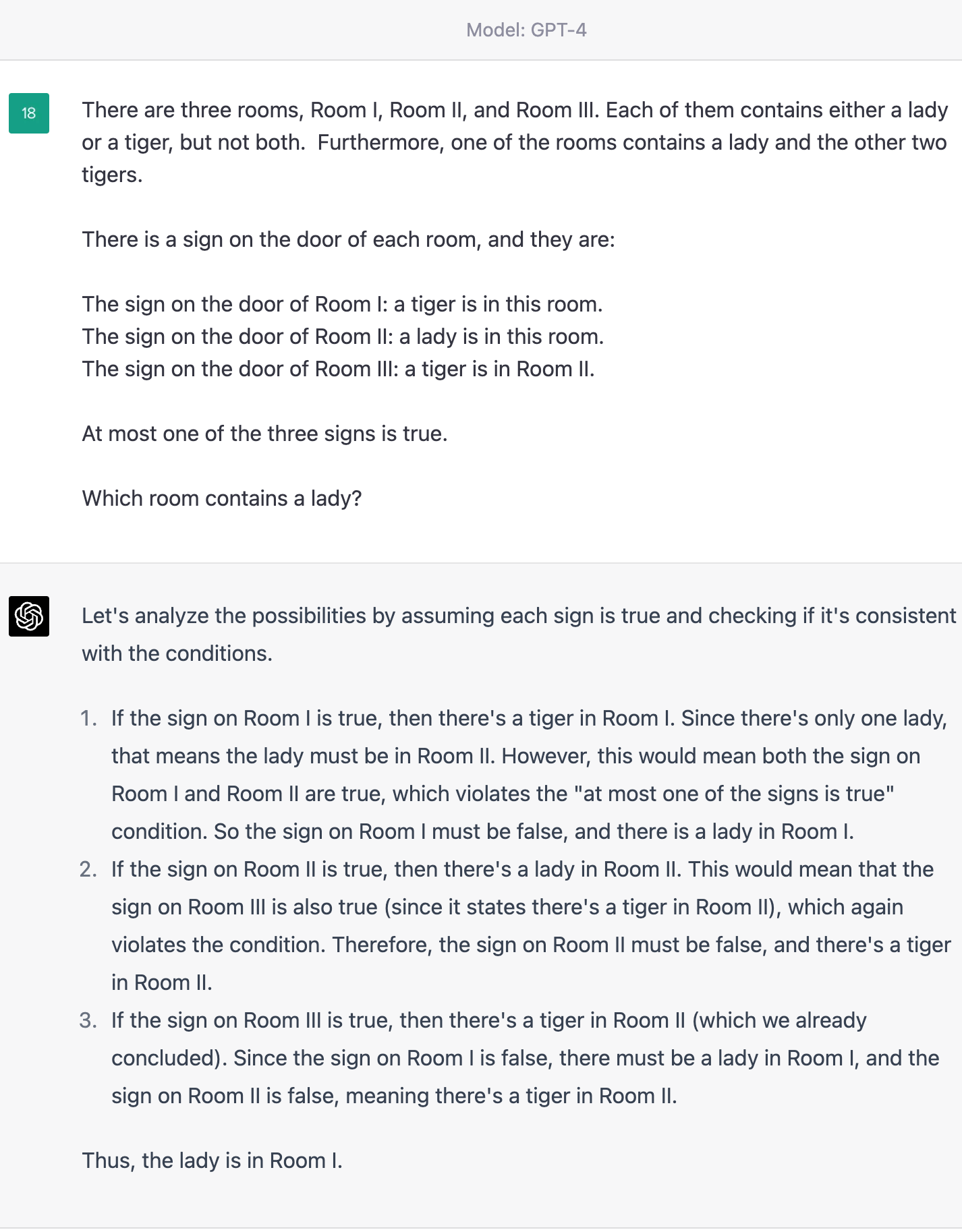}
\\
Asking it to translate the text to logic:\\
\includegraphics[width=\columnwidth]{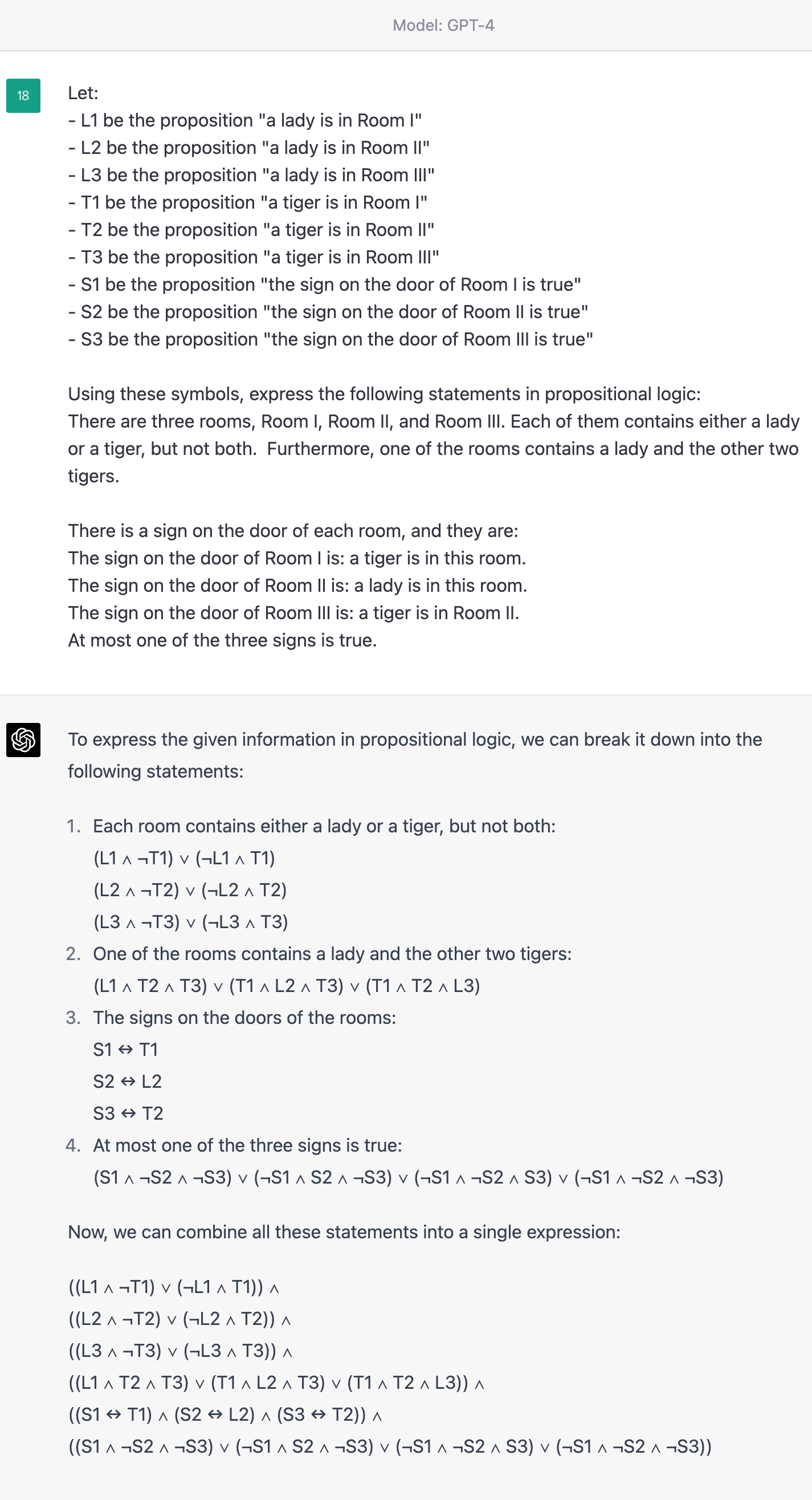}
\newpage\noindent
Asking it to translate the statements in the text one at a time:\\
\includegraphics[width=\columnwidth]{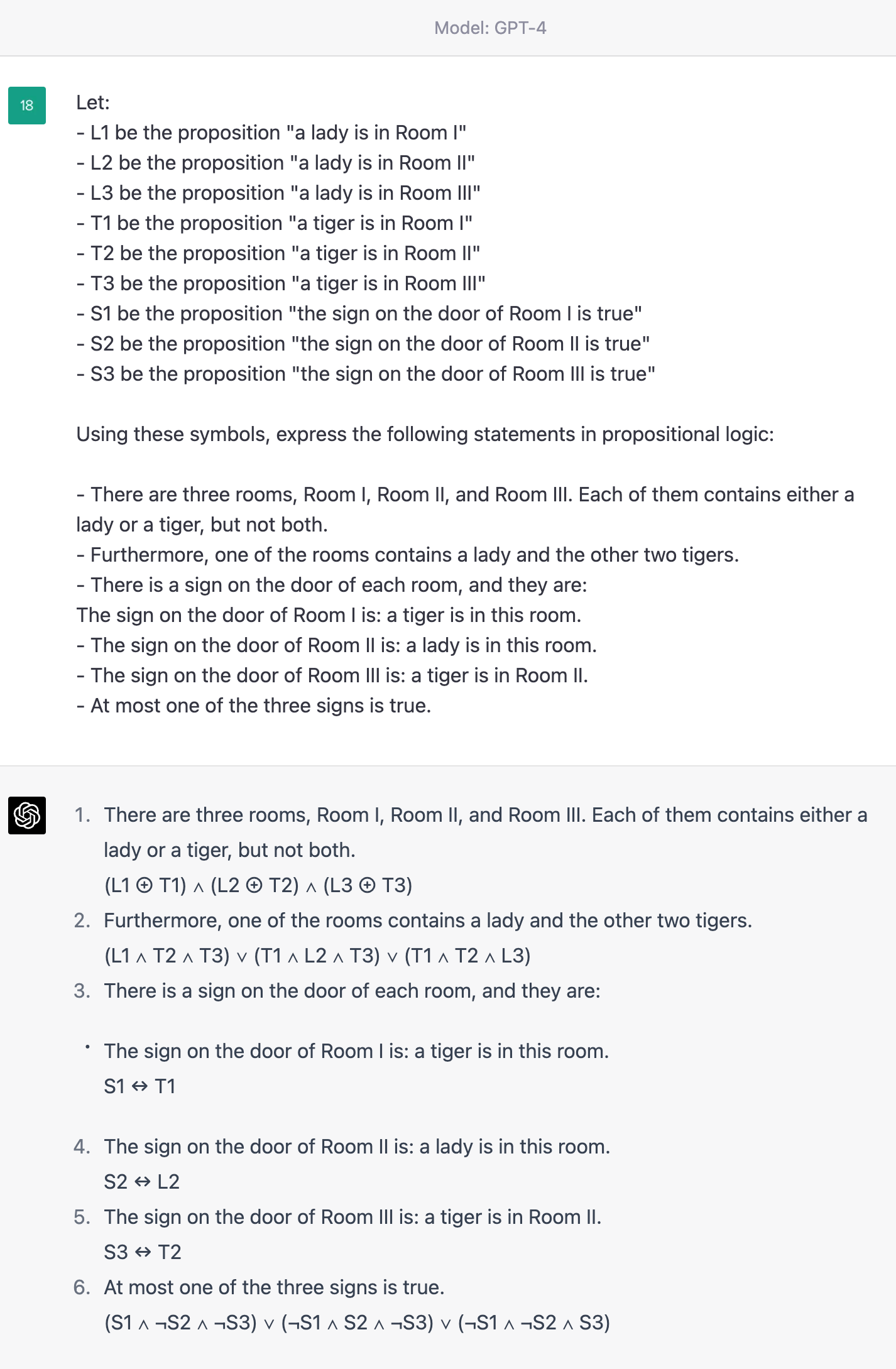}
\\
\subsection*{Trial X}
Direct query to GPT4:\\
\includegraphics[width=\columnwidth]{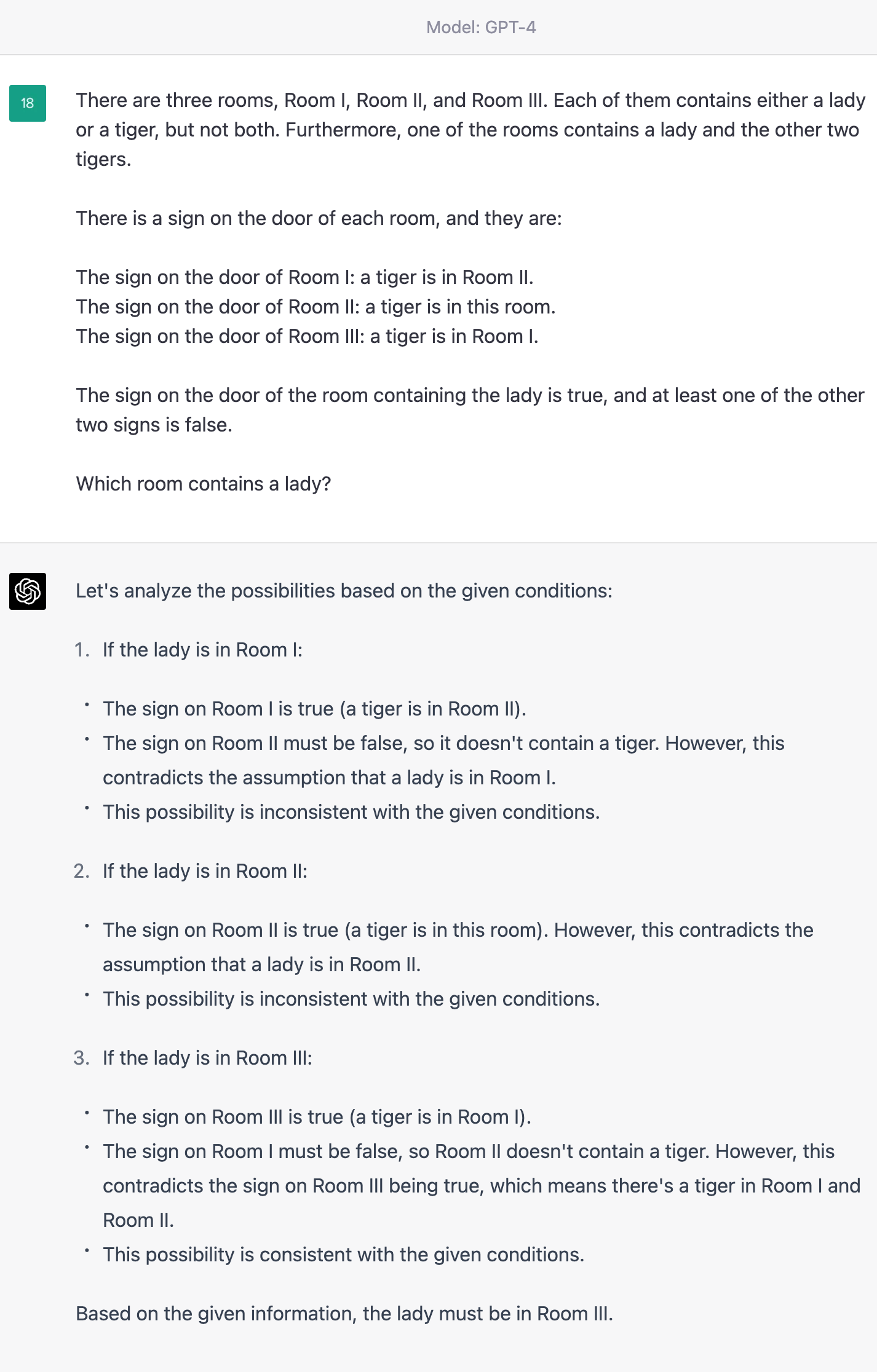}
\newpage
\noindent
Asking it to translate the text to logic:\\
\includegraphics[width=\columnwidth]{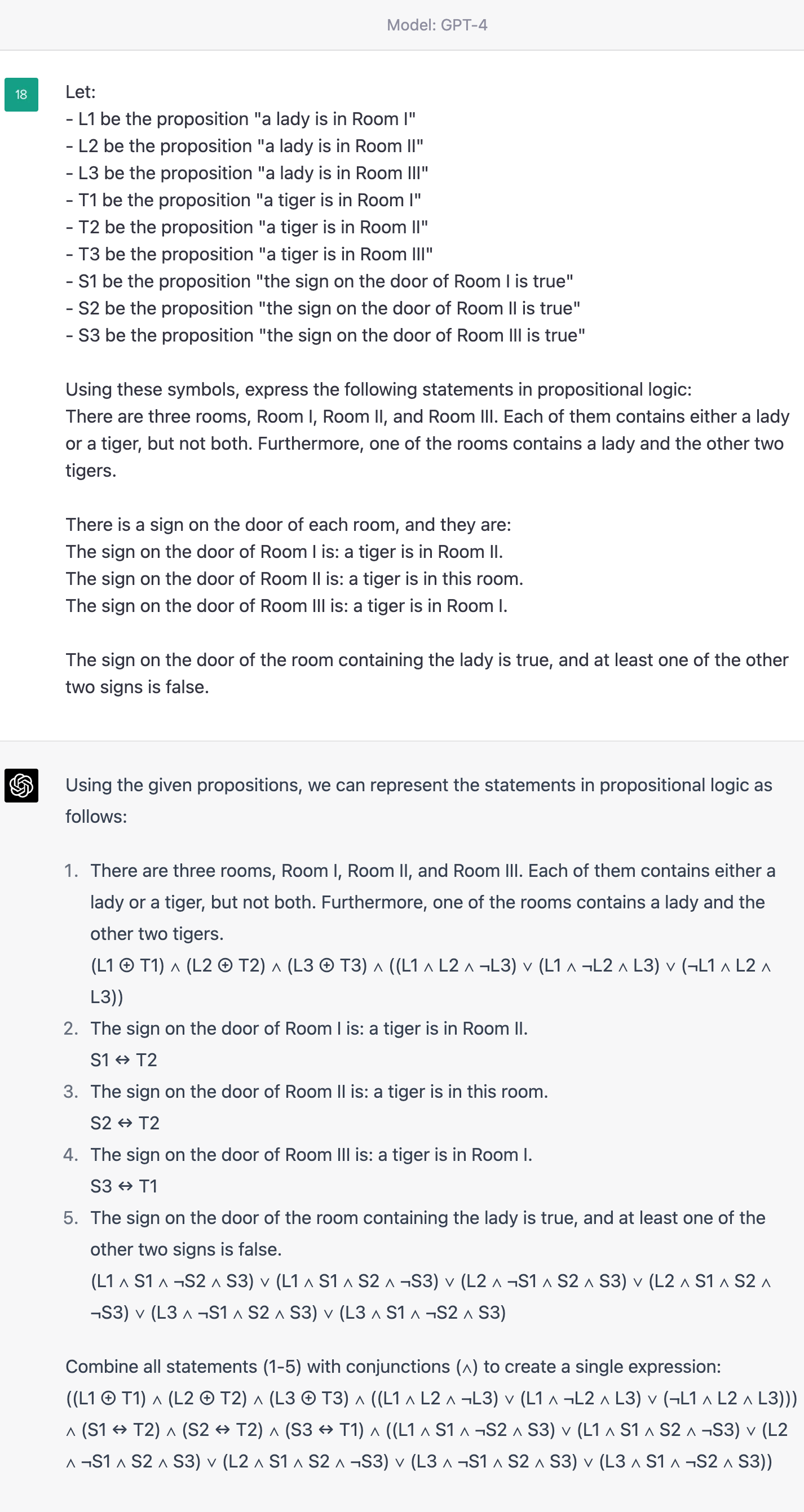}
\newpage\noindent
Asking it to translate the statements in the text one at a time:\\
\includegraphics[width=.89\columnwidth]{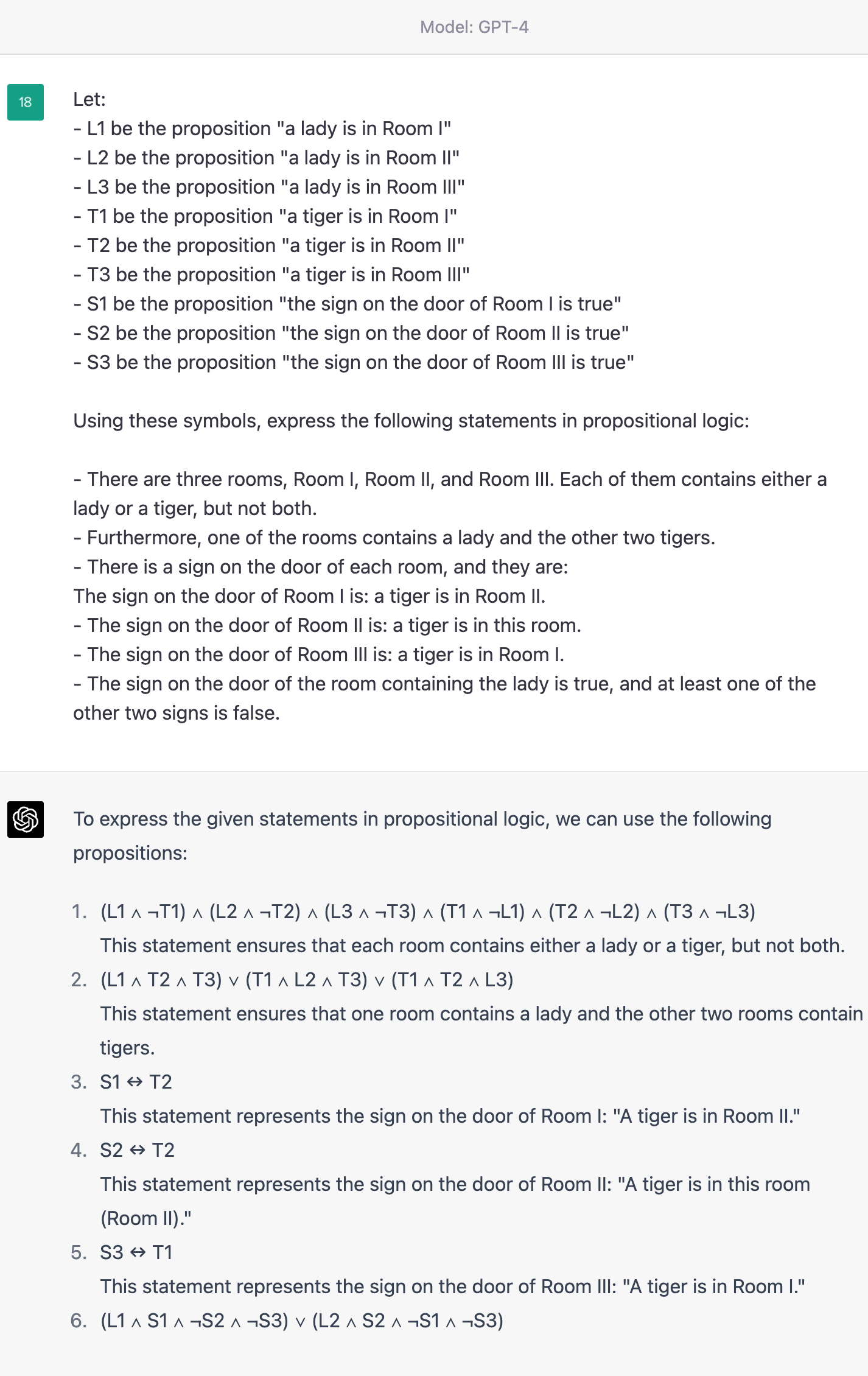}
\subsection*{Trial XI}
Direct query to GPT4:\\
\includegraphics[width=.89\columnwidth]{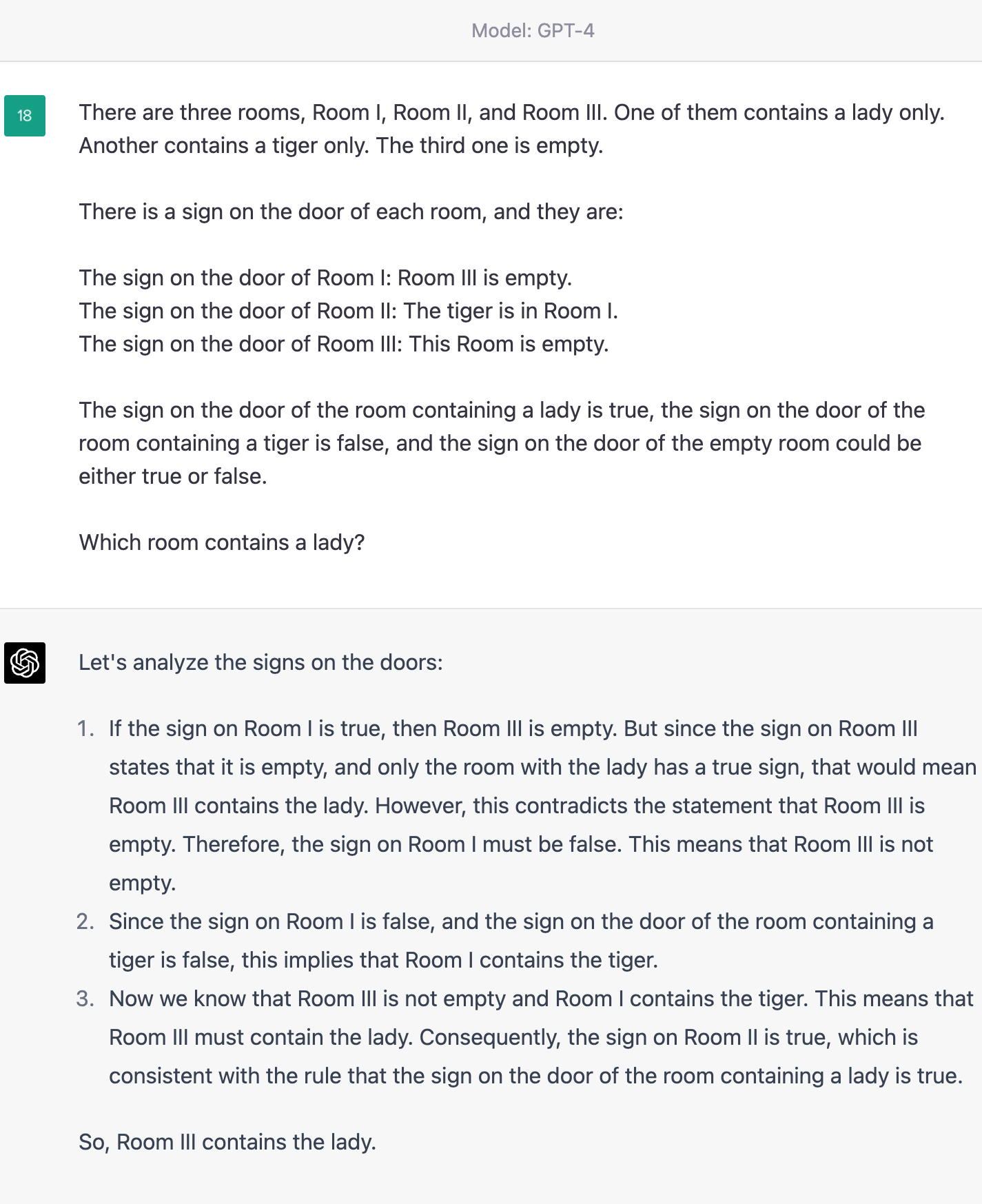}
\\
Asking it to translate the text to logic:\\
\includegraphics[width=\columnwidth]{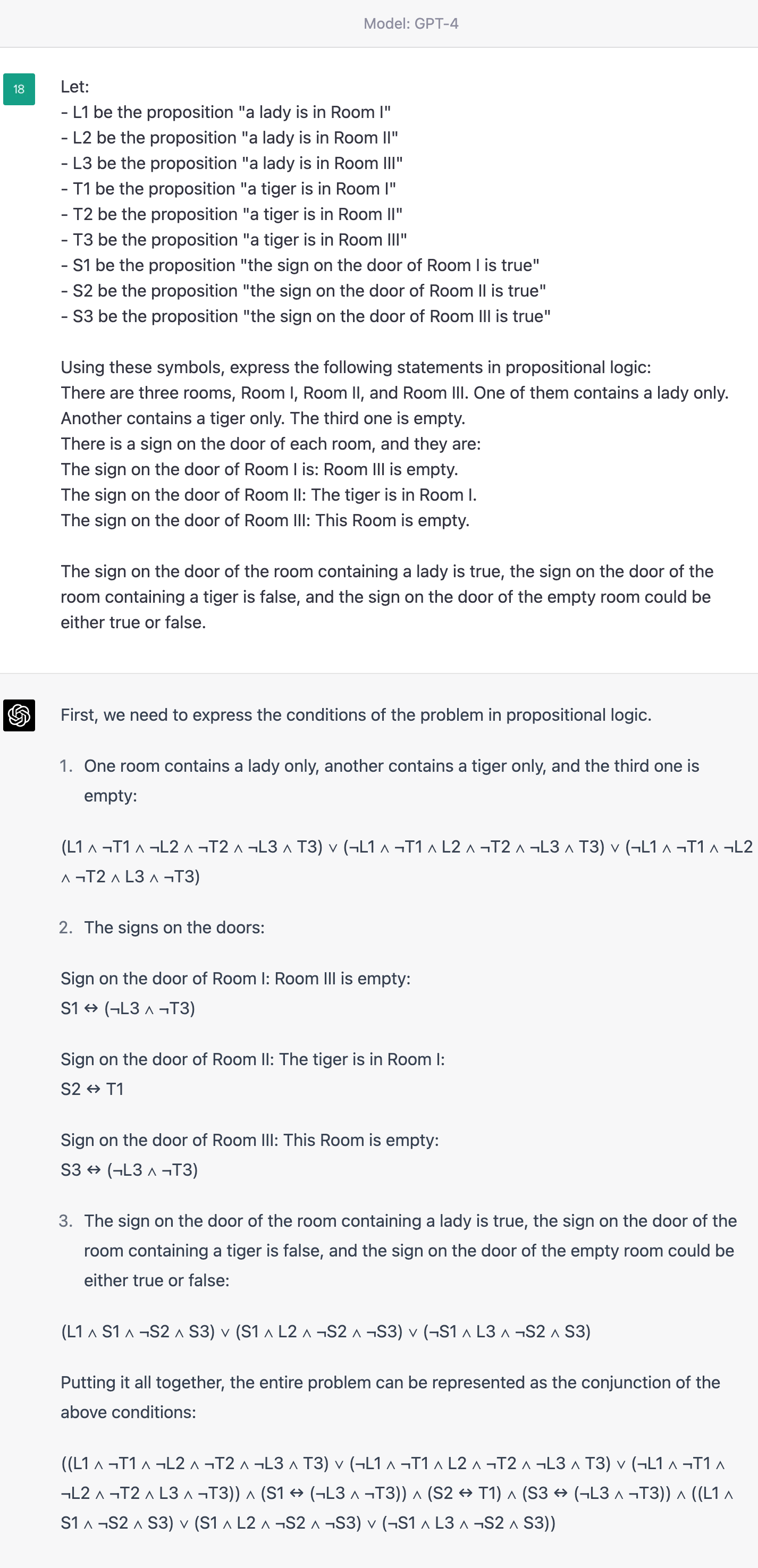}
\newpage\noindent
Asking it to translate the statements in the text one at a time:\\
\includegraphics[width=\columnwidth]{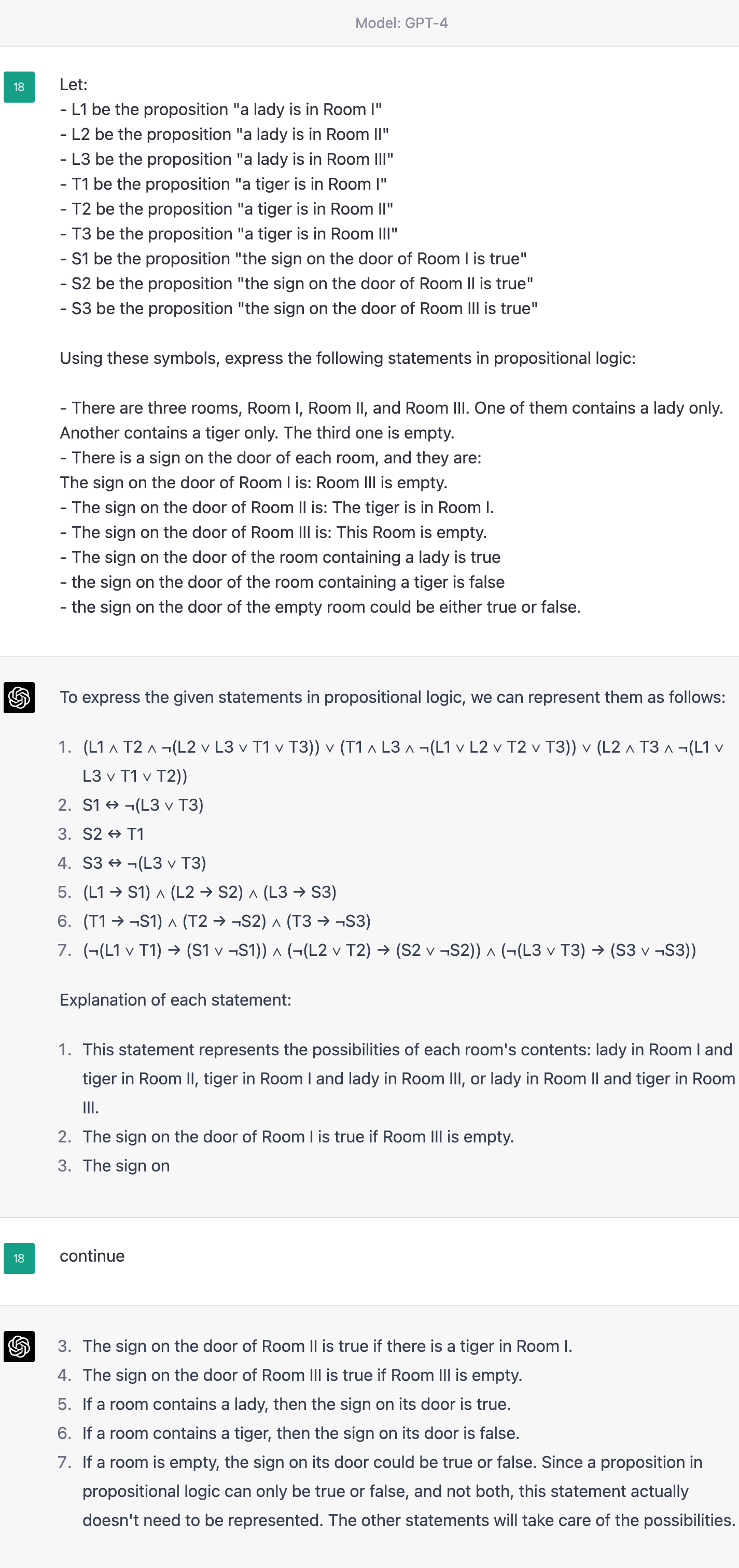}
\newpage
\subsection*{Trial XII}
Direct query to GPT4:\\
\includegraphics[width=.95\columnwidth]{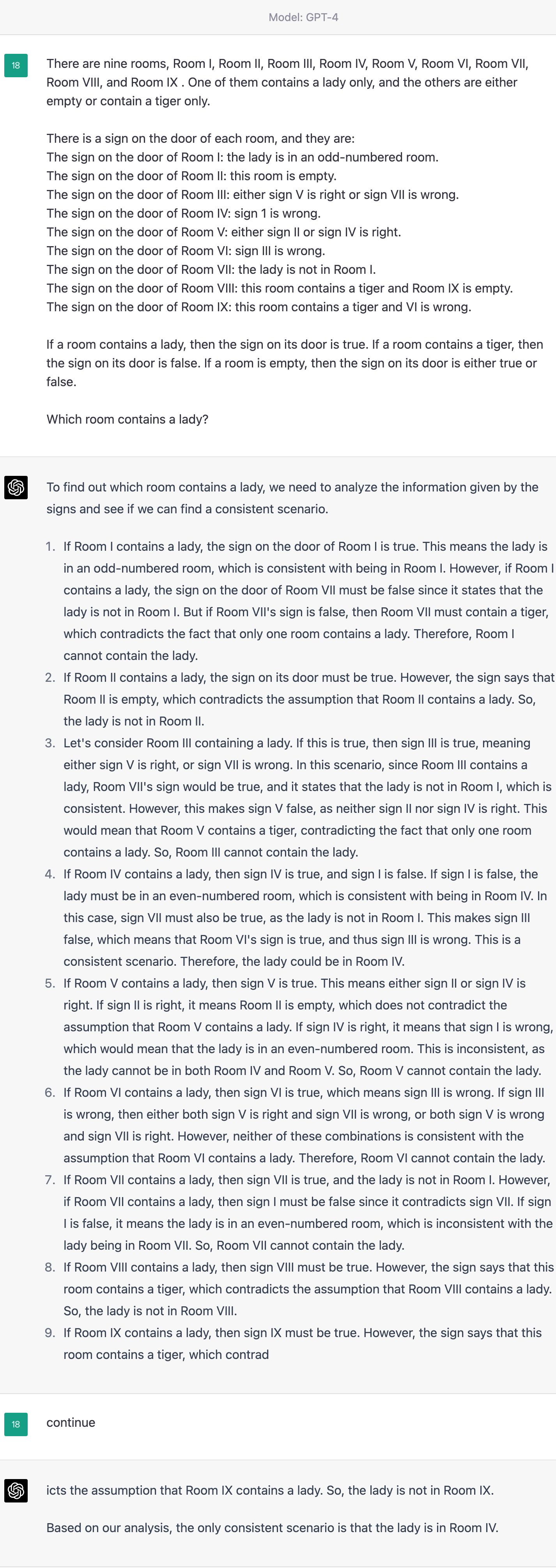}
\\
Asking it to translate the text to logic:\\[-.1ex]
\includegraphics[width=.9\columnwidth]{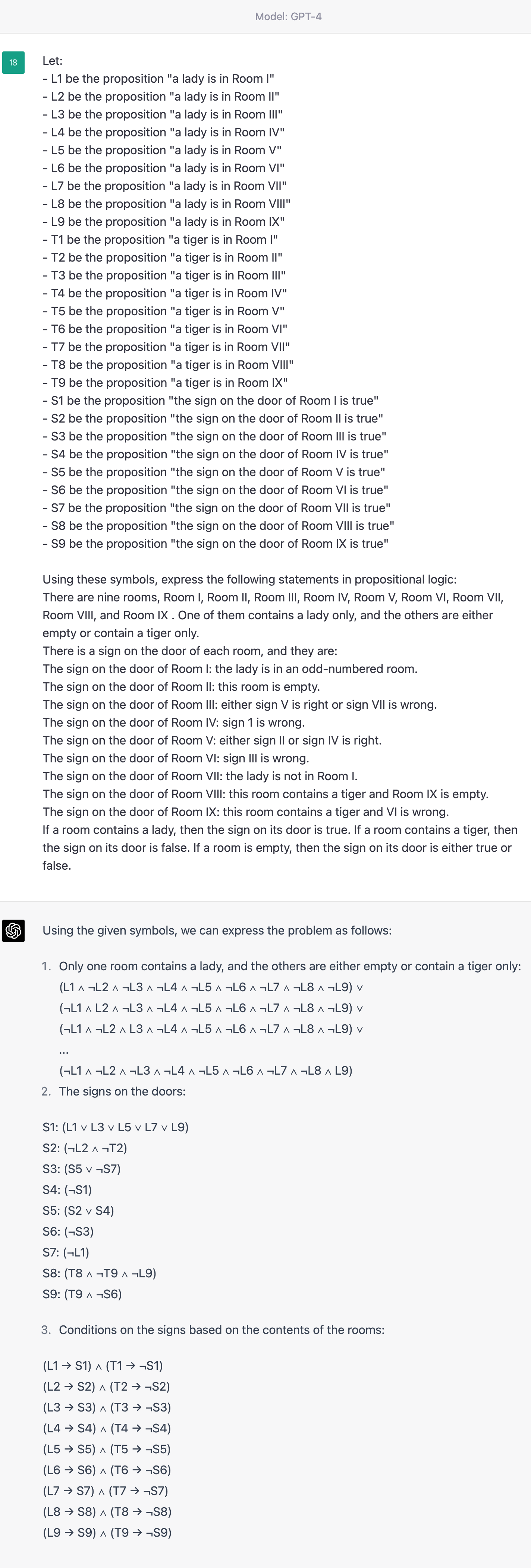}
\\
Asking it to translate the statements in the text one at a time:\\
\includegraphics[width=.9\columnwidth]{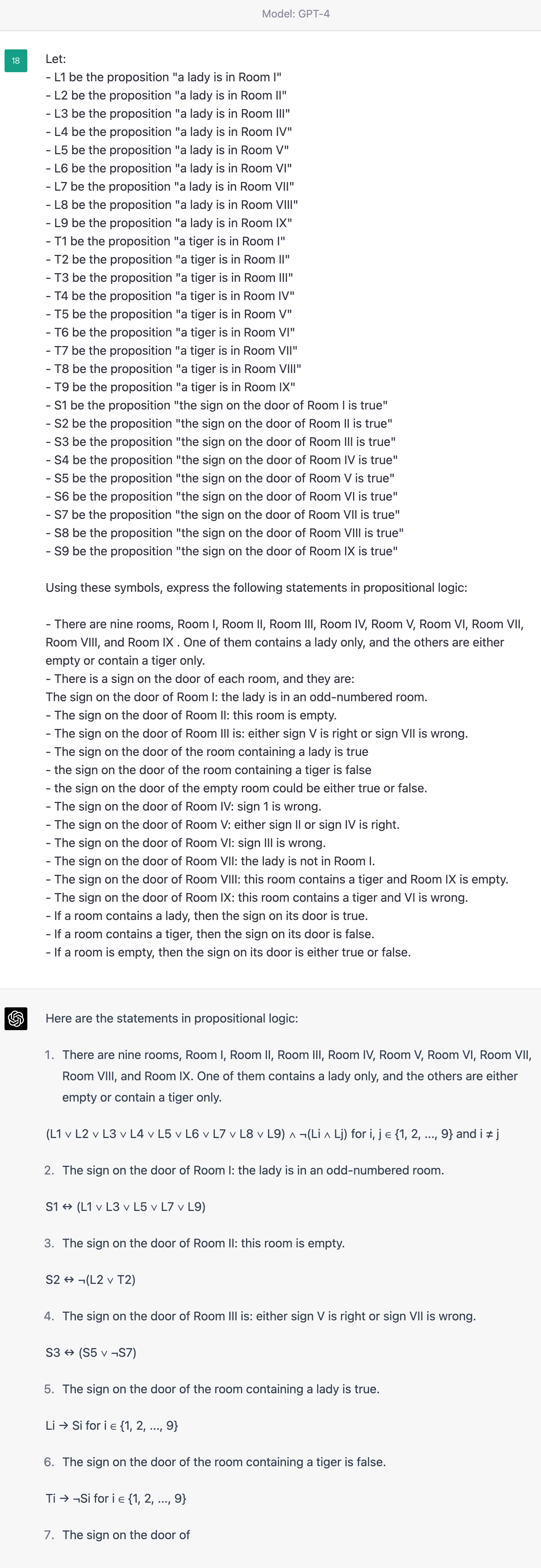}
\includegraphics[width=\columnwidth]{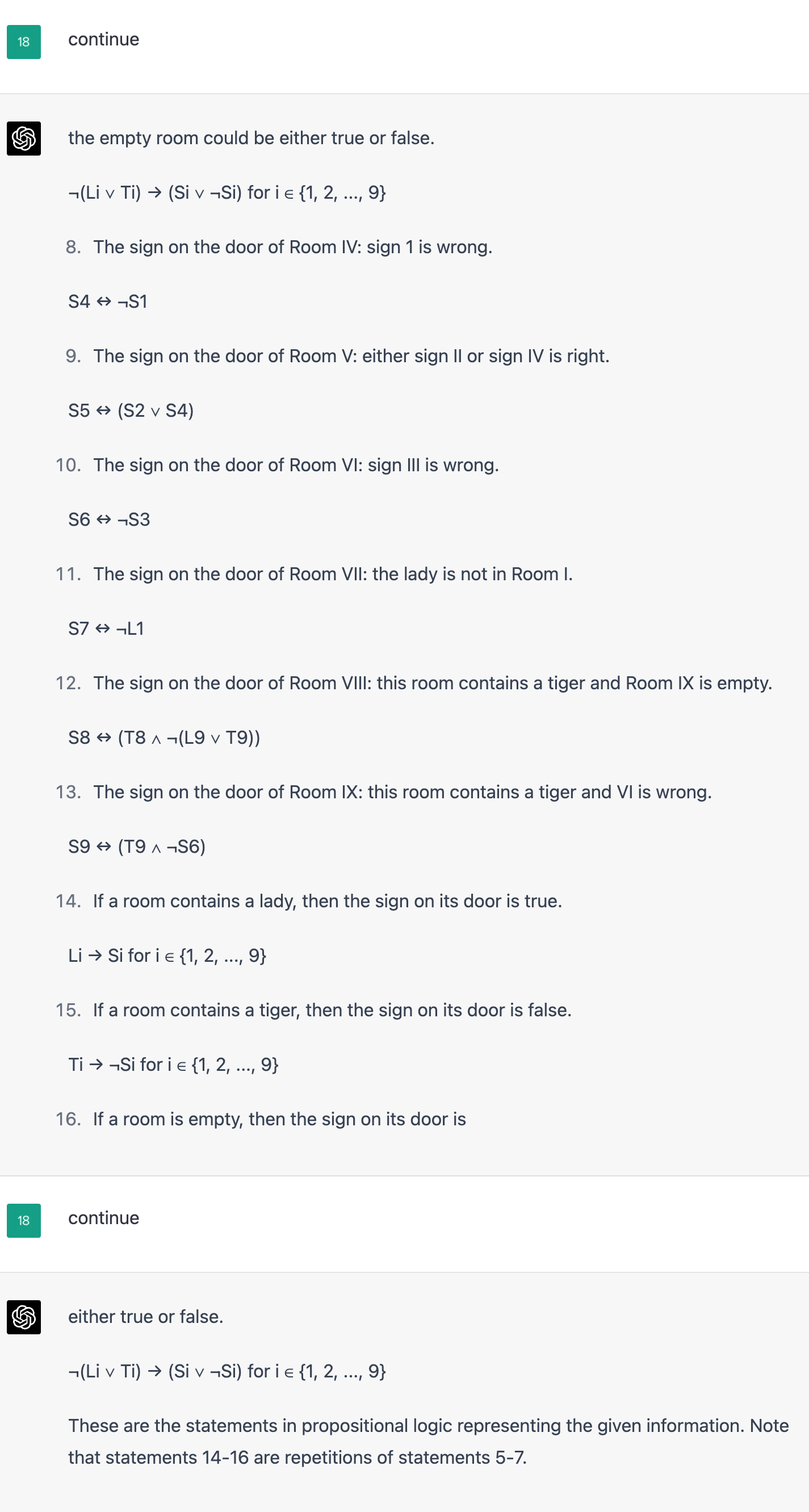}

\end{document}